\documentclass{article}

\PassOptionsToPackage{numbers,sort&compress}{natbib}
\usepackage[preprint]{neurips_2026}

\makeatletter
\renewcommand{\@noticestring}{%
  Accepted at the 40th Conference on Neural Information Processing Systems (NeurIPS 2026).%
}
\makeatother

\usepackage[T1]{fontenc}
\usepackage[utf8]{inputenc}
\usepackage{amsmath,amssymb,amsthm,mathtools}
\usepackage{array,tabularx}
\usepackage{booktabs}
\usepackage{enumitem}
\usepackage{graphicx}
\usepackage{placeins}
\usepackage{xcolor}
\usepackage{hyperref}
\usepackage{url}

\hypersetup{
  colorlinks=true,
  linkcolor=blue!50!black,
  citecolor=blue!50!black,
  urlcolor=blue!50!black
}

\newtheorem{theorem}{Theorem}
\newtheorem{assumption}{Assumption}
\newtheorem{corollary}{Corollary}
\newtheorem{proposition}{Proposition}
\newtheorem{lemma}{Lemma}
\theoremstyle{definition}
\newtheorem{definition}{Definition}
\newtheorem{example}{Example}
\theoremstyle{remark}

\theoremstyle{plain}
\newtheorem{repeatedtheorem}{Theorem}

\DeclareMathOperator{\odds}{odds}

\newcommand{\E}{\mathbb{E}}

\title{Hidden Tails: Certifying Tail-Risk Claims under Selective Labels}
\author{%
  El Mustapha Mansouri \\
  School of Engineering, Institute of Science Tokyo \\
  Yokohama, Kanagawa 226-8501, Japan \\
  \texttt{mansouri.e.2224@m.isct.ac.jp}
}
\date{}

\begin{document}
\maketitle
\begin{abstract}
Historical decisions determine which outcomes are observed, potentially concealing a predictor’s largest errors. Even unlimited data from the same observation process can therefore leave population tail risk unresolved. Here we derive sharp bounds on conditional value-at-risk, the average loss in a specified upper tail, for binary outcomes and bounded losses. These bounds span exactly the risks compatible with the observations and a stated limit on how outcome odds differ between observed and hidden cases at the same covariate values. Under stated regularity conditions, calibrated statistical certificates cover these ranges simultaneously across candidate predictors, supporting risk limits and model comparisons while separating statistical uncertainty from persistent ambiguity. For a fixed predictor and loss threshold, we derive a minimum-cost allocation for reducing excess-loss ambiguity by prioritizing hidden outcomes according to ambiguity removed per unit cost. This allocation treats audited population contributions as known and retains the original sensitivity bounds separately for audited and still-hidden outcomes. In controlled masking experiments with known loan repayments, recovering about 16\% of hidden labels removes a median 61\% of estimated fixed-threshold ambiguity, compared with 16\% for random recovery. These results connect population tail-risk claims to the available evidence and identify where additional outcomes can sharpen their assessment.
\end{abstract}

\section{Introduction}
\label{sec:introduction}
Population tail-risk claims guide decisions about whether a predictor meets an acceptable loss limit and whether one predictor has lower tail risk than another. Yet historical decisions determine which outcomes are available for evaluation \citep{kleinberg2015prediction,kleinberg2018human,lakkaraju2017selective}, and the missing cases can carry the largest prediction losses. For example, a hospital assessing a model for 30-day adverse events may lack event status for patients who receive subsequent care outside its network; record linkage can later recover those labels. Conditional value-at-risk (CVaR) averages the upper tail of prediction loss \citep{acerbi2002expected,artzner1999coherent,rockafellar2000optimization}. Certifying this quantity for all target patients therefore requires an explicit account of hidden outcomes.

The obstacle persists even if the observed distribution is known exactly. Suppose labels are
observed for 25\% of a population with constant covariates, with equal proportions of zeros and ones
among observed outcomes. Under zero-one loss and CVaR at \(\alpha=.5\), the constant predictors
\(f=0\) and \(g=1\) have identical observed loss distributions. If every hidden label is zero, their
population CVaRs are \((.25,1)\); if every hidden label is one, they are \((1,.25)\). Both worlds
produce the same records and have positive observation probability. This illustration allows
unrestricted hidden outcomes and shows why historical records alone cannot certify an ordering
(Appendix~\ref{app:examples}).

\begin{figure}[t]
\centering
\includegraphics[width=\textwidth]{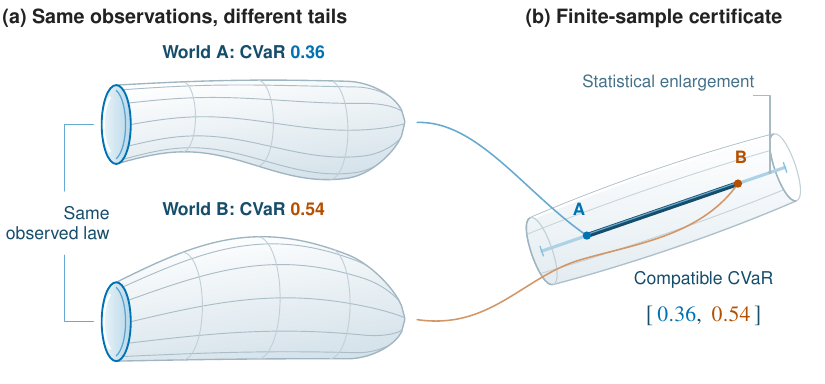}
\caption{\textbf{Risk ambiguity and statistical uncertainty for one fixed predictor.} (a) The same observed law is compatible with hidden-outcome laws giving population CVaR 0.36 and 0.54. (b) The dark segment is the sharp population interval; pale extensions account for estimating its endpoints from a finite sample. Colors link the laws to the endpoints. Geometry is schematic. Appendix~\ref{app:examples} gives the numerical construction.}
\label{fig:framework}
\end{figure}

We make the assumptions needed for certification explicit through a finite, externally specified
sensitivity level \(\Gamma\), which limits how much outcome odds can differ between hidden and
observed cases at the same covariates. The compatible population laws yield a sharp CVaR interval:
every value in it is attained under the observations and the restriction. Its width measures
ambiguity that persists even with unlimited historical data. Figure~\ref{fig:framework} depicts this
range for one fixed predictor; the additional enlargement accounts for estimating its endpoints from
a finite sample.

Our goal is to connect three questions: which risks are compatible with the evidence, how precisely
their bounds can be estimated, and which additional labels most reduce unresolved ambiguity. An
upper bound below a chosen loss limit certifies that criterion; separated intervals establish a
predictor ordering. When these rules leave a decision unresolved, targeted outcome collection can
reduce the underlying information gap.

\paragraph{Contributions.}
\begin{enumerate}
\item \textbf{Sharp population identification.} For binary outcomes and bounded losses, we derive attainable CVaR endpoints and show that every intermediate risk is compatible. For each endpoint, one hidden-outcome law is extremal at every loss threshold, reducing its calculation to a scalar convex minimization (Section~\ref{sec:population}).

\item \textbf{Uniform finite-sample certification.} Paired, locally robust endpoint estimates give simultaneous bounds across candidate predictors at fixed sensitivity and tail levels. The bounds expose the costs of assessing a smaller tail, weaker observation overlap, and auxiliary-estimation error, and support comparisons and predictor selection using the same data (Section~\ref{sec:learning}).

\item \textbf{Minimum-cost revelation.} Under a specified information model, ranking tail identification value per unit cost gives the exact optimal revelation allocation for a fixed predictor and threshold. A zero-threshold bound also controls optimized interval width (Section~\ref{sec:revelation}).
\end{enumerate}

Revelation recovers existing outcomes for a fixed target population. Our information model treats
audited contributions as known and retains the original sensitivity bounds separately for audited
and still-hidden groups.

Synthetic experiments separate persistent identification width from decreasing endpoint error
(Figure~\ref{fig:finite-sample}). Historical LendingClub data illustrate changes in population
assessment, while controlled masking of known outcomes evaluates coverage and comparisons
(Figure~\ref{fig:lendingclub}). Fixed and sequential audits examine how acquisition choices change
estimated ambiguity (Figure~\ref{fig:revelation-results}).

\paragraph{Positioning.}
This work connects selective-label evaluation \citep{chen2025selective,coston2021fairness,lakkaraju2017selective} and partial identification \citep{manski1989anatomy,manski2003partial,scharfstein1999adjusting,tan2006distributional,zhao2019sensitivity} with CVaR optimization \citep{delage2010distributionally,duchi2021uniform,rockafellar2002conditional} and orthogonal estimation \citep{chernozhukov2018double,kallus2024localized,robins1994estimation}. Appendix~\ref{app:connections} develops these connections. We first make the target risk and the assumptions linking observed and hidden outcomes precise.

\section{Tail Risk under Selective Labels}
\label{sec:setup}

Tail-risk certification requires a target population and loss, a criterion for large errors, and a restriction on hidden outcomes. CVaR measures average loss within a chosen upper tail, making the severity of the largest prediction errors part of the assessment. The observation model specifies what historical records identify; the sensitivity model bounds how observed and hidden outcome probabilities can differ. Together, these choices define the population risks that our certificates must cover.

\subsection{Target population and observation process}
\label{sec:target}
Let \(\mathcal F\) be the candidate predictor class and \(P\) the full-data distribution of \((X,S,Y)\), with covariate marginal \(P_X\). We evaluate \(f\in\mathcal F\) using loss \(\ell(f(X),Y)\), where \(Y\in\{0,1\}\). In decision settings, \(Y\) denotes \(Y(a^\star)\) under one specified action held fixed across the target population and predictor comparisons. The indicator \(S\in\{0,1\}\) equals one when the outcome is observed. We observe \(n\) independent copies of \(O=(X,S,Y^{\mathrm{obs}})\), where \(Y^{\mathrm{obs}}=SY\). The observed law \(P_{\mathrm{obs}}\) identifies \(P_X\) and
\[
 \pi(x)=P(S=1\mid X=x),
 \qquad p_1(x)=P(Y=1\mid X=x,S=1),
\]
while \(p_0(x)=P(Y=1\mid X=x,S=0)\) is hidden wherever \(\pi(x)<1\). Here \(\pi\) is the historical observation probability and \(p_1\) the outcome probability among observed cases; both are auxiliary to the predictor \(f\) being assessed.

Bounded loss gives a common risk scale and supports finite-sample error control. Population overlap makes the observed-outcome distribution available as a reference throughout the target covariate population. We formalize these conditions as follows.

\begin{assumption}[Bounded loss and population overlap]
\label{ass:regularity}
For every \(f\in\mathcal F\) and \(y\in\{0,1\}\), the map \(x\mapsto\ell(f(x),y)\) is measurable and takes values in \([0,1]\), and \(0<\pi_{\min}\le\pi(X)\le1\) almost surely.
\end{assumption}

Technical regularity and target-label conventions are in Appendix~\ref{app:compatibility}. Write \(z_+=\max\{z,0\}\) and \(L_f=\ell(f(X),Y)\). At level \(\alpha\in(0,1)\), \(\beta=1-\alpha\) is the upper-tail fraction. The threshold \(\eta\) is a candidate loss cutoff, and \((L_f-\eta)_+\) is the excess above it. At \(\alpha=.95\), CVaR averages the worst 5\% of population losses, allowing fractional mass at a tied cutoff \citep{rockafellar2002conditional}:
\begin{equation}
 \rho_{\alpha,P}(f)=\operatorname{CVaR}_{\alpha,P}(L_f)
 =\inf_{\eta\in[0,1]}\left\{\eta+\beta^{-1}\mathbb E_P[(L_f-\eta)_+]\right\}.
\label{eq:cvar-definition}
\end{equation}
The expectation includes both observed and hidden cases.

\subsection{Sensitivity model and certification objective}
\label{sec:sensitivity}
Outcome-dependent observation can change label prevalence even at fixed covariates. We use the observed outcome probability as a reference and express the allowed departure of hidden outcomes through a multiplicative bound on their odds. For a finite, externally specified \(\Gamma\ge1\), we impose the hidden-to-observed outcome-odds restriction \citep[Section~7.2]{zhao2019sensitivity}:
\begin{equation}
\begin{gathered}
 p_0(x)\in[\underline p_\Gamma(x),\overline p_\Gamma(x)]
 =[q_-(p_1(x)),q_+(p_1(x))],\qquad \pi(x)<1,\\
 q_-(p)=\frac{p}{p+\Gamma(1-p)},
 \qquad q_+(p)=\frac{\Gamma p}{1-p+\Gamma p}.
\end{gathered}
\label{eq:sensitivity-model}
\end{equation}
For interior probabilities, hidden-to-observed outcome odds lie between \(\Gamma^{-1}\) and \(\Gamma\); the formulas also cover \(p_1\in\{0,1\}\). At \(\Gamma=1\), \(p_0(x)=p_1(x)\): hidden and observed outcomes have the same conditional distribution given \(X\), called conditional ignorability. For example, \(p_1(x)=.5\) and \(\Gamma=2\) allow \(p_0(x)\in[1/3,2/3]\). Larger \(\Gamma\) permits greater departure. The chosen \(\Gamma\) defines a declared sensitivity scenario, supported by domain knowledge or additional outcome information (Appendix~\ref{app:gamma-calibration}).

Let \(\mathcal P_\Gamma(P_{\mathrm{obs}})\) contain all full-data laws with observed marginal \(P_{\mathrm{obs}}\) and hidden regression satisfying this restriction. Their risk endpoints are
\begin{equation}
 \underline\rho_{\alpha,\Gamma}(f)
 =\inf_{P\in\mathcal P_\Gamma}\rho_{\alpha,P}(f),
 \qquad
 \overline\rho_{\alpha,\Gamma}(f)
 =\sup_{P\in\mathcal P_\Gamma}\rho_{\alpha,P}(f).
\label{eq:population-endpoints}
\end{equation}
For fixed \(\alpha\) and \(\Gamma\), our statistical objective is an interval covering every compatible risk simultaneously for every \(f\in\mathcal F\), with probability at least \(1-\delta_{\mathrm{tot}}\) over the observed sample. Here \(\delta_{\mathrm{tot}}\in(0,1)\) is the total failure budget; Section~\ref{sec:learning} accounts for sample fluctuation and auxiliary-estimation error within this budget. Thus \(\Gamma\) specifies the admissible hidden outcomes, while \(\delta_{\mathrm{tot}}\) specifies the statistical guarantee.

We next derive these population risk endpoints from the observed law and show that every value between them is attainable.

\section{Sharp Population CVaR Bounds}
\label{sec:population}
\label{sec:sharp-intervals}
For binary outcomes, both possible prediction losses can be computed from \(f(x)\), even when the actual label is hidden. Sensitivity bounds constrain the probability assigned to these two losses. This reduces population identification to two scalar convex minimizations.

Fix a predictor \(f\) and a cutoff \(\eta\in[0,1]\). Write
\[
 a^y_{f,\eta}(x)=(\ell(f(x),y)-\eta)_+,
 \qquad
 d_{f,\eta}(x)=a^1_{f,\eta}(x)-a^0_{f,\eta}(x).
\]
Here \(a^y\) is the excess loss if the label is \(y\), and \(d\) is its change when the label changes from zero to one. A hidden outcome probability \(q\) gives expected excess loss \((1-q)a^0+qa^1=a^0+qd\). With \(d_+=(d_{f,\eta})_+\) and \(d_-=(-d_{f,\eta})_+\), its largest and smallest allowed values are
\begin{equation}
\begin{aligned}
 b^U_{f,\eta}&=a^0_{f,\eta}+\overline p_\Gamma d_+-\underline p_\Gamma d_-,\\
 b^L_{f,\eta}&=a^0_{f,\eta}+\underline p_\Gamma d_+-\overline p_\Gamma d_-.
\end{aligned}
\label{eq:endpoint-values}
\end{equation}
All quantities are evaluated at the same covariate value. The upper bound assigns the greatest allowed probability to the label with larger loss; the lower bound reverses the assignment. Averaging the historically observed and hidden contributions gives
\[
 m^\sigma_{f,\eta}
 =\mathbb E_{\mathrm{obs}}\!\left[
 S a^{Y^{\mathrm{obs}}}_{f,\eta}(X)+(1-S)b^\sigma_{f,\eta}(X)
 \right],
 \qquad \sigma\in\{L,U\},
\]
where \(\mathbb E_{\mathrm{obs}}\) integrates over the observed-data law. Adding the cutoff term forms the CVaR objectives
\[
 L_{f,\eta}=\eta+\beta^{-1}m^L_{f,\eta},
 \qquad U_{f,\eta}=\eta+\beta^{-1}m^U_{f,\eta}.
\]
Dependence on \(\Gamma\) is implicit.

\begin{theorem}[Sharp population CVaR interval]
\label{thm:cvar-sharp}
Under Assumption~\ref{ass:regularity}, for every \(f\in\mathcal F\) and finite \(\Gamma\ge1\),
\[
 \underline\rho_{\alpha,\Gamma}(f)=\min_{\eta\in[0,1]}L_{f,\eta},
 \qquad
 \overline\rho_{\alpha,\Gamma}(f)=\min_{\eta\in[0,1]}U_{f,\eta}.
\]
Both endpoints and every intermediate value are attained by compatible full-data laws. Consequently, the set of risks \(\{\rho_{\alpha,P}(f):P\in\mathcal P_\Gamma(P_{\mathrm{obs}})\}\) is exactly \([\underline\rho_{\alpha,\Gamma}(f),\overline\rho_{\alpha,\Gamma}(f)]\).
\end{theorem}

The key mechanism is that thresholding preserves the ordering of the two label-specific losses. For
each endpoint, the same hidden-outcome distribution is therefore extremal at every cutoff, including
the optimizing cutoff. Appendices~\ref{app:extremal}--\ref{app:sharpness} give the construction and
sharpness proof, including discrete losses and ties.

The identification width
\[
 I_\Gamma(f)=\overline\rho_{\alpha,\Gamma}(f)-\underline\rho_{\alpha,\Gamma}(f)
\]
is the unresolved range of population risk even when the observed distribution is known exactly. Sharpness means that narrowing this range requires additional information or stronger assumptions. Increasing \(\Gamma\) expands the compatible class and can only widen the interval; at \(\Gamma=1\), the population risk is identified under conditional ignorability.

For a chosen acceptable risk level \(r\), \(\overline\rho_{\alpha,\Gamma}(f)\le r\) certifies that
every compatible risk meets the limit. If
\(\overline\rho_{\alpha,\Gamma}(f)<\underline\rho_{\alpha,\Gamma}(g)\), the ordering holds under
every compatible law. Precision and risk level are distinct: a narrow interval can still lie above
the acceptable limit. Section~\ref{sec:learning} adds sampling uncertainty to these decisions.

\section{Estimating and Using Tail-Risk Certificates}
\label{sec:learning}
The observed distribution must be estimated from a finite sample. We control endpoint error uniformly across the candidate class so that the same certificate event supports comparison and selection after examining the data.

\subsection{Cross-fitted endpoint estimation}
\label{sec:estimation}
The auxiliary functions \(p=p_1\) and \(\pi\) describe the outcome among historically observed cases and the probability of historical observation. They are fitted to estimate the risk of \(f\). Substituting an estimated \(p\) into the sensitivity bounds introduces estimation bias. We correct its leading effect using observed outcome residuals.

Write \(\dot b^\sigma=\partial b^\sigma/\partial p\), for \(\sigma\in\{L,U\}\). The locally robust score combines observed excess loss, the sensitivity-bounded hidden contribution, and the residual correction:
\begin{equation}
 \begin{aligned}
 \phi^\sigma_{f,\eta}(O;p,\pi)
 ={}&S a^{Y^{\mathrm{obs}}}_{f,\eta}(X)
 +(1-S)b^\sigma_{f,\eta}(X;p(X))\\
 &+S\frac{1-\pi(X)}{\pi(X)}
 \dot b^\sigma_{f,\eta}(X;p(X))\{Y^{\mathrm{obs}}-p(X)\}.
 \end{aligned}
\label{eq:robust-score}
\end{equation}
At the true auxiliary functions, its expectation equals \(m^\sigma_{f,\eta}\). The correction leaves quadratic and product error terms, quantified in Section~\ref{sec:uniform-certificates} and Appendix~\ref{app:score-bias}.

Split the sample into \(K\) balanced folds. Fit \(p,\pi\) outside each evaluation fold, then clip their predictions to \([0,1]\) and \([\pi_{\min}/2,1]\), respectively. Thus each score uses auxiliary fits trained without its evaluation observations. Let \(n_{\min}\) be the smallest fold size and \(k(i)\) the fold containing observation \(i\). Define
\begin{equation}
 \widehat\rho^\sigma(f)
 =\inf_{\eta\in[0,1]}
 \left\{\eta+\frac1{n\beta}\sum_{i=1}^n
 \phi^\sigma_{f,\eta}\!\left(
 O_i;\widehat p^{(-k(i))},\widehat\pi^{(-k(i))}
 \right)\right\}.
\label{eq:crossfit-endpoint}
\end{equation}
The endpoints share auxiliary fits and minimize their cutoffs separately. For each fixed \(f\) and realized auxiliary fits, exact cutoff minimization checks \(0,1\), and \(\ell(f(X_i),y)\) for every \(i=1,\ldots,n\) and \(y\in\{0,1\}\) (Proposition~\ref{prop:b:threshold-search}).

\subsection{Uniform certificates and decision guarantees}
\label{sec:uniform-certificates}
To turn these estimates into certificates, we control the errors of both endpoints uniformly over the candidate class. The bound combines sampling fluctuation with auxiliary-estimation error.

Write \(\rho^L=\underline\rho_{\alpha,\Gamma}\) and \(\rho^U=\overline\rho_{\alpha,\Gamma}\).

\begin{theorem}[Uniform statistical certification]
\label{thm:learning}
Fix \(\alpha\in(0,1)\), \(1\le\Gamma<\infty\), \(\delta\in(0,1)\), and \(\delta_\nu\in[0,1-\delta)\). Assume the bounded-loss, overlap, and sampling conditions of Section~\ref{sec:setup} and the cross-fitting and measurability conventions of Appendix~\ref{app:b}. Suppose that, with probability at least \(1-\delta_\nu\), the clipped auxiliary fits satisfy jointly
\[
 \max_k\|\widehat p^{(-k)}-p\|_{L^2(P_X)}\le r_p,
 \qquad
 \max_k\|\widehat\pi^{(-k)}-\pi\|_{L^2(P_X)}\le r_\pi.
\]
For the calibrated radius \(r_n(\delta)\) in \eqref{eq:b:radius}, with probability at least \(1-\delta-\delta_\nu\),
\begin{equation}
 \sup_{f\in\mathcal F}\max_{\sigma\in\{L,U\}}
 |\widehat\rho^\sigma(f)-\rho^\sigma(f)|\le r_n(\delta).
\label{eq:uniform-endpoints}
\end{equation}
\end{theorem}

Allocate \(\delta+\delta_\nu\le\delta_{\mathrm{tot}}\) to obtain the confidence level specified in
Section~\ref{sec:setup}. Here \(\delta\) covers sample fluctuation and \(\delta_\nu\) covers failure
of the joint auxiliary-error bounds. The \(L^2(P_X)\) bounds measure prediction error across the
target covariate distribution, including regions with few observed outcomes. Computing the
calibrated radius requires valid bounds on these errors and on score complexity
(Appendix~\ref{app:uniform-estimation}).

The radius separates sample fluctuation from auxiliary-estimation bias. Smaller tail mass \(\beta\)
amplifies both; weak observation overlap and larger sensitivity levels also enlarge the bound. The
remaining bias contains a squared outcome-regression error \(r_p^2\) and the product \(r_pr_\pi\).
At \(\Gamma=1\), the squared-error coefficient vanishes. For \(\Gamma>1\), accurate outcome
regression remains important even when the observation model is known. Cross-fitting supplies the
conditional independence used for concentration; the auxiliary-error bounds are established
separately.

\paragraph{Outer certificates and decisions.}\label{cor:certificate}
Write \(r_n=r_n(\delta)\). On the same event, simultaneously for all predictors,
\begin{equation}
 [\underline\rho_{\alpha,\Gamma}(f),\overline\rho_{\alpha,\Gamma}(f)]
 \subseteq
 \widehat C_{n,\Gamma}(f)
 :=
 [\max\{0,\widehat\rho^L(f)-r_n\},
  \min\{1,\widehat\rho^U(f)+r_n\}].
\label{eq:outer-certificate}
\end{equation}
The certificate's width lies between \(I_\Gamma(f)\) and \(I_\Gamma(f)+4r_n\). Its upper limit can certify a chosen risk limit. If \(\widehat\rho^U(f)+r_n<\widehat\rho^L(g)-r_n\), then \(f\) has lower CVaR than \(g\) under every compatible law. Minimizing \(\widehat\rho^U\) targets the smallest compatible upper risk in \(\mathcal F\); Appendix~\ref{app:decision-guarantees} gives a \(2r_n\) selection-error bound with an additional term for optimization tolerance.

These decisions retain their guarantee after predictor selection within \(\mathcal F\). As sampling
error shrinks, the identification width remains. Additional labels can reduce that remaining
ambiguity.

\section{Resolving Tail Ambiguity through Revelation}
\label{sec:revelation}
We next choose which hidden outcomes to reveal to reduce ambiguity about population risk. We begin with a fixed predictor and a common loss cutoff for the lower and upper CVaR objectives. Under the information model below, their gap decomposes across the population, making ambiguity reduction per unit cost the basis for a minimum-cost allocation.

\subsection{Tail identification value}
\label{sec:tiv}
Fix \(f,\Gamma\), and a cutoff \(\eta\in[0,1]\). Throughout this section, expectations are over the target covariate distribution \(P_X\). An audit reveals existing outcomes: its indicator \(A\le1-S\) and intensity \(\lambda(x)=P(A=1\mid X=x,S=0)\) specify the share of originally hidden outcomes recovered at \(x\).

We hold the historical observed law and the audited population contribution fixed; the remaining
hidden outcomes retain the specified sensitivity bounds. Formally, let \(p_A(x)=P(Y=1\mid X=x,A=1)\)
and \(q(x)=P(Y=1\mid X=x,S=0,A=0)\). We impose the original interval \([\underline
p_\Gamma(x),\overline p_\Gamma(x)]\) separately on these audited and still-hidden outcome
probabilities wherever their component has positive mass; \(q\) is otherwise free within it. This
defines the retained-sensitivity information model (Appendix~\ref{app:retained-class}).
Identification treats the audited population contribution as known; finite audited samples require
estimation of that contribution.

The retained lower and upper objectives, \(L_{f,\eta}(\lambda)\) and \(U_{f,\eta}(\lambda)\), share the known contribution and differ only on the remaining hidden mass (Appendix~\ref{app:tiv}). Write \(s_\Gamma=\overline p_\Gamma-\underline p_\Gamma\). Their gap is
\begin{equation}
\begin{aligned}
\tau_{f,\eta}(x)&:=\frac{1-\pi(x)}{\beta}s_\Gamma(x)|d_{f,\eta}(x)|,\\
W_{f,\eta}(\lambda)&:=U_{f,\eta}(\lambda)-L_{f,\eta}(\lambda)\\
&=\mathbb E[(1-\lambda)\tau_{f,\eta}]
=W_{f,\eta}(0)-\mathbb E[\lambda\tau_{f,\eta}].
\end{aligned}
\label{eq:tiv-identity}
\end{equation}
The tail identification value (TIV) density \(\tau_{f,\eta}\) combines three quantities: the population share with hidden outcomes, the allowed range of their outcome probability, and the change in excess loss between the two labels. It measures ambiguity per unit target-population mass. Revealing a fraction \(\lambda\) removes exactly \(\mathbb E[\lambda\tau_{f,\eta}]\), a value calculable before seeing the audited labels. When both labels produce the same excess loss, this value is zero even if the outcome is uncertain.

\subsection{Minimum-cost revelation and certificate precision}
\label{sec:frontier}
Let \(c(x)\ge0\) be an integrable cost density per unit target-population mass, so allocation \(\lambda\) costs \(\mathbb E[c\lambda]\). If one already-hidden label costs \(\kappa(x)\), then \(c=(1-\pi)\kappa\). For a target gap \(\varepsilon\ge0\), the minimum cost is
\begin{equation}
 \operatorname{Cert}_{f,\eta}(\varepsilon)
 :=\inf_{0\le\lambda\le1}
 \{\mathbb E[c\lambda]:W_{f,\eta}(\lambda)\le\varepsilon\}.
\label{eq:cert-frontier}
\end{equation}

\begin{theorem}[Minimum-cost fixed-threshold revelation]
\label{thm:cert-frontier}
Under the retained-sensitivity model, this infimum is attained. An optimum reveals zero-cost positive-TIV locations first and, on \(c>0\), satisfies
\[
 \lambda^\star(x)=
 \begin{cases}
 1,&\tau_{f,\eta}(x)/c(x)>t,\\
 0,&\tau_{f,\eta}(x)/c(x)<t,
 \end{cases}
\]
with fractional revelation on ties, for some \(t\in[0,\infty]\).
\end{theorem}

Thus acquisition follows decreasing ambiguity reduction per unit cost until the target is reached, using a partial terminal tie group when needed. This is the fractional-knapsack rule \citep{dantzig1957discrete}. Appendices~\ref{app:frontier} and~\ref{app:finite-cell} give the proof, boundary cases, and finite-cell computation. Among already-hidden candidates with positive label cost, its operational score is
\[
 \frac{\tau_{f,\eta}(x)}{c(x)}
 =\frac{s_\Gamma(x)|d_{f,\eta}(x)|}{\beta\kappa(x)}.
\]
The hidden-population fraction cancels because the candidate is already known to be hidden.

\label{sec:sequential}
We now connect the fixed-cutoff gap to the width of the optimized CVaR interval. The retained endpoints minimize their cutoffs separately, giving
\begin{equation}
 I^{\mathrm{ret}}_{\Gamma,\lambda}(f)
 :=\min_{\eta\in[0,1]}U_{f,\eta}(\lambda)
   -\min_{\eta\in[0,1]}L_{f,\eta}(\lambda)
 \le W_{f,0}(\lambda).
\label{prop:uniform-threshold}
\end{equation}
Theorem~\ref{thm:cert-frontier} at \(\eta=0\) therefore minimizes cost for the sufficient requirement \(W_{f,0}(\lambda)\le\varepsilon\), which guarantees optimized width at most \(\varepsilon\). Appendix~\ref{app:optimized-width} proves this bound and formulates direct optimization of the optimized interval width.

For sequential acquisition, we use TIV as a ranking rule that adapts as labels accumulate. After each batch, the procedure retrains the predictor and refreshes TIV at the upper-endpoint minimizing cutoff of the retained plug-in objective. Its
historical outcome fit remains fixed, preserving the original sensitivity bounds. Audited labels
enter through their observed losses. Candidates are ranked by estimated
\(s_\Gamma|d_{f,\eta}|/(\beta\kappa)\), a batch is revealed, and the procedure repeats.
Figure~\ref{fig:revelation-results} evaluates the fixed and sequential protocols;
Appendix~\ref{app:repayment-audits} specifies fitting and evaluation splits.

\section{Empirical Evaluation}
\label{sec:experiments}
We evaluate three questions: how accurately endpoints can be estimated, how selective
observation changes model assessment, and which additional labels reduce unresolved
ambiguity. Simulated populations provide exact identification targets. Historical
applications test the consequences of changing the evaluation population. Controlled masking
of known outcomes permits direct checks of estimated risks and audit performance.
Appendix~\ref{app:experiments} provides the complete methods, results, and figure
provenance.

\begin{figure}[!b]
\centering
\includegraphics[width=\textwidth]{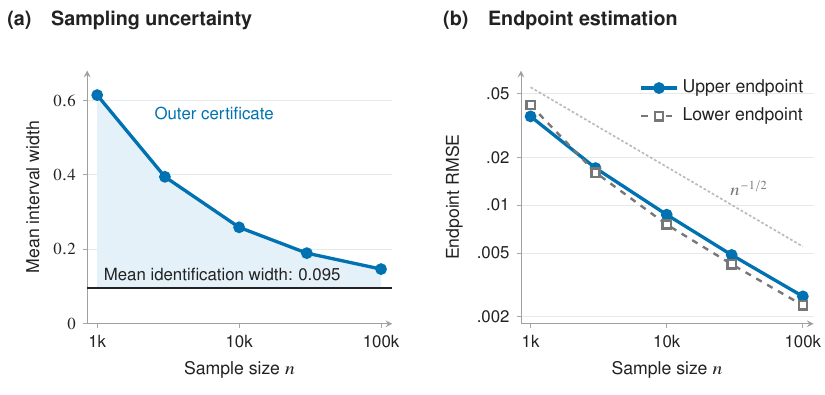}
\caption{\textbf{Statistical uncertainty contracts around an identification floor.} (a) Mean
oracle-calibrated outer width and mean exact population width; shading shows their
difference. (b) Cross-fitted endpoint RMSE with learned nuisances; the faint line is an
$n^{-1/2}$ guide. Each sample size uses 60 simulated populations and 80 datasets per
population. The oracle coverage target is the entire sharp interval.
Appendix~\ref{app:finite-sample-experiments} gives the design and complete results.}
\label{fig:finite-sample}
\end{figure}

\subsection{Identification versus statistical uncertainty}
\label{sec:empirical-uncertainty}

We separate statistical uncertainty from population ambiguity in finite-cell populations
with Brier loss $\ell(f(x),y)=(f(x)-y)^2$, $\Gamma=2$, and $\alpha=.75$. An oracle
certificate uses the true outcome probability $p_1$ among historically observed cases and a
finite-cell Hoeffding radius. On the same sampled datasets, two-fold cross-fitting learns
the nuisance functions; root-mean-square error (RMSE) assesses the resulting endpoint
estimates. Each sample size uses 60 simulated populations and 80 datasets per population.

As $n$ increases from $10^3$ to $10^5$, mean oracle outer width falls from .614 to .147,
while mean exact population width remains .0949 (Figure~\ref{fig:finite-sample}a). The 95\%
oracle certificate contains the entire sharp interval in every replication; its simultaneous
protection across thresholds is conservative. Upper- and lower-endpoint RMSE decrease from
.0361 and .0425 to .0027 and .0023 (panel b). Larger samples sharpen estimation of the
interval, while the original observation process leaves its identification width intact. The
learned-nuisance experiment measures endpoint accuracy; coverage for learned fits requires
the calibration conditions of Theorem~\ref{thm:learning}.

\subsection{Historical applicant-level sensitivity analysis}
\label{sec:historical}
From the LendingClub data \citep{george2019lendingclub}, we construct a 2012 cohort of 390,644 applications, including 53,367 mature issued loans with known repayment outcomes (Appendix~\ref{app:historical-lending}). The applicant target is
default under loan issuance; this outcome is unavailable for rejected applicants. The cohort
includes 180,903 applications with recorded credit-score values below 660, the 2012 FICO
minimum \citep{lendingclub2012registration}, and no issued loans in that group. We therefore
specify an issued-loan outcome regression extended to all recorded applicant covariates. Its
predictions supply the reference probabilities in \eqref{eq:robust-score}, together with the
observed-residual correction. The historical results are residual-adjusted sensitivity
estimates under this regression extension. Appendix~\ref{app:historical-lending} documents
unsupported regions and differences in how source fields were measured.

We compare logistic regression, default-weighted logistic regression, and gradient boosting
with five-fold cross-fitting. Default examples receive weight 2 in the weighted model.
Standard classification objectives represent existing prediction pipelines whose population
tail risks must be assessed. Brier loss is bounded and makes both possible label losses
explicit. We assess tail levels $\alpha\in\{.90,.95\}$.

At $\alpha=.95$, accepted-only CVaR favors the default-weighted model (.661). At $\Gamma=1$,
its estimated applicant risk is .813, compared with .733 for boosting and .805 for logistic
regression: two of three pairwise orderings reverse. Figure~\ref{fig:lendingclub}a gives the
intervals at $\Gamma=2$; Appendix~\ref{app:historical-lending} reports complete sensitivity
curves and approximate breakdown values.

\subsection{Validation with known repayment outcomes}
\label{sec:known-outcomes}
To check estimated intervals and model comparisons against known risks, we impose 20 outcome-dependent masks on the mature issued loans, with sensitivity $\Gamma_0=3$ and approximately 45\% of labels observed. This study targets the issued-loan
reference population. Predictors are trained on observed labels using five-fold
cross-fitting, and their full-outcome CVaRs provide reference risks.

\begin{figure}[t]
\centering
\includegraphics[width=\textwidth]{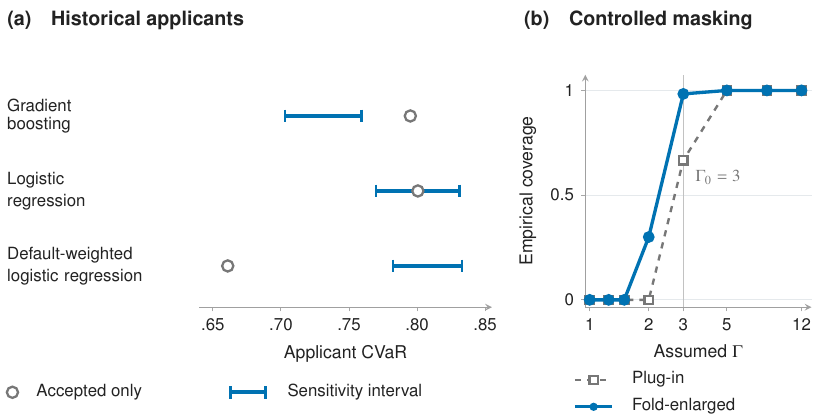}
\caption{\textbf{Observation and sensitivity change population assessment.} (a) Accepted-only CVaR
(hollow points) and estimated applicant sensitivity endpoints (blue segments) at
$\alpha=.95$, $\Gamma=2$, under the regression extension of
Appendix~\ref{app:historical-lending}. Segments precede statistical enlargement. (b)
Coverage of full-outcome reference CVaR under controlled masking of mature issued loans, pooled across three
predictors and 20 shared masks at $\alpha=.95$. The guide marks the masking mechanism's
sensitivity $\Gamma_0=3$; fold enlargement is empirical.
Appendix~\ref{app:controlled-masking} provides the complete results.}
\label{fig:lendingclub}
\end{figure}

At $\Gamma_0$, plug-in intervals---the estimated endpoint intervals before statistical
widening---contain reference CVaR in 11--14 of 20 masks per predictor and tail level.
Enlargement using variability across five fold-specific objectives raises coverage to 19--20
(Figure~\ref{fig:lendingclub}b). Coverage here targets the full-outcome reference CVaR of
each fitted predictor. Fold enlargement provides an empirical calibration diagnostic.
Plug-in coverage is zero at every tested $\Gamma\le2$.

Three unordered predictor pairs in each of 20 masks give 60 comparisons per tail level. At
$\Gamma=3$, $\alpha=.95$, strict interval separation resolves 37 pairs correctly and leaves
23 unresolved. At $\Gamma=1$, the rule declares all 120 orderings across both tail levels
and makes 16 errors. Allowing the masking sensitivity changes both the inferred ordering and
whether separation supports a decision. Appendix~\ref{app:controlled-masking} gives the
mechanism, enlargement rule, and complete results.

\begin{samepage}
\subsection{Fixed and sequential audits}
\label{sec:empirical-revelation}
We next evaluate how targeted revelation reduces the remaining ambiguity.
In 120 synthetic finite-cell populations with heterogeneous costs, TIV per cost matches the
exact fractional-knapsack frontier for fixed predictor and threshold
(Appendix~\ref{app:frontier-experiments}). We then audit the masked issued-loan population
using boosting, $\alpha=.95$, $\Gamma=3$, and unit cost per hidden label.

\end{samepage}

The fixed protocol freezes the initial predictor, historical nuisance estimates, and
upper-endpoint threshold. Newly audited outcomes become known contributions, while the
original sensitivity bounds remain on unaudited outcomes. The sequential protocol retrains
the predictor and reoptimizes thresholds after each batch; retained evaluation keeps the
historical outcome fit fixed, and a reserved split measures predictive loss. The
lowest-propensity baseline selects cases with the lowest estimated label availability; it
uses historical availability in the fixed protocol and refitted current known-label
availability in the sequential protocol.

\begin{figure}[t]
\centering
\includegraphics[width=\textwidth]{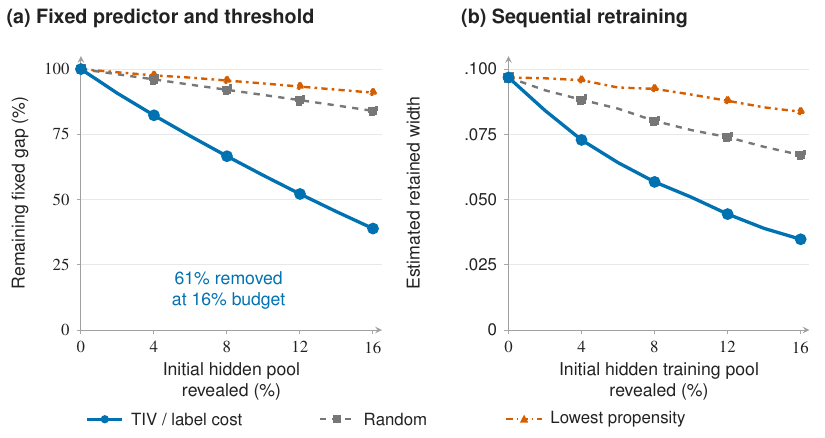}
\caption{\textbf{Targeted audits reduce estimated ambiguity at equal within-protocol budgets.} (a)
Median remaining fixed-threshold gap, normalized within each mask by its initial estimated
gap. (b) Median estimated retained-sensitivity width for the predictor refitted after each
audit batch. Both use boosting, $\alpha=.95$, $\Gamma=3$, unit hidden-label cost, and 20
masks. Lines join eight recorded batches. Appendix~\ref{app:repayment-audits} specifies both
protocols. Budgets are median fractions of the initially hidden cases in each eligible pool: the original pool in (a) and the 80\% training pool in (b).}
\label{fig:revelation-results}
\end{figure}

At the recorded budget of approximately 16\%, TIV removes a median 61\% of the initial
estimated fixed-threshold gap, compared with 16\% for random acquisition
(Figure~\ref{fig:revelation-results}a). This concentration of removable ambiguity explains
the gain from targeting. For Brier loss, predictions near .5 have similar losses under
either label, so uncertainty sampling can favor cases with small excess-loss contrasts and
little identification value.

With sequential retraining, median estimated retained width falls from .097 to .035 under
TIV (panel b). This width describes the current predictor; it can increase when retraining
changes the assessed losses, as the other trajectories illustrate
(Table~\ref{tab:audit-protocols}). Terminal median held-out CVaR is .8541 for TIV and .8682
for random, while mean Brier loss is .1362 and .1347, respectively
(Table~\ref{tab:sequential-performance}). Relative to random acquisition, these runs show
improved tail performance and precision alongside a small mean-loss trade-off.

The constant forecast $f\equiv1/2$ has loss $1/4$ under either label. Its mean Brier loss
and every CVaR therefore equal $1/4$, while identification width and TIV are zero. Its tail
loss is below that of the reported fitted models. Table~\ref{tab:sequential-performance}
shows the competing mean-loss advantage of the audited boosting models, whose mean Brier
losses are near .135. The reference separates average prediction quality, extreme-tail loss,
and certainty about that loss.
Appendix~\ref{app:repayment-audits} gives all policies and both protocols.

\FloatBarrier
\section{Limitations}
\label{sec:limitations}
The sensitivity range must be justified for the target population and covariate resolution. Statistical certificates additionally require overlap, controlled score complexity, and valid nuisance-error bounds. Small tail mass and scarce observed labels can make the bounds conservative. For \(\Gamma>1\), outcome-regression error contributes quadratically even with a correct observation model. Choosing \(\Gamma\) from data or examining multiple sensitivity levels requires corresponding joint error control (Appendix~\ref{app:gamma-calibration}).

The historical applicant analysis depends on regression extrapolation into unsupported regions and
on differently measured source fields. Controlled masking checks known issued-loan outcomes.
Repayment under an untaken lending action requires a causal or experimental model beyond passive
recovery of existing outcomes.

The revelation optimum fixes the predictor and threshold and assumes componentwise retained
sensitivity bounds. Known outcome-independent audit randomization can identify more, and direct
minimization of the reoptimized interval can favor a different allocation. Finite audits must
estimate their newly known contributions. Learned acquisition scores, retraining, and data-dependent
stopping require uncertainty control across the resulting audit history; the sequential trajectories
provide empirical performance evidence. Extensions to richer outcomes, unbounded losses, and other
application domains require additional modeling and validation.

\section{Conclusion}
\label{sec:conclusion}
Selective observation leaves a range of population tail risks compatible with the available outcomes. Sharp CVaR endpoints characterize that range, and calibrated uniform outer certificates support risk limits, model comparisons, and conservative selection within a candidate class. Under retained sensitivity, TIV quantifies the ambiguity removed by revelation for a fixed predictor and threshold. Ranking this value per cost gives the optimal population allocation for that objective. These outputs connect a specified risk decision to the observations and assumptions required to support it, and identify where further outcome collection can improve precision.

\label{main-text-end}
\FloatBarrier
\clearpage
\begin{ack}
The author declares no competing interests.
\end{ack}
\bibliographystyle{unsrtnat}
\bibliography{references}

@article{rockafellar2000optimization,
  author  = {Rockafellar, R. Tyrrell and Uryasev, Stanislav},
  title   = {Optimization of Conditional Value-at-Risk},
  journal = {Journal of Risk},
  volume  = {2},
  number  = {3},
  pages   = {21--41},
  year    = {2000},
  doi     = {10.21314/JOR.2000.038}
}

@article{artzner1999coherent,
  author  = {Artzner, Philippe and Delbaen, Freddy and Eber, Jean-Marc and Heath, David},
  title   = {Coherent Measures of Risk},
  journal = {Mathematical Finance},
  volume  = {9},
  number  = {3},
  pages   = {203--228},
  year    = {1999},
  doi     = {10.1111/1467-9965.00068}
}

@article{rockafellar2002conditional,
  author  = {Rockafellar, R. Tyrrell and Uryasev, Stanislav},
  title   = {Conditional Value-at-Risk for General Loss Distributions},
  journal = {Journal of Banking \& Finance},
  volume  = {26},
  number  = {7},
  pages   = {1443--1471},
  year    = {2002},
  doi     = {10.1016/S0378-4266(02)00271-6}
}

@article{acerbi2002expected,
  author  = {Acerbi, Carlo and Tasche, Dirk},
  title   = {Expected Shortfall: A Natural Coherent Alternative to Value at Risk},
  journal = {Economic Notes},
  volume  = {31},
  number  = {2},
  pages   = {379--388},
  year    = {2002},
  doi     = {10.1111/1468-0300.00091}
}

@article{kleinberg2015prediction,
  author  = {Kleinberg, Jon and Ludwig, Jens and Mullainathan, Sendhil and Obermeyer, Ziad},
  title   = {Prediction Policy Problems},
  journal = {American Economic Review},
  volume  = {105},
  number  = {5},
  pages   = {491--495},
  year    = {2015},
  doi     = {10.1257/aer.p20151023}
}

@article{kleinberg2018human,
  author  = {Kleinberg, Jon and Lakkaraju, Himabindu and Leskovec, Jure and Ludwig, Jens and Mullainathan, Sendhil},
  title   = {Human Decisions and Machine Predictions},
  journal = {The Quarterly Journal of Economics},
  volume  = {133},
  number  = {1},
  pages   = {237--293},
  year    = {2018},
  doi     = {10.1093/qje/qjx032}
}

@inproceedings{lakkaraju2017selective,
  author    = {Lakkaraju, Himabindu and Kleinberg, Jon and Leskovec, Jure and Ludwig, Jens and Mullainathan, Sendhil},
  title     = {The Selective Labels Problem: Evaluating Algorithmic Predictions in the Presence of Unobservables},
  booktitle = {Proceedings of the 23rd ACM SIGKDD International Conference on Knowledge Discovery and Data Mining},
  pages     = {275--284},
  year      = {2017},
  doi       = {10.1145/3097983.3098066}
}

@article{delage2010distributionally,
  author  = {Delage, Erick and Ye, Yinyu},
  title   = {Distributionally Robust Optimization Under Moment Uncertainty with Application to Data-Driven Problems},
  journal = {Operations Research},
  volume  = {58},
  number  = {3},
  pages   = {595--612},
  year    = {2010},
  doi     = {10.1287/opre.1090.0741}
}

@article{duchi2021uniform,
  author  = {Duchi, John C. and Namkoong, Hongseok},
  title   = {Learning Models with Uniform Performance via Distributionally Robust Optimization},
  journal = {The Annals of Statistics},
  volume  = {49},
  number  = {3},
  pages   = {1378--1406},
  year    = {2021},
  doi     = {10.1214/20-AOS2004}
}

@article{mohajerin2018data,
  author  = {Mohajerin Esfahani, Peyman and Kuhn, Daniel},
  title   = {Data-Driven Distributionally Robust Optimization Using the {Wasserstein} Metric: Performance Guarantees and Tractable Reformulations},
  journal = {Mathematical Programming},
  volume  = {171},
  pages   = {115--166},
  year    = {2018},
  doi     = {10.1007/s10107-017-1172-1}
}

@article{blanchet2019quantifying,
  author  = {Blanchet, Jose and Murthy, Karthyek},
  title   = {Quantifying Distributional Model Risk via Optimal Transport},
  journal = {Mathematics of Operations Research},
  volume  = {44},
  number  = {2},
  pages   = {565--600},
  year    = {2019},
  doi     = {10.1287/moor.2018.0936}
}

@inproceedings{coston2021fairness,
  author    = {Coston, Amanda and Rambachan, Ashesh and Chouldechova, Alexandra},
  title     = {Characterizing Fairness Over the Set of Good Models Under Selective Labels},
  booktitle = {Proceedings of the 38th International Conference on Machine Learning},
  series    = {Proceedings of Machine Learning Research},
  volume    = {139},
  pages     = {2144--2155},
  year      = {2021},
  publisher = {PMLR},
  url       = {https://proceedings.mlr.press/v139/coston21a.html}
}

@book{little2019statistical,
  author    = {Little, Roderick J. A. and Rubin, Donald B.},
  title     = {Statistical Analysis with Missing Data},
  edition   = {3rd},
  publisher = {Wiley},
  year      = {2019}
}

@article{manski1989anatomy,
  author  = {Manski, Charles F.},
  title   = {Anatomy of the Selection Problem},
  journal = {Journal of Human Resources},
  volume  = {24},
  number  = {3},
  pages   = {343--360},
  year    = {1989},
  doi     = {10.2307/145818}
}

@article{manski1990nonparametric,
  author  = {Manski, Charles F.},
  title   = {Nonparametric Bounds on Treatment Effects},
  journal = {American Economic Review},
  volume  = {80},
  number  = {2},
  pages   = {319--323},
  year    = {1990}
}

@book{manski2003partial,
  author    = {Manski, Charles F.},
  title     = {Partial Identification of Probability Distributions},
  publisher = {Springer},
  year      = {2003}
}

@article{scharfstein1999adjusting,
  author  = {Scharfstein, Daniel O. and Rotnitzky, Andrea and Robins, James M.},
  title   = {Adjusting for Nonignorable Drop-Out Using Semiparametric Nonresponse Models},
  journal = {Journal of the American Statistical Association},
  volume  = {94},
  number  = {448},
  pages   = {1096--1120},
  year    = {1999},
  doi     = {10.1080/01621459.1999.10473862}
}

@article{tan2006distributional,
  author  = {Tan, Zhiqiang},
  title   = {A Distributional Approach for Causal Inference Using Propensity Scores},
  journal = {Journal of the American Statistical Association},
  volume  = {101},
  number  = {476},
  pages   = {1619--1637},
  year    = {2006},
  doi     = {10.1198/016214506000000023}
}

@article{zhao2019sensitivity,
  author  = {Zhao, Qingyuan and Small, Dylan S. and Bhattacharya, Bhaswar B.},
  title   = {Sensitivity Analysis for Inverse Probability Weighting Estimators via the Percentile Bootstrap},
  journal = {Journal of the Royal Statistical Society: Series B (Statistical Methodology)},
  volume  = {81},
  number  = {4},
  pages   = {735--761},
  year    = {2019},
  doi     = {10.1111/rssb.12327}
}

@book{rosenbaum2002observational,
  author    = {Rosenbaum, Paul R.},
  title     = {Observational Studies},
  edition   = {2nd},
  publisher = {Springer},
  year      = {2002}
}

@article{robins1994estimation,
  author  = {Robins, James M. and Rotnitzky, Andrea and Zhao, Lue Ping},
  title   = {Estimation of Regression Coefficients When Some Regressors Are Not Always Observed},
  journal = {Journal of the American Statistical Association},
  volume  = {89},
  number  = {427},
  pages   = {846--866},
  year    = {1994},
  doi     = {10.1080/01621459.1994.10476818}
}

@article{chernozhukov2018double,
  author  = {Chernozhukov, Victor and Chetverikov, Denis and Demirer, Mert and Duflo, Esther and Hansen, Christian and Newey, Whitney and Robins, James},
  title   = {Double/Debiased Machine Learning for Treatment and Structural Parameters},
  journal = {The Econometrics Journal},
  volume  = {21},
  number  = {1},
  pages   = {C1--C68},
  year    = {2018},
  doi     = {10.1111/ectj.12097}
}

@inproceedings{chen2025selective,
  author    = {Chen, Jian and Li, Zhehao and Mao, Xiaojie},
  title     = {Learning with Selectively Labeled Data from Multiple Decision-makers},
  booktitle = {Proceedings of the 42nd International Conference on Machine Learning},
  series    = {Proceedings of Machine Learning Research},
  volume    = {267},
  pages     = {8480--8519},
  year      = {2025},
  publisher = {PMLR},
  url       = {https://proceedings.mlr.press/v267/chen25al.html}
}

@article{kallus2024localized,
  author  = {Kallus, Nathan and Mao, Xiaojie and Uehara, Masatoshi},
  title   = {Localized Debiased Machine Learning: Efficient Inference on Quantile Treatment Effects and Beyond},
  journal = {Journal of Machine Learning Research},
  volume  = {25},
  number  = {16},
  pages   = {1--59},
  year    = {2024},
  url     = {https://jmlr.org/papers/v25/23-0661.html}
}

@article{bartlett2002rademacher,
  author  = {Bartlett, Peter L. and Mendelson, Shahar},
  title   = {Rademacher and {Gaussian} Complexities: Risk Bounds and Structural Results},
  journal = {Journal of Machine Learning Research},
  volume  = {3},
  pages   = {463--482},
  year    = {2002},
  url     = {https://jmlr.org/papers/v3/bartlett02a.html}
}

@book{van1996weak,
  author    = {van der Vaart, Aad W. and Wellner, Jon A.},
  title     = {Weak Convergence and Empirical Processes},
  publisher = {Springer},
  year      = {1996}
}

@article{dantzig1957discrete,
  author  = {Dantzig, George B.},
  title   = {Discrete-Variable Extremum Problems},
  journal = {Operations Research},
  volume  = {5},
  number  = {2},
  pages   = {266--288},
  year    = {1957},
  doi     = {10.1287/opre.5.2.266}
}

@incollection{yu1997assouad,
  author    = {Yu, Bin},
  title     = {Assouad, {Fano}, and {Le Cam}},
  booktitle = {Festschrift for {Lucien Le Cam}},
  editor    = {Pollard, David and Torgersen, Erik and Yang, Grace L.},
  pages     = {423--435},
  publisher = {Springer},
  year      = {1997},
  doi       = {10.1007/978-1-4612-1880-7_29}
}

@article{pedregosa2011scikit,
  author  = {Pedregosa, Fabian and Varoquaux, Ga{\"e}l and Gramfort, Alexandre and Michel, Vincent and Thirion, Bertrand and Grisel, Olivier and Blondel, Mathieu and Prettenhofer, Peter and Weiss, Ron and Dubourg, Vincent and Vanderplas, Jake and Passos, Alexandre and Cournapeau, David and Brucher, Matthieu and Perrot, Matthieu and Duchesnay, {\'E}douard},
  title   = {Scikit-learn: Machine Learning in {Python}},
  journal = {Journal of Machine Learning Research},
  volume  = {12},
  pages   = {2825--2830},
  year    = {2011},
  url     = {https://jmlr.org/papers/v12/pedregosa11a.html}
}

@inproceedings{maurer2016vector,
  title     = {A Vector-Contraction Inequality for {R}ademacher Complexities},
  author    = {Maurer, Andreas},
  booktitle = {Algorithmic Learning Theory},
  series    = {Lecture Notes in Computer Science},
  volume    = {9925},
  pages     = {3--17},
  publisher = {Springer},
  year      = {2016}
}

@misc{george2019lendingclub,
  author       = {George, Nathan},
  title        = {All {Lending Club} Loan Data},
  howpublished = {Kaggle dataset, \url{https://www.kaggle.com/datasets/wordsforthewise/lending-club}},
  note         = {License: CC0 Public Domain},
  year         = {2019}
}

@inproceedings{chen2016xgboost,
  author    = {Chen, Tianqi and Guestrin, Carlos},
  title     = {{XGBoost}: A Scalable Tree Boosting System},
  booktitle = {Proceedings of the 22nd ACM SIGKDD International Conference on Knowledge Discovery and Data Mining},
  pages     = {785--794},
  publisher = {ACM},
  year      = {2016},
  doi       = {10.1145/2939672.2939785}
}

@article{cohn1996active,
  author  = {Cohn, David A. and Ghahramani, Zoubin and Jordan, Michael I.},
  title   = {Active Learning with Statistical Models},
  journal = {Journal of Artificial Intelligence Research},
  volume  = {4},
  pages   = {129--145},
  year    = {1996},
  url     = {https://arxiv.org/abs/cs/9603104}
}

@misc{lendingclub2012registration,
  author = {{LendingClub Corporation}},
  title = {Amendment No. 3 to Form {S-1} Registration Statement},
  year = {2012},
  month = apr,
  note = {Filed April 5, 2012; minimum credit criteria},
  url = {https://www.sec.gov/Archives/edgar/data/1409970/000119312512151868/d306143ds1a.htm}
}

@misc{lendingclubdictionary,
  author = {{LendingClub}},
  title = {Loan and Rejected-Application Data Dictionary},
  note = {LoanStats and RejectStats worksheets; accessed November 7, 2026},
  url = {https://resources.lendingclub.com/LCDataDictionary.xlsx}
}
\clearpage
\appendix
\numberwithin{theorem}{section}
\numberwithin{equation}{section}
% Auxiliary statements share one appendix-local sequence; restatements keep
% the corresponding main theorem number through mainrestatement.
\makeatletter
\let\c@lemma\c@theorem
\let\c@proposition\c@theorem
\let\c@corollary\c@theorem
\let\c@definition\c@theorem
\let\c@example\c@theorem
\let\c@remark\c@theorem
\makeatother
\renewcommand{\thelemma}{\thetheorem}
\renewcommand{\theproposition}{\thetheorem}
\renewcommand{\thecorollary}{\thetheorem}
\renewcommand{\thedefinition}{\thetheorem}
\renewcommand{\theexample}{\thetheorem}
\renewcommand{\theremark}{\thetheorem}
\renewcommand{\theHlemma}{\theHtheorem}
\renewcommand{\theHproposition}{\theHtheorem}
\renewcommand{\theHcorollary}{\theHtheorem}
\renewcommand{\theHdefinition}{\theHtheorem}
\renewcommand{\theHexample}{\theHtheorem}
\renewcommand{\theHremark}{\theHtheorem}
\allowdisplaybreaks[2]
\setlist{itemsep=2pt,topsep=4pt,parsep=0pt}
\section*{Guide to the appendices}
\label{app:guide}
\addcontentsline{toc}{section}{Guide to the appendices}
Appendix~\ref{app:a} gives compatible laws, examples, and identification proofs;
Appendix~\ref{app:b} gives score bias, uniform estimation, calibration, and the
learning lower bound; Appendix~\ref{app:c} gives retained-information revelation,
its cost frontier, and computation. Appendix~\ref{app:d} records scientific
protocols and complete experimental results; Appendix~\ref{app:e} treats
sensitivity calibration, extensions, and related approaches. Appendix~\ref{app:f}
collects additional mechanism, learning, and revelation diagnostics. Implementation
provenance is available in the public repository described in
Appendix~\ref{app:implementation-inventory}.

Throughout, $\beta=1-\alpha$, $\Gamma<\infty$, and the target population and outcome
are fixed. $I_\Gamma(f)$ is identification width; $\widehat C_{n,\Gamma}(f)$ is a
statistical outer certificate. After revelation, $W_{f,\eta}(\lambda)$ is the gap
at a common threshold, whereas $I^{\mathrm{ret}}_{\Gamma,\lambda}(f)$ optimizes the
two endpoints separately. Auxiliary results and equations use appendix-local numbering.

\section{Population Identification and Sensitivity}
\label{app:a}

\subsection{Compatible full-data laws and sensitivity bounds}
\label{app:compatibility}\label{app:target-outcomes}

The target outcome and population are fixed across predictors and compatible laws. In a decision problem, $Y=Y(a^\star)$ is the outcome under a specified action (e.g., default under loan issuance), and $S$ records its availability \citep{lakkaraju2017selective,kleinberg2015prediction,kleinberg2018human}. Using an observed outcome requires consistency with that action and outcome definition. Comparing prediction rules holds $a^\star$ and $P_X$ fixed; changing the action, recovering an outcome under an untaken action, or transferring to a new population requires an additional causal or transport argument. The sensitivity model supplies neither exchangeability nor causal identification. Passive revelation recovers an already-realized target outcome, as in record linkage or controlled masking. Thus mature issued loans supply the masking target, whereas repayment under issuance remains a potential outcome for rejected applicants in the historical analysis.

Let $\mathcal X$ be standard Borel, and let $P_{\mathrm{obs}}$ be the law of $O=(X,S,Y^{\mathrm{obs}})$ with $Y^{\mathrm{obs}}=SY$. Use measurable versions
\[
\pi(x)=P_{\mathrm{obs}}(S=1\mid X=x),\qquad
p_1(x)=P_{\mathrm{obs}}(Y^{\mathrm{obs}}=1\mid X=x,S=1),
\]
and losses $z_f^y(x)=\ell(f(x),y)\in[0,1]$. Assumption~\ref{ass:regularity} gives $\pi\geq\pi_{\min}>0$, so $\pi\,dP_X$ and $P_X$ have the same null sets and $p_1$ is identified $P_X$-almost surely. All unqualified almost-sure statements use $P_X$. Write $H=\{\pi<1\}$. For a measurable $q:\mathcal X\to[0,1]$ and its Bernoulli kernel $K_q$, define
\begin{equation}
\begin{split}
P^q(dx,\{1\},dy)&=P_X(dx)\pi(x)K_{p_1}(x,dy),\\
P^q(dx,\{0\},dy)&=P_X(dx)[1-\pi(x)]K_q(x,dy).
\end{split}
\label{eq:a:law-construction}
\end{equation}
Set $q=p_1$ on $H^c$, where its value carries zero hidden mass.

\begin{lemma}[Kernel representation of compatible laws]
\label{lemma:a:kernels}
Every full-data law inducing $P_{\mathrm{obs}}$ has the form \eqref{eq:a:law-construction}, with $q(x)=P(Y=1\mid X=x,S=0)$ on $H$. Conversely, every measurable $q:\mathcal X\to[0,1]$ in \eqref{eq:a:law-construction} induces $P_{\mathrm{obs}}$. With $q=p_1$ on $H^c$, these kernels identify the compatible laws up to $P_X$-null sets.
\end{lemma}

\begin{proof}
Standard Borel disintegration fixes the $(X,S)$ law and selected-label kernel, leaving one binary probability on $S=0$. Conversely, the two displayed measures sum to one and have the prescribed pushforward under $(x,s,y)\mapsto(x,s,sy)$. Equal full-data laws have equal hidden kernels $(1-\pi)dP_X$-almost surely, equivalently $P_X$-almost surely on $H$; the canonical assignment settles $H^c$.
\end{proof}

For finite $\Gamma\geq1$, the odds restriction is equivalent, at interior $p_1$, to the bounds $\underline p_\Gamma=q_-(p_1;\Gamma)$ and $\overline p_\Gamma=q_+(p_1;\Gamma)$, where
\begin{equation}
q_-(p;\Gamma)=\frac{p}{p+\Gamma(1-p)},\qquad
q_+(p;\Gamma)=\frac{\Gamma p}{1-p+\Gamma p}.
\label{eq:a:sensitivity-maps}
\end{equation}
Their denominators are at least one. Thus $p_1=0$ forces $q=0$ and $p_1=1$ forces $q=1$ at every finite $\Gamma$; these substantive boundary restrictions merit scrutiny when apparent certainty comes from a fitted model. For interior $p_1$, the endpoints tend to zero and one as $\Gamma\to\infty$. Let $\mathcal Q_\Gamma$ contain measurable kernels within these bounds on $H$, canonically assigned on $H^c$. Then
\begin{equation}
\mathcal P_\Gamma(P_{\mathrm{obs}})=\{P^q:q\in\mathcal Q_\Gamma\}.
\label{eq:a:compatible-class}
\end{equation}
It is nonempty since $\underline p_\Gamma\leq p_1\leq\overline p_\Gamma$; at $\Gamma=1$ it is the single law satisfying $Y\perp S\mid X$.

For $\beta=1-\alpha\in(0,1)$, the CVaR representation \citep{rockafellar2002conditional} reduces to
\begin{equation}
\rho_{\alpha,P}(f)=\inf_{\eta\in\mathbb R}\{\eta+\beta^{-1}\mathbb E_P(L_f-\eta)_+\}
=\min_{\eta\in[0,1]}\{\eta+\beta^{-1}\mathbb E_P(L_f-\eta)_+\}.
\label{eq:a:ru}
\end{equation}
For $\eta<0$, the objective is $\beta^{-1}\mathbb E_PL_f+(1-\beta^{-1})\eta$ and exceeds its value at zero; for $\eta>1$, it equals $\eta>1$. Continuity on $[0,1]$ gives attainment, including distributions with atoms.

\subsection{Observational equivalence and motivating constructions}
\label{app:examples}

\begin{theorem}[Observational equivalence with reversed CVaR rankings]
\label{thm:main:reversal}
For every $\alpha\in(0,1)$, there exist an observed law satisfying strict overlap, two predictors $f,g$ with identical selected-loss distributions, and two full-data laws $P^\star,Q^\star$ inducing that observed law such that
\[
\rho_{\alpha,P^\star}(f)<\rho_{\alpha,P^\star}(g),
\qquad
\rho_{\alpha,Q^\star}(f)>\rho_{\alpha,Q^\star}(g).
\]
\end{theorem}

\begin{proof}
Take constant $X$, $0<\pi<\min\{1,2\beta\}$, $S\sim\operatorname{Bernoulli}(\pi)$, and $Y\mid S=1\sim\operatorname{Bernoulli}(1/2)$. Under zero-one loss, the constant predictors $f=0,g=1$ have the same selected-loss law. Set $Y=0$ on $S=0$ under $P^\star$ and $Y=1$ there under $Q^\star$. Both yield observed masses $\pi/2,\pi/2,1-\pi$ on $(S,Y^{\mathrm{obs}})=(1,1),(1,0),(0,0)$ and satisfy overlap with $\pi_{\min}=\pi$.

For $B\sim\operatorname{Bernoulli}(r)$, the threshold objective is $r/\beta+\eta(1-r/\beta)$ on $[0,1]$, hence
\begin{equation}
\operatorname{CVaR}_\alpha(B)=\min\{r/\beta,1\}.
\label{eq:a:bernoulli-cvar}
\end{equation}
Under $P^\star$, the loss probabilities are $\pi/2$ and $1-\pi/2$, so $\rho_{\alpha,P^\star}(f)=\pi/(2\beta)<\min\{(1-\pi/2)/\beta,1\}=\rho_{\alpha,P^\star}(g)$. Under $Q^\star$ they exchange.
\end{proof}
This construction concerns the unrestricted compatible class: its hidden probabilities zero and one lie outside the finite-$\Gamma$ interval $[1/(1+\Gamma),\Gamma/(1+\Gamma)]$ when $p_1=1/2$.

\paragraph{Figure~\ref{fig:framework}.}
For squared loss, $f=1/5$, $\pi=1/2$, $p_1=1/3$, $\Gamma=2$, and $\alpha=1/2$, the loss is $1/25+(3/5)B$ with $B\sim\operatorname{Bernoulli}(1/6+q/2)$ and $q\in[1/5,1/2]$. Translation and positive scaling of~\eqref{eq:a:ru}, followed by~\eqref{eq:a:bernoulli-cvar}, give
\begin{equation}
\begin{gathered}
\rho_{1/2,P^q}(f)=\frac1{25}+\frac65\left(\frac16+\frac q2\right),\\
\rho_{1/2,P^{1/5}}(f)=\frac9{25}=0.36,\quad
\rho_{1/2,P^{1/2}}(f)=\frac{27}{50}=0.54.
\end{gathered}
\label{eq:a:figure-one}
\end{equation}
These endpoints describe the risk of one fixed predictor under the same observed law.

\subsection{Conditional excess-loss envelopes and common extremal laws}
\label{app:extremal}

Write $a^y_{f,\eta}=(z_f^y-\eta)_+$, $d_{f,\eta}=a^1_{f,\eta}-a^0_{f,\eta}$, $d_+=(d_{f,\eta})_+$, and $d_-=(-d_{f,\eta})_+$. The hidden conditional excess loss is $a^0+qd$, with envelopes
\begin{equation}
\begin{split}
b^U_{f,\eta}&=a^0_{f,\eta}+\overline p_\Gamma d_+-\underline p_\Gamma d_-,\\
b^L_{f,\eta}&=a^0_{f,\eta}+\underline p_\Gamma d_+-\overline p_\Gamma d_-.
\end{split}
\label{eq:a:conditional-envelopes}
\end{equation}
The observable objectives are
\begin{equation}
\begin{split}
m^\sigma_{f,\eta}&=\mathbb E_{\mathrm{obs}}[S a^{Y^{\mathrm{obs}}}_{f,\eta}(X)+(1-S)b^\sigma_{f,\eta}(X)],\quad\sigma\in\{L,U\},\\
L_{f,\eta}&=\eta+\beta^{-1}m^L_{f,\eta},\qquad U_{f,\eta}=\eta+\beta^{-1}m^U_{f,\eta}.
\end{split}
\label{eq:a:population-objectives}
\end{equation}

\begin{lemma}[Common extremal kernels and threshold regularity]
\label{lemma:a:common-extrema}\label{lemma:a:threshold-regularity}
On $H$, define
\begin{equation}
\begin{split}
q_f^U(x)&=\overline p_\Gamma(x)\mathbf{1}\{z_f^1(x)\geq z_f^0(x)\}
 +\underline p_\Gamma(x)\mathbf{1}\{z_f^1(x)<z_f^0(x)\},\\
q_f^L(x)&=\underline p_\Gamma(x)\mathbf{1}\{z_f^1(x)\geq z_f^0(x)\}
 +\overline p_\Gamma(x)\mathbf{1}\{z_f^1(x)<z_f^0(x)\},
\end{split}
\label{eq:a:common-kernels}
\end{equation}
and set both kernels to $p_1$ on $H^c$. These kernels belong to $\mathcal Q_\Gamma$. For every $q\in\mathcal Q_\Gamma$ and every $\eta\in[0,1]$,
\begin{equation}
m^L_{f,\eta}
\leq\mathbb E_{P^q}[(L_f-\eta)_+]
\leq m^U_{f,\eta},
\qquad
\mathbb E_{P^{q_f^\sigma}}[(L_f-\eta)_+]=m^\sigma_{f,\eta}.
\label{eq:a:simultaneous-attainment}
\end{equation}
The objectives $L_{f,\eta},U_{f,\eta}$ are convex and $(1+\beta^{-1})$-Lipschitz in $\eta$, and attain their minima on $[0,1]$.
\end{lemma}

\begin{proof}
The kernels are measurable and satisfy the bounds. Since $z\mapsto(z-\eta)_+$ is nondecreasing, thresholding preserves the ordering of $z_f^1,z_f^0$; it can create a tie but cannot reverse the sign. Thus the same kernels extremize the affine expression $a^0+qd$ at every threshold, including ties. Disintegration gives
\begin{equation}
\begin{split}
\mathbb E_{P^q}(L_f-\eta)_+
={}&\mathbb E_{\mathrm{obs}}[S a^{Y^{\mathrm{obs}}}_{f,\eta}(X)]\\
&+\mathbb E_{P_X}[(1-\pi)(a^0_{f,\eta}+qd_{f,\eta})],
\end{split}
\label{eq:a:excess-disintegration}
\end{equation}
proving~\eqref{eq:a:simultaneous-attainment}, with zero contribution from $H^c$. Each endpoint objective is therefore the CVaR objective of one fixed law. Convexity and the Lipschitz bound follow from the convex, 1-Lipschitz hinge; compactness gives attainment.
\end{proof}

\subsection{Sharpness, endpoint attainment, and interval filling}
\label{app:sharpness}

\begin{proof}[Proof of Theorem~\ref{thm:cvar-sharp}]
Put $G_f(q,\eta)=\eta+\beta^{-1}\mathbb E_{P^q}(L_f-\eta)_+$. Lemma~\ref{lemma:a:common-extrema} gives $L_{f,\eta}=G_f(q_f^L,\eta)\leq G_f(q,\eta)\leq G_f(q_f^U,\eta)=U_{f,\eta}$ simultaneously for all thresholds. Minimizing these inequalities and choosing the respective common kernels proves
\begin{equation}
\underline\rho_{\alpha,\Gamma}(f)=\min_{\eta\in[0,1]}L_{f,\eta},\qquad
\overline\rho_{\alpha,\Gamma}(f)=\min_{\eta\in[0,1]}U_{f,\eta}.
\label{eq:a:sharp-endpoints}
\end{equation}
Both threshold minima and both optimized endpoints are attained. In particular, the upper equality uses one kernel attaining the supremum at every threshold, without an abstract minimax interchange.

For functions with finite infima on a common domain,
\begin{equation}
\left|\inf_\eta F(\eta)-\inf_\eta G(\eta)\right|
\leq\sup_\eta|F(\eta)-G(\eta)|,
\label{eq:a:infimum-stability}
\end{equation}
because each function is bounded below by the other minus their uniform distance. Equation~\eqref{eq:a:excess-disintegration} and $|d_{f,\eta}|\leq1$ therefore imply
\[
|\rho_{\alpha,P^q}(f)-\rho_{\alpha,P^{\widetilde q}}(f)|
\leq\sup_\eta|G_f(q,\eta)-G_f(\widetilde q,\eta)|
\leq\beta^{-1}\mathbb E[(1-\pi)|q-\widetilde q|].
\]
The mixture $q_t=(1-t)q_f^L+tq_f^U$, $t\in[0,1]$, remains measurable and compatible, and satisfies
\begin{equation}
|\rho_{\alpha,P^{q_t}}(f)-\rho_{\alpha,P^{q_s}}(f)|
\leq\frac{|t-s|}{\beta}\mathbb E[(1-\pi)|q_f^U-q_f^L|]
\leq\frac{|t-s|}{\beta}.
\label{eq:a:mixture-lipschitz}
\end{equation}
This continuous path joins the two attained endpoints. The intermediate value theorem, including the singleton case, yields
\begin{equation}
\{\rho_{\alpha,P}(f):P\in\mathcal P_\Gamma(P_{\mathrm{obs}})\}
=[\underline\rho_{\alpha,\Gamma}(f),\overline\rho_{\alpha,\Gamma}(f)].
\label{eq:a:sharp-interval}
\end{equation}
\end{proof}

\subsection{Sensitivity analysis and population decision criteria}
\label{app:sensitivity}\label{app:comparison-scope}

\begin{proposition}[Nested classes and continuous endpoint curves]
\label{prop:a:sensitivity}
If $1\leq\Gamma_1\leq\Gamma_2<\infty$, then
\[
\mathcal P_{\Gamma_1}(P_{\mathrm{obs}})\subseteq\mathcal P_{\Gamma_2}(P_{\mathrm{obs}}).
\]
For each $f$, the lower endpoint is nonincreasing in $\Gamma$, the upper endpoint is nondecreasing, and the identification width
$I_\Gamma(f)=\overline\rho_{\alpha,\Gamma}(f)-\underline\rho_{\alpha,\Gamma}(f)$ is nondecreasing. The interval is a singleton at $\Gamma=1$. Moreover, for $\sigma\in\{L,U\}$, writing $\rho^L_{\alpha,\Gamma}=\underline\rho_{\alpha,\Gamma}$ and $\rho^U_{\alpha,\Gamma}=\overline\rho_{\alpha,\Gamma}$,
\begin{equation}
|\rho^\sigma_{\alpha,\Gamma_2}(f)-\rho^\sigma_{\alpha,\Gamma_1}(f)|
\leq\frac{\mathbb E_{P_X}(1-\pi)}{4\beta}|\Gamma_2-\Gamma_1|.
\label{eq:a:gamma-lipschitz}
\end{equation}
In particular, both endpoint curves are continuous at every finite sensitivity level.
\end{proposition}

\begin{proof}
For $p\in[0,1]$, $\partial_\Gamma q_-=-p(1-p)/\{p+\Gamma(1-p)\}^2\in[-1/4,0]$ and $\partial_\Gamma q_+=p(1-p)/\{1-p+\Gamma p\}^2\in[0,1/4]$. Their signs prove nesting and endpoint monotonicity; at one both maps equal $p$. Their magnitude bounds and the $\Gamma$-independent loss ordering in~\eqref{eq:a:common-kernels} bound the difference of the extremal kernels by $|\Gamma_2-\Gamma_1|/4$. Apply~\eqref{eq:a:excess-disintegration} and~\eqref{eq:a:infimum-stability} to obtain~\eqref{eq:a:gamma-lipschitz}.
\end{proof}

\begin{corollary}[Population risk budgets, comparisons, and selection]
\label{cor:a:decisions}
For any risk budget $r$,
\begin{equation}
\rho_{\alpha,P}(f)\leq r\quad\hbox{for every }P\in\mathcal P_\Gamma(P_{\mathrm{obs}})
\quad\Longleftrightarrow\quad
\overline\rho_{\alpha,\Gamma}(f)\leq r.
\label{eq:a:risk-budget}
\end{equation}
For two predictors, the sufficient separation condition
\begin{equation}
\overline\rho_{\alpha,\Gamma}(f)<\underline\rho_{\alpha,\Gamma}(g)
\label{eq:a:dominance}
\end{equation}
implies $\rho_{\alpha,P}(f)<\rho_{\alpha,P}(g)$ under every compatible law. If $\overline\rho_{\alpha,\Gamma}(f_\varepsilon)\leq\inf_{g\in\mathcal F}\overline\rho_{\alpha,\Gamma}(g)+\varepsilon$, then every such law satisfies $\rho_{\alpha,P}(f_\varepsilon)\leq\inf_{g\in\mathcal F}\overline\rho_{\alpha,\Gamma}(g)+\varepsilon$. A nonempty finite class permits $\varepsilon=0$.
\end{corollary}

\begin{proof}
Use the attained upper endpoint for~\eqref{eq:a:risk-budget}, the chain $\rho_P(f)\leq\overline\rho(f)<\underline\rho(g)\leq\rho_P(g)$ for comparison, and $\rho_P(f_\varepsilon)\leq\overline\rho(f_\varepsilon)$ for selection.
\end{proof}
If $\underline\rho(f)>r$, every law violates the budget; if $\underline\rho(f)\leq r<\overline\rho(f)$, the two extremal laws respectively satisfy and violate it. Thus sharpness covers all budget and equality cases.

\paragraph{Common-law comparisons.}
Define $D_{f,g}(\Gamma)=\overline\rho_{\alpha,\Gamma}(f)-\underline\rho_{\alpha,\Gamma}(g)$. Marginal extrema can use different hidden kernels, whereas the direct contrast is
\begin{equation}
\Delta_\Gamma(f,g)=\sup_{P\in\mathcal P_\Gamma(P_{\mathrm{obs}})}
\{\rho_{\alpha,P}(f)-\rho_{\alpha,P}(g)\}\leq D_{f,g}(\Gamma).
\label{eq:e:common-law-comparison}
\end{equation}
Consequently, $D_{f,g}<0$ proves common-law dominance, but overlapping marginal intervals leave its sign unresolved by this criterion. Determining $\Delta_\Gamma$ requires joint optimization under one hidden kernel. The uniform event of Appendix~\ref{app:decision-guarantees} makes the marginal comparison and selection rules valid simultaneously over the stipulated predictor class using the same observations.

\paragraph{First breakdown of an interval-certified comparison.}
Set $\Gamma_{\mathrm{break}}(f,g)=\inf\{\Gamma\geq1:D_{f,g}(\Gamma)\geq0\}$, with $\inf\varnothing=\infty$. The gap is continuous and nondecreasing. If $D_{f,g}(1)\geq0$, breakdown is one and no strict interval certificate holds. If $D_{f,g}(1)<0$, breakdown exceeds one; when finite, continuity forces $D_{f,g}(\Gamma_{\mathrm{break}})=0$ and strict separation holds exactly on $[1,\Gamma_{\mathrm{break}})$. A nonzero gap at that boundary would persist across it, contradicting first failure. Infinite breakdown means separation at every finite level. Differences between two nondecreasing upper endpoints need not be monotone, so the upper-endpoint minimizer may change independently of these pairwise breakdown values.

\section{Statistical Certificates and Learning Complexity}
\label{app:b}

This appendix proves the statistical results for a fixed tail level
$\alpha\in(0,1)$ and a fixed finite sensitivity level $\Gamma\geq 1$.
Write $\beta=1-\alpha$, $p=p_1$, and $P=P_{\mathrm{obs}}$; thus $Pg$
denotes integration of an observed-data function $g$ against the common
observed law. For brevity, set
\[
 \rho^L(f)=\underline{\rho}_{\alpha,\Gamma}(f),\qquad
 \rho^U(f)=\overline{\rho}_{\alpha,\Gamma}(f),\qquad
 I_\Gamma(f)=\rho^U(f)-\rho^L(f).
\]
The population characterization in Appendix~\ref{app:a} gives
$\rho^\sigma(f)=\inf_{\eta\in[0,1]}
\{\eta+\beta^{-1}m^\sigma_{f,\eta}\}$ for
$\sigma\in\{L,U\}$. All $L_2$ norms below are with respect to $P_X$.

For uniform empirical-process statements, assume the paired maps
$x\mapsto(\ell(f(x),0),\ell(f(x),1))$, $f\in\mathcal F$, form a pointwise-separable
class of measurable functions with the envelope supplied by
Assumption~\ref{ass:regularity}. A countable pointwise-dense subclass and
rational thresholds including $0,1$ determine the suprema below, since the
score transformations are continuous. Measurable, pointwise-separable
versions of the fitted score classes suffice as well.

\subsection{Score regularity and exact nuisance-bias expansion}
\label{app:score-bias}

Recall $a^y_{f,\eta}(x)=(\ell(f(x),y)-\eta)_+$ and
$d_{f,\eta}(x)=a^1_{f,\eta}(x)-a^0_{f,\eta}(x)$. For a real number $d$,
write $d_+=\max(d,0)$ and $d_-=\max(-d,0)$. Define
\begin{align}
 H_p^U(d)&=q_+(p)d_+-q_-(p)d_-,&
 H_p^L(d)&=q_-(p)d_+-q_+(p)d_-, \label{eq:b:H}\\
 G_p^U(d)&=q_+'(p)d_+-q_-'(p)d_-,&
 G_p^L(d)&=q_-'(p)d_+-q_+'(p)d_- . \label{eq:b:G}
\end{align}
Consequently,
$b^\sigma_{f,\eta}(x;p)=a^0_{f,\eta}(x)+H_p^\sigma(d_{f,\eta}(x))$
and $\dot b^\sigma_{f,\eta}(x;p)=G_p^\sigma(d_{f,\eta}(x))$.

\begin{lemma}[Regularity of the endpoint maps]
\label{lemma:b:regularity}
For $p\in[0,1]$, the derivatives of the sensitivity endpoints satisfy
\begin{align}
 q_+'(p)&=\frac{\Gamma}{\{1+(\Gamma-1)p\}^2},&
 q_-'(p)&=\frac{\Gamma}{\{\Gamma-(\Gamma-1)p\}^2},\label{eq:b:first-derivatives}\\
 q_+''(p)&=-\frac{2\Gamma(\Gamma-1)}{\{1+(\Gamma-1)p\}^3},&
 q_-''(p)&=\frac{2\Gamma(\Gamma-1)}{\{\Gamma-(\Gamma-1)p\}^3}.
 \label{eq:b:second-derivatives}
\end{align}
In particular, $\Gamma^{-1}\leq q_\pm'(p)\leq\Gamma$ and
$|q_\pm''(p)|\leq 2\Gamma(\Gamma-1)$. Uniformly in $(f,\eta,x,p,\sigma)$,
\begin{equation}
 \begin{gathered}
 0\leq b^\sigma_{f,\eta}(x;p)\leq1,\qquad
 |\dot b^\sigma_{f,\eta}(x;p)|\leq\Gamma,\\
 |\partial_p^2 b^\sigma_{f,\eta}(x;p)|\leq2\Gamma(\Gamma-1).
 \end{gathered}
 \label{eq:b:regularity-bounds}
\end{equation}
Moreover, $H_p^\sigma$ is $1$-Lipschitz and $G_p^\sigma$ is
$\Gamma$-Lipschitz on the real line; both vanish at zero.
\end{lemma}

\begin{proof}
Differentiating $q_+(p)=\Gamma p/\{1+(\Gamma-1)p\}$ and
$q_-(p)=p/\{\Gamma-(\Gamma-1)p\}$ gives
\eqref{eq:b:first-derivatives}--\eqref{eq:b:second-derivatives}.
The denominators before taking powers lie in $[1,\Gamma]$, giving the
derivative bounds. The value $a^0+H_p^\sigma(a^1-a^0)$ is a convex
combination of $a^0,a^1\in[0,1]$; its derivatives are bounded by
$\Gamma|a^1-a^0|$ and $2\Gamma(\Gamma-1)|a^1-a^0|$, respectively.
This proves \eqref{eq:b:regularity-bounds}. On either half-line the slopes
of $H_p^\sigma$ and $G_p^\sigma$ lie in $[0,1]$ and
$[\Gamma^{-1},\Gamma]$, respectively. Both maps are continuous at zero,
so these slope bounds also prove the Lipschitz bounds across loss ties.
\end{proof}

For candidate nuisance functions $p_c:\mathcal X\to[0,1]$ and
$\pi_c:\mathcal X\to[\pi_{\min}/2,1]$, the score is
\begin{align}
 \phi^\sigma_{f,\eta}(O;p_c,\pi_c)
 ={}&S a^{Y^{\mathrm{obs}}}_{f,\eta}(X)
 +(1-S)b^\sigma_{f,\eta}(X;p_c(X)) \notag\\
 &+S\frac{1-\pi_c(X)}{\pi_c(X)}
 \dot b^\sigma_{f,\eta}(X;p_c(X))
 \{Y^{\mathrm{obs}}-p_c(X)\}.
 \label{eq:b:score}
\end{align}
The sum of its first two terms belongs to $[0,1]$, whereas its correction
has absolute value at most $2\Gamma/\pi_{\min}$. Therefore
\begin{equation}
 \begin{split}
 -\frac{2\Gamma}{\pi_{\min}}&\leq
 \phi^\sigma_{f,\eta}(O;p_c,\pi_c)
 \leq1+\frac{2\Gamma}{\pi_{\min}},\\
 |\phi^\sigma_{f,\eta}(O;p_c,\pi_c)|&\leq
 B:=1+\frac{2\Gamma}{\pi_{\min}}.
 \end{split}
 \label{eq:b:score-envelope}
\end{equation}
The score can be negative; the concentration argument below explicitly
allows signed classes.

\begin{lemma}[Exact conditional nuisance bias]
\label{lemma:b:bias}
Suppress $(f,\eta,\sigma,X)$ inside $b$ and its derivatives. With
$e=p_c-p$, the conditional difference between the score expectation and
the target excess-loss integrand is exactly
\begin{equation}
 (1-\pi)\{b(p_c)-b(p)-\dot b(p_c)e\}
 +\frac{\pi_c-\pi}{\pi_c}\dot b(p_c)e.
 \label{eq:b:conditional-bias}
\end{equation}
Taking $P_X$ expectations of \eqref{eq:b:conditional-bias} gives
$P\phi^\sigma_{f,\eta}(\cdot;p_c,\pi_c)-m^\sigma_{f,\eta}$, and
\begin{equation}
 \begin{aligned}
 &\left|P\phi^\sigma_{f,\eta}(\cdot;p_c,\pi_c)-m^\sigma_{f,\eta}\right|\\
 &\quad\leq\Gamma(\Gamma-1)\|p_c-p\|_2^2
 +\frac{2\Gamma}{\pi_{\min}}\|p_c-p\|_2\|\pi_c-\pi\|_2.
 \end{aligned}
 \label{eq:b:bias-bound}
\end{equation}
\end{lemma}

\begin{proof}
Conditional on $X$, the observed excess-loss contribution has mean
$\pi\{(1-p)a^0+pa^1\}$ and
$\mathbb E[S(Y^{\mathrm{obs}}-p_c)\mid X]=\pi(p-p_c)$.
Subtracting the target integrand
$\pi\{(1-p)a^0+pa^1\}+(1-\pi)b(p)$ leaves
\[
 (1-\pi)\{b(p_c)-b(p)\}
 +\frac{\pi(1-\pi_c)}{\pi_c}\dot b(p_c)(p-p_c).
\]
The identity
$\pi(1-\pi_c)/\pi_c=(1-\pi)+(\pi-\pi_c)/\pi_c$
reduces this expression to \eqref{eq:b:conditional-bias}.
Taylor's theorem, expanded about $p_c$ along the segment joining $p_c$
and $p$, and Lemma~\ref{lemma:b:regularity} give
\[
 |b(p_c)-b(p)-\dot b(p_c)(p_c-p)|
 \leq\Gamma(\Gamma-1)(p_c-p)^2.
\]
Use $0\leq1-\pi\leq1$ for the first term, and
$|\dot b(p_c)/\pi_c|\leq2\Gamma/\pi_{\min}$ followed by
Cauchy--Schwarz for the second term. This yields \eqref{eq:b:bias-bound}
uniformly over all score indices.
\end{proof}

The score is unbiased at the true nuisances and has $O(t^2)$ bias along
admissible nuisance paths with $L_2$ perturbations of order $t$.
The sign of $d_{f,\eta}(x)$ is independent of the nuisance argument, so
endpoint selection adds no remainder. At $\Gamma=1$ the curvature term
vanishes; for $\Gamma>1$, knowing $\pi$ alone generally leaves quadratic
error in $p$. Thus the result is an orthogonal expansion, not generic
double robustness. Its overlap premise and clipping ranges are substantive:
clipping cannot supply labels or identification where overlap fails.

\subsection{Cross-fitting and uniform empirical-process control}
\label{app:empirical-process}

Let $I_1,\ldots,I_K$ be a deterministic partition of the observation indices,
with $n_k=|I_k|\geq1$, $n_{\min}=\min_k n_k$, and $w_k=n_k/n$.
A random partition independent of the observations is also allowed by
conditioning on that partition. Write $D_{-k}$ for the training observations
outside $I_k$ and $P_{n,k}$ for the empirical measure on $I_k$.
The clipped fits $\widehat p^{(-k)},\widehat\pi^{(-k)}$ are measurable
functions of $D_{-k}$. Any independent fitting randomization can be included
in this conditioning. Define
\[
 \phi^{\sigma,(k)}_{f,\eta}
 =\phi^\sigma_{f,\eta}
 (\cdot;\widehat p^{(-k)},\widehat\pi^{(-k)}),\qquad
 \Phi_k^\sigma=\{\phi^{\sigma,(k)}_{f,\eta}:f\in\mathcal F,
 \ \eta\in[0,1]\}.
\]
The empirical Rademacher complexities used here are
\begin{equation}
 \begin{split}
 \widehat{\mathfrak R}^{\sigma,k}_{\mathrm{abs}}
 &=\mathbb E_\varepsilon\!\left[
 \sup_{f,\eta}\left|\frac1{n_k}
 \sum_{i\in I_k}\varepsilon_i\phi^{\sigma,(k)}_{f,\eta}(O_i)\right|
 \,\middle|\,D_n\right],\\
 \widehat{\mathfrak R}_{n,\mathrm{abs}}^\sigma
 &=\sum_{k=1}^K w_k\widehat{\mathfrak R}^{\sigma,k}_{\mathrm{abs}}.
 \end{split}
 \label{eq:b:empirical-rademacher}
\end{equation}
Here $D_n$ includes the observations, partition, and fitting seeds.
The expectation averages only independent uniform signs in $\{-1,1\}$;
the fitted functions and evaluation observations are fixed.

\begin{lemma}[Empirical-complexity bound for a signed class]
\label{lemma:b:empirical-rademacher}
Let $Z_1,\ldots,Z_m$ be independent with law $Q$ and empirical measure $Q_m$,
and let $\mathcal H$ be a fixed pointwise-separable class with $|h|\leq B$. For $0<\xi<1$,
with probability at least $1-\xi$,
\begin{equation}
 \sup_{h\in\mathcal H}|(Q_m-Q)h|
 \leq2\widehat{\mathfrak R}_{m,\mathrm{abs}}(\mathcal H)
 +6B\sqrt{\frac{\log(2/\xi)}{2m}},
 \label{eq:b:signed-concentration}
\end{equation}
where the empirical complexity uses the absolute supremum as in
\eqref{eq:b:empirical-rademacher}.
\end{lemma}

\begin{proof}
Put $Z=\sup_h|(Q_m-Q)h|$ and
$R=\widehat{\mathfrak R}_{m,\mathrm{abs}}(\mathcal H)$.
Symmetrization gives $\mathbb EZ\leq2\mathbb ER$
\cite{bartlett2002rademacher,van1996weak}. Changing one observation changes
each of $Z,R$ by at most $2B/m$. With
$t=2B\sqrt{\log(2/\xi)/(2m)}$, McDiarmid's inequality bounds each of
$\mathbb P(Z>\mathbb EZ+t)$ and $\mathbb P(\mathbb ER>R+t)$ by $\xi/2$.
Thus $Z\leq2R+3t$ with probability at least $1-\xi$, proving
\eqref{eq:b:signed-concentration} for signed classes with $|h|\leq B$.
\end{proof}

Conditional on $D_{-k}$, the observations in $I_k$ are independent with
law $P$, and $\Phi_k^\sigma$ is fixed with envelope $B$.
Apply Lemma~\ref{lemma:b:empirical-rademacher} with
$m=n_k$ and $\xi=\delta/(2K)$. Its conditional failure probability is
at most $\delta/(2K)$, so the same bound holds unconditionally.
A union bound over the $2K$ fold--endpoint pairs yields an event
$\mathcal E_{\mathrm{emp}}$ of probability at least $1-\delta$ on which
\begin{align}
 \sup_{f,\eta}\left|
 \sum_{k=1}^K w_k(P_{n,k}-P)\phi^{\sigma,(k)}_{f,\eta}
 \right|
 &\leq2\widehat{\mathfrak R}_{n,\mathrm{abs}}^\sigma
 +6B\sum_{k=1}^K w_k\sqrt{\frac{\log(4K/\delta)}{2n_k}}\notag\\
 &\leq2\widehat{\mathfrak R}_{n,\mathrm{abs}}^\sigma
 +6B\sqrt{\frac{\log(4K/\delta)}{2n_{\min}}}
 \label{eq:b:fold-bound}
\end{align}
for both endpoints. Here $P\phi^{\sigma,(k)}_{f,\eta}$ integrates the
fitted function against a fresh observation. The training sets overlap;
no independence between folds or between their success events is used.
In particular, the proof never conditions simultaneously on all fitted
nuisances, which could destroy the needed held-out independence.

\subsection{Uniform endpoint estimation and complexity rates}
\label{app:uniform-estimation}

Define the cross-fitted empirical objectives and their infima by
\begin{equation}
 \begin{aligned}
 \widehat Q_f^\sigma(\eta)&=\eta+\frac1\beta
 \sum_{k=1}^K w_kP_{n,k}\phi^{\sigma,(k)}_{f,\eta},\\
 \widehat\rho^\sigma(f)&=\inf_{\eta\in[0,1]}\widehat Q_f^\sigma(\eta).
 \end{aligned}
 \label{eq:b:empirical-objectives}
\end{equation}

\begin{proof}[Proof of Theorem~\ref{thm:learning}]
Under the theorem's assumptions and the measurability convention above,
let $\mathcal E_\nu$ be the simultaneous nuisance event
\begin{equation}
 \max_k\|\widehat p^{(-k)}-p\|_2\leq r_p,\qquad
 \max_k\|\widehat\pi^{(-k)}-\pi\|_2\leq r_\pi,
 \label{eq:b:nuisance-event}
\end{equation}
with probability at least $1-\delta_\nu$. For
$\delta\in(0,1)$ and $\delta_\nu\in[0,1-\delta)$, the
explicit calibrated radius is
\begin{equation}
 \begin{aligned}
 r_n(\delta)=\frac1\beta\Bigg[
 &2\max_{\sigma\in\{L,U\}}\widehat{\mathfrak R}_{n,\mathrm{abs}}^\sigma
 +6B\sqrt{\frac{\log(4K/\delta)}{2n_{\min}}}\\
 &+\Gamma(\Gamma-1)r_p^2
 +\frac{2\Gamma}{\pi_{\min}}r_pr_\pi\Bigg].
 \end{aligned}
 \label{eq:b:radius}
\end{equation}
On $\mathcal E_\nu$, Lemma~\ref{lemma:b:bias}, applied to each realized
pair of clipped fits, implies
\[
 \sup_{f,\eta,\sigma}\left|
 \sum_{k=1}^K w_kP\phi^{\sigma,(k)}_{f,\eta}
 -m^\sigma_{f,\eta}\right|
 \leq\Gamma(\Gamma-1)r_p^2+\frac{2\Gamma}{\pi_{\min}}r_pr_\pi.
\]
The intersection $\mathcal E=\mathcal E_\nu\cap\mathcal E_{\mathrm{emp}}$
has probability at least $1-\delta-\delta_\nu$ by the union bound.
On this event, \eqref{eq:b:fold-bound} and the last display give
\[
 \sup_{f,\eta,\sigma}\left|
 \sum_{k=1}^K w_kP_{n,k}\phi^{\sigma,(k)}_{f,\eta}
 -m^\sigma_{f,\eta}\right|\leq\beta r_n(\delta).
\]
Applying the infimum stability inequality \eqref{eq:a:infimum-stability}
to $\widehat Q_f^\sigma$ and its population counterpart yields, on the same
event of probability at least $1-\delta-\delta_\nu$,
\begin{equation}
 \sup_{f\in\mathcal F}\max_{\sigma\in\{L,U\}}
 |\widehat\rho^\sigma(f)-\rho^\sigma(f)|\leq r_n(\delta).
 \label{eq:b:uniform-endpoints}
\end{equation}
This requires neither convexity of the signed empirical objective nor a
unique threshold or loss density there.
\end{proof}

For a total confidence level $1-\delta_{\mathrm{tot}}$, allocate
$\delta+\delta_\nu\leq\delta_{\mathrm{tot}}$. The empirical complexity in
\eqref{eq:b:radius} is an expectation over all Rademacher signs, conditional
on the realized data. A Monte Carlo estimate is not automatically an upper
bound for this expectation. Any random complexity or nuisance-error bound
used instead must hold simultaneously on a specified validity event, whose
failure probability is included in the confidence allocation.

We next bound the score complexity. For a sample $x_1,\ldots,x_m$,
let $\widehat{\mathfrak R}_{m,\mathrm{abs}}(\mathcal V)$ denote its
absolute empirical complexity. Define
\[
 \mathcal L_y=\{x\mapsto\ell(f(x),y):f\in\mathcal F\},\qquad
 \mathcal A_y=\{x\mapsto(\ell(f(x),y)-\eta)_+:
 f\in\mathcal F,\ \eta\in[0,1]\}.
\]
Define the threshold-augmented loss-class complexity by
\begin{equation}
 \mathfrak C_n=\sum_{k=1}^K w_k\left\{
 \widehat{\mathfrak R}_{n_k,\mathrm{abs}}(\mathcal L_0)
 +\widehat{\mathfrak R}_{n_k,\mathrm{abs}}(\mathcal L_1)
 +n_k^{-1/2}\right\},
 \label{eq:b:loss-complexity}
\end{equation}
where each term uses the covariates in its evaluation fold.
For fixed fitted nuisances and $o=(x,s,y^{\mathrm{obs}})$, put
$p_c=\widehat p^{(-k)}(x)$ and $w=(1-\widehat\pi^{(-k)}(x)) /
\widehat\pi^{(-k)}(x)$. The score is
$\Psi_o^\sigma(a^0_{f,\eta}(x),a^1_{f,\eta}(x))$, where
\begin{align}
 \Psi_o^\sigma(u_0,u_1)
 ={}&s\{(1-y^{\mathrm{obs}})u_0+y^{\mathrm{obs}}u_1\}\notag\\
 &+(1-s)\{u_0+H_{p_c}^\sigma(u_1-u_0)\}\notag\\
 &+swG_{p_c}^\sigma(u_1-u_0)(y^{\mathrm{obs}}-p_c).
 \label{eq:b:composition}
\end{align}
It vanishes at $(0,0)$. Lemma~\ref{lemma:b:regularity},
$w\leq2/\pi_{\min}$, and
$|\Delta u_1-\Delta u_0|\leq2\|\Delta u\|_\infty$ show that it is
$L_*=3+4\Gamma/\pi_{\min}$-Lipschitz in the maximum norm,
and hence also in Euclidean norm.

The vector-contraction inequality \cite{maurer2016vector} therefore gives,
for a universal constant $C$,
\begin{equation}
 \begin{aligned}
 \widehat{\mathfrak R}^{\sigma,k}_{\mathrm{abs}}
 \leq C L_*\Big\{&
 \widehat{\mathfrak R}_{n_k,\mathrm{abs}}(\mathcal L_0)
 +\widehat{\mathfrak R}_{n_k,\mathrm{abs}}(\mathcal L_1)\\
 &+n_k^{-1/2}\Big\}.
 \end{aligned}
 \label{eq:b:score-complexity}
\end{equation}
To handle the absolute supremum, use that the score class contains zero
at $\eta=1$: the absolute supremum is at most the sum of the positive and
negative suprema, whose expectations agree by sign symmetry. Vector
contraction on the positive supremum, followed by separating the two
coordinate suprema, bounds it by a universal multiple of
$L_*\sum_y\widehat{\mathfrak R}_{n_k,\mathrm{abs}}(\mathcal A_y)$.
Apply the same argument to the $\sqrt2$-Lipschitz hinge map
$(z,t)\mapsto(z-t)_+$, adjoining zero to its input class if needed.
Its coordinate complexities are those of $\mathcal L_y$ and the constant
threshold class; the latter is
$m^{-1}\mathbb E_\varepsilon|\sum_i\varepsilon_i|\leq m^{-1/2}$.
This proves \eqref{eq:b:score-complexity}. All contraction maps are
continuous endpoint values or derivatives, including at ties; no
discontinuous selector class is introduced.

If $\mathcal L_y$ has VC-subgraph dimension at most $d\geq1$, then
$\log N(\epsilon,\mathcal L_y,L_2(Q))\leq C_0d\log(C_1/\epsilon)$
for $0<\epsilon<1$ and every probability measure $Q$.
After adjoining zero and the negative class, the standard entropy-to-complexity
bound \cite{van1996weak,bartlett2002rademacher} gives, uniformly over samples,
\begin{equation}
 \widehat{\mathfrak R}_{m,\mathrm{abs}}(\mathcal L_y)
 \leq C'\sqrt{\frac{d+1}{m}}.
 \label{eq:b:VC-complexity}
\end{equation}
It therefore applies to the fixed fitted nuisances in each conditional argument.

For fixed $K$ and balanced folds, $n_{\min}\geq c_Kn$ for a positive
constant $c_K$ once all folds are nonempty. Equations
\eqref{eq:b:score-complexity}--\eqref{eq:b:VC-complexity} give
\begin{equation}
 \begin{aligned}
 &\frac1\beta\left[
 2\max_\sigma\widehat{\mathfrak R}_{n,\mathrm{abs}}^\sigma
 +6B\sqrt{\frac{\log(4K/\delta)}{2n_{\min}}}\right]\\
 &\qquad\leq \frac{C_{\Gamma,\pi_{\min},K}}{\beta}
 \sqrt{\frac{d+\log(1/\delta)}{n}}.
 \end{aligned}
 \label{eq:b:VC-rate}
\end{equation}
For fixed $\Gamma,\pi_{\min},\beta$, nuisance rates $o(n^{-1/4})$ make
the nuisance contribution in \eqref{eq:b:radius} $o(n^{-1/2})$.
Joint high-probability nuisance bounds remain necessary; cross-fitting
does not establish them. All uniformity is over $f,\eta$ at fixed
$\alpha,\Gamma$.

\begin{example}[Conditional sparse-nuisance specialization]
\label{ex:b:sparse-nuisances}
Suppose sparse logistic nuisance fits in $D$ covariates satisfy simultaneous
$L_2(P_X)$ error bounds of orders
\[
 r_p\lesssim\sqrt{\frac{s_p\log(eD)}{n_{\mathrm{sel}}}},\qquad
 r_\pi\lesssim\sqrt{\frac{s_\pi\log(eD)}{n_{\mathrm{tr}}}},
\]
where $s_p,s_\pi$ are sparsities, $n_{\mathrm{tr}}$ is the smallest training-fold
size, and $n_{\mathrm{sel}}>0$ is a simultaneous lower bound for the number of
revealed training labels in every fold. Substitution in the proved remainder
gives, up to the constants in these assumed rates,
\[
 \frac1\beta\left\{
 \Gamma(\Gamma-1)\frac{s_p\log(eD)}{n_{\mathrm{sel}}}
 +\frac{2\Gamma}{\pi_{\min}}
 \frac{\sqrt{s_ps_\pi}\log(eD)}{\sqrt{n_{\mathrm{sel}}n_{\mathrm{tr}}}}
 \right\}.
\]
For a fixed number of balanced folds with nonempty training sets,
$n_{\mathrm{tr}}\asymp n$. These substitutions
require the stated model, design, tuning, and confidence conditions.
A selected-population bound transfers to $P_X$ because, for every measurable $e$,
\begin{equation}
 \|e\|_{L_2(P_X)}^2
 \leq\frac{P(S=1)}{\pi_{\min}}
 \|e\|_{L_2(P_{X\mid S=1})}^2.
 \label{eq:b:selected-norm-transfer}
\end{equation}
Indeed, $\mathbb E[\pi(X)e(X)^2]
=P(S=1)\mathbb E[e(X)^2\mid S=1]$ and $\pi(X)\geq\pi_{\min}$.
Learning $p$ from revealed outcomes must therefore control target-population error.
\end{example}

\begin{proposition}[Finite threshold search for the fitted objectives]
\label{prop:b:threshold-search}
For a fixed $f$ and realized nuisance fits, each objective in
\eqref{eq:b:empirical-objectives} attains its minimum on
\[
 T_n(f)=\{0,1\}\cup
 \{\ell(f(X_i),y):1\leq i\leq n,\ y\in\{0,1\}\}.
\]
\end{proposition}

\begin{proof}
Thresholding preserves the label-loss ordering at each $X_i$.
Thus $H_{p_c}^\sigma(d_{f,\eta})$ and $G_{p_c}^\sigma(d_{f,\eta})$
are fixed coefficients times $d_{f,\eta}$ throughout $[0,1]$; at
$d_{f,\eta}=0$ both representations vanish. Each score is a fixed linear
combination of the two label-specific hinges, making the objective continuous
and affine between successive knots of $T_n(f)$. Each such interval has
an endpoint minimizer, even if the objective is constant there.
Signed correction coefficients may prevent convexity but preserve this result.
\end{proof}

An exact implementation can sort the at most $2n+2$ breakpoints once
per predictor. Write the fitted score as
$c_{i0}^\sigma(\ell(f(X_i),0)-\eta)_+
+c_{i1}^\sigma(\ell(f(X_i),1)-\eta)_+$, with coefficients fixed across
$\eta$. Starting from the objective at zero, sweep the sorted breakpoints:
the objective is updated by its current slope times the distance to the
next breakpoint, and the slope increases by $c_{iy}^\sigma/(n\beta)$
when the corresponding hinge vanishes. Combine all coefficients at a
tied breakpoint and initialize the right slope using only losses above
zero. Checking all knot values gives both global minima in
$O(n\log n)$ time and $O(n)$ storage, after the fitted probabilities
and label-specific losses have been evaluated.

\subsection{Outer certificates, comparisons, and predictor selection}
\label{app:decision-guarantees}

\begin{proof}[Proof of the decision guarantees in Section~\ref{sec:uniform-certificates}]
Work on the joint event \eqref{eq:b:uniform-endpoints}, of probability at
least $1-\delta-\delta_\nu$, and use the certificate
\eqref{eq:outer-certificate}. Since $0\leq\rho^L(f)\leq\rho^U(f)\leq1$,
clipping preserves containment
of the sharp interval and hence every compatible CVaR. Containment gives
the lower width bound; clipping decreases width, while
\[
 \widehat\rho^U(f)-\widehat\rho^L(f)+2r_n
 \leq\rho^U(f)-\rho^L(f)+4r_n,
\]
proving the upper bound. If
$\widehat\rho^U(f)+r_n<\widehat\rho^L(g)-r_n$, then for every compatible law,
\begin{align*}
 \rho_{\alpha,P^\star}(f)&\leq\rho^U(f)
 \leq\widehat\rho^U(f)+r_n\\
 &<\widehat\rho^L(g)-r_n\leq\rho^L(g)
 \leq\rho_{\alpha,P^\star}(g).
\end{align*}
Finally, any measurable $\varepsilon_{\mathrm{opt}}$-approximate minimizer
$\widehat f\in\mathcal F$ of $\widehat\rho^U$ satisfies
\begin{align*}
 \rho^U(\widehat f)&\leq\inf_f\widehat\rho^U(f)+\varepsilon_{\mathrm{opt}}+r_n\\
 &\leq\inf_f\rho^U(f)+\varepsilon_{\mathrm{opt}}+2r_n.
\end{align*}
The last inequality takes infima of the pointwise endpoint bound, so no
exact minimizer is needed. Uniformity permits all predictors in these
claims to be chosen using the same data.
\end{proof}

The selection benchmark is the smallest compatible upper endpoint in
the class. For every compatible $P^\star$, the same argument gives
$\rho_{\alpha,P^\star}(\widehat f)\leq
\inf_f\rho^U(f)+2r_n+\varepsilon_{\mathrm{opt}}$.
Clipped endpoint errors are at most $2r_n$ on the same event. Convergence
in probability follows if $r_n\to0$ in probability and
$\delta_n+\delta_{\nu,n}\to0$; a fixed failure probability remains fixed.

\subsection{Oracle calibration for finite-cell experiments}
\label{app:oracle-calibration}

For the oracle coverage experiment, suppose $p=p_1$ is known and define
\begin{equation}
 \begin{split}
 h^\sigma_{f,\eta}(O)&=S a^{Y^{\mathrm{obs}}}_{f,\eta}(X)
 +(1-S)b^\sigma_{f,\eta}(X;p(X)),\\
 \widehat\rho^\sigma_{\mathrm{or}}(f)&=
 \inf_{\eta\in[0,1]}\{\eta+\beta^{-1}P_nh^\sigma_{f,\eta}\}.
 \end{split}
 \label{eq:b:oracle-score}
\end{equation}
This score lies in $[0,1]$ and has mean $m^\sigma_{f,\eta}$; estimating
$\pi$ is unnecessary for this oracle construction.

\begin{proposition}[Finite-cell oracle certificate]
\label{prop:b:oracle-calibration}
Fix a predictor $f$ independently of the evaluation observations.
Suppose $P_X$ is supported on $J$ covariate values $x_1,\ldots,x_J$.
Let
\[
 T(f)=\{0,1\}\cup\{\ell(f(x_j),y):1\leq j\leq J,
 \ y\in\{0,1\}\},\qquad M=|T(f)|\leq2J+2.
\]
With probability at least $1-\delta$,
\begin{equation}
 \max_{\sigma\in\{L,U\}}
 |\widehat\rho^\sigma_{\mathrm{or}}(f)-\rho^\sigma(f)|
 \leq r_{\mathrm{or}}(\delta):=
 \frac1\beta\sqrt{\frac{\log(4M/\delta)}{2n}}.
 \label{eq:b:oracle-radius}
\end{equation}
Thus the clipped interval obtained from \eqref{eq:outer-certificate}
using these oracle estimators and $r_{\mathrm{or}}$ contains the entire
sharp population interval with probability at least $1-\delta$.
\end{proposition}

\begin{proof}
By the piecewise-affine argument of Proposition~\ref{prop:b:threshold-search},
both $h^\sigma_{f,\eta}(O)$ and its expectation have knots only in $T(f)$.
The absolute empirical-minus-population difference is maximal at a knot:
\[
 \sup_{\eta\in[0,1]}|(P_n-P)h^\sigma_{f,\eta}|
 =\max_{\eta\in T(f)}|(P_n-P)h^\sigma_{f,\eta}|.
\]
Hoeffding's inequality and a union bound over the $2M$ endpoint--threshold
pairs bound the failure probability at tolerance $u$ by $4Me^{-2nu^2}$.
Taking $u=\sqrt{\log(4M/\delta)/(2n)}$ and applying
\eqref{eq:a:infimum-stability} proves \eqref{eq:b:oracle-radius};
The containment argument in Appendix~\ref{app:decision-guarantees} gives
whole-interval coverage.
\end{proof}

The population-support set $T(f)$ includes losses at cells absent from a
particular sample; observed knots $T_n(f)$ suffice for empirical minimization
but cannot replace $T(f)$ in this concentration argument.
Conditioning covers a predictor trained on independent data. Selection using
evaluation data requires a uniform bound: for a fixed finite family
$\mathcal F_0$, replace $M$ by $\sum_{f\in\mathcal F_0}|T(f)|$.
Known $p$ supplies the oracle experiment's premise; coverage for learned-nuisance
estimators instead requires Theorem~\ref{thm:learning}'s controls.

\subsection{Revealed-tail learning lower bound}
\label{app:learning-lower-bound}

We now examine the difficulty of learning a predictor in the \(\Gamma=1\) submodel, where population risks are identified. The lower-bound construction uses equal tail-cell masses and constant observation probability.

\begin{theorem}[Revealed-tail observation complexity]
\label{thm:main:lower-bound}
Fix $\alpha\in(0,1)$, $\beta=1-\alpha$, $\pi_0\in(0,1]$, and $n\geq1$.
For every integer $1\leq d\leq n\pi_0\beta$, there exist a binary
classification class $\mathcal F$ of VC dimension $d$ and a family
$\mathcal P$ of full-data laws with $\Gamma=1$ and $\pi(X)=\pi_0$ such that
\begin{equation}
 \begin{aligned}
 &\inf_{\widehat f}\sup_{P^\star\in\mathcal P}
 \mathbb E_{(P^\star_{\mathrm{obs}})^{\otimes n}}
 \left[\rho_{\alpha,P^\star}(\widehat f)
       -\inf_{f\in\mathcal F}\rho_{\alpha,P^\star}(f)\right]\\
 &\qquad\geq\frac1{16}\sqrt{\frac{d}{n\pi_0\beta}}.
 \end{aligned}
 \label{eq:b:learning-lower-bound}
\end{equation}
The infimum ranges over measurable learners taking values in $\mathcal F$,
including randomized learners.
\end{theorem}

\begin{proof}
We use a direct binary-testing form of the hypercube argument
\cite{yu1997assouad}. Fix
$\pi_0\in(0,1]$, $n\geq1$, and an integer
$1\leq d\leq n\pi_0\beta$.
Put
\begin{equation}
 \gamma=\frac18\sqrt{\frac{d}{n\pi_0\beta}}\leq\frac18.
 \label{eq:b:lower-gamma}
\end{equation}
Let $\mathcal X=\{\star,1,\ldots,d\}$ with
$P_X(\star)=1-\beta$ and $P_X(j)=\beta/d$.
For $v\in\{-1,1\}^d$, define
$f_v(\star)=0$ and $f_v(j)=(1+v_j)/2$, and set
$\mathcal F=\{f_v:v\in\{-1,1\}^d\}$.
The class shatters $\{1,\ldots,d\}$, and cannot shatter any larger set
because its value at $\star$ is fixed. Its VC dimension is exactly $d$.
Use ordinary zero-one loss $\ell(f(x),y)=\mathbf1\{f(x)\ne y\}$.

For each $\theta\in\{-1,1\}^d$, construct a full-data law $P_\theta$
with $Y=0$ at $\star$ and
\[
 P_\theta(Y=1\mid X=j)=\frac12+\gamma\theta_j,
 \qquad j=1,\ldots,d.
\]
Let $S\sim\operatorname{Bernoulli}(\pi_0)$ independently of $(X,Y)$,
and observe $(X,S,SY)$. These laws satisfy overlap with
$\pi(X)=\pi_0$ and conditional ignorability, hence belong to the
$\Gamma=1$ model.
All losses vanish at $\star$. The conditional error within
$\{1,\ldots,d\}$ is
\begin{equation}
 R_\theta(f_v)=\frac12-\gamma+
 \frac{2\gamma}{d}\sum_{j=1}^d\mathbf1\{v_j\ne\theta_j\}.
 \label{eq:b:hypercube-risk}
\end{equation}
The full-population loss is Bernoulli with parameter
$q_{\theta,v}=\beta R_\theta(f_v)\leq\beta$; the Bernoulli identity
\eqref{eq:a:bernoulli-cvar} gives
\begin{equation}
 \begin{split}
 \rho_{\alpha,P_\theta}(f_v)&=R_\theta(f_v),\\
 \rho_{\alpha,P_\theta}(f_v)-\inf_{f\in\mathcal F}
 \rho_{\alpha,P_\theta}(f)
 &=\frac{2\gamma}{d}\sum_{j=1}^d\mathbf1\{v_j\ne\theta_j\}.
 \end{split}
 \label{eq:b:CVaR-hamming}
\end{equation}
The minimizing predictor is $f_\theta$.

Let $Q_\theta=P_{\theta,\mathrm{obs}}^{\otimes n}$ and let
$\theta^{(j)}$ be obtained by reversing the $j$th coordinate.
For one observation, the two distributions differ only on
$\{X=j,S=1\}$, which has probability $\beta\pi_0/d$ under both laws.
Conditioning on $(X,S)$ and using the chain rule for relative entropy
therefore gives
\begin{align}
 D_{\mathrm{KL}}(Q_\theta\Vert Q_{\theta^{(j)}})
 &=\frac{n\beta\pi_0}{d}
 D_{\mathrm{KL}}\!\left(
 \operatorname{Bernoulli}(1/2+\gamma)
 \,\Vert\,\operatorname{Bernoulli}(1/2-\gamma)\right)\notag\\
 &\leq\frac{16n\beta\pi_0\gamma^2}{d}=\frac14.
 \label{eq:b:KL}
\end{align}
Here the symmetric Bernoulli divergences both equal
$2\gamma\log\{(1+2\gamma)/(1-2\gamma)\}\leq16\gamma^2$ for
$\gamma\leq1/4$, and independence supplies the factor $n$.
Pinsker's inequality gives
\begin{equation}
 \operatorname{TV}(Q_\theta,Q_{\theta^{(j)}})
 \leq\sqrt{D_{\mathrm{KL}}(Q_\theta\Vert Q_{\theta^{(j)}})/2}
 \leq\frac1{2\sqrt2}\leq\frac12.
 \label{eq:b:TV}
\end{equation}

Consider any measurable learner taking values in $\mathcal F$,
written $\widehat f=f_{\widehat v}$. For a fixed coordinate $j$, pair
each $\theta$ with $\theta_j=1$ with $\theta^{(j)}$, and write
$A_j=\{\widehat v_j=1\}$. The sum of the two testing errors satisfies
\begin{align*}
 Q_\theta(\widehat v_j\ne\theta_j)
 +Q_{\theta^{(j)}}(\widehat v_j\ne\theta_j^{(j)})
 &=Q_\theta(A_j^c)+Q_{\theta^{(j)}}(A_j)\\
 &\geq1-\operatorname{TV}(Q_\theta,Q_{\theta^{(j)}})\geq\frac12.
\end{align*}
There are $2^{d-1}$ such pairs. Averaging over all $2^d$ vertices gives
$2^{-d}\sum_\theta Q_\theta(\widehat v_j\ne\theta_j)\geq1/4$.
Sum this inequality over $j$ and use \eqref{eq:b:CVaR-hamming} to obtain
\[
 \frac1{2^d}\sum_\theta
 \mathbb E_{Q_\theta}\left[
 \rho_{\alpha,P_\theta}(\widehat f)-
 \inf_{f\in\mathcal F}\rho_{\alpha,P_\theta}(f)\right]
 \geq\frac{2\gamma}{d}\frac d4
 =\frac1{16}\sqrt{\frac{d}{n\pi_0\beta}}.
\]
The supremum over $\theta$ is at least this average, for every learner.
Taking the infimum over learners proves Theorem~\ref{thm:main:lower-bound}.
Randomized learners satisfy the same bound: append their independent
random seed to each experiment, which leaves the neighboring total
variation unchanged, and apply the identical testing argument.
\end{proof}

Only the expected $n\pi_0\beta$ revealed observations in the tail stratum
distinguish the vertices. This is an expected excess-CVaR lower bound for
$\mathcal F$-valued learners, already with fully identified risks at $\Gamma=1$.

\section{Revelation Theory and Algorithms}
\label{app:c}
Fix $f$, $\alpha\in(0,1)$, $\beta=1-\alpha$, and finite $\Gamma\geq1$. Use the
historical $\pi,p_1$, sensitivity bounds $[\underline p_\Gamma,\overline p_\Gamma]$,
and $s_\Gamma=\overline p_\Gamma-\underline p_\Gamma$. Write $\mu=P_X$ and
$\mathbb E=\mathbb E_\mu$. All kernels and policies are measurable; conditional
statements apply on components of positive population mass.

\subsection{Retained-information model and compatible laws}
\label{app:retained-class}\label{app:audit-information}
Additional revelation $A\in\{0,1\}$ satisfies $A\leq1-S$, with intensity
$\lambda(x)=P(A=1\mid X=x,S=0)\in[0,1]$. The audited population law has known
$p_A(x)=P(Y=1\mid X=x,A=1)$ satisfying
\begin{equation}
 \underline p_\Gamma\leq p_A\leq\overline p_\Gamma
 \quad\text{on }\{(1-\pi)\lambda>0\}.
 \label{eq:c-audit-bounds}
\end{equation}
Extend $p_A$ within these bounds on its zero-mass region; this has no effect on risks.

\begin{definition}[Retained-sensitivity compatibility]\label{def:c-retained-class}
The class $\mathcal P^{\mathrm{ret}}_{\Gamma,\lambda}$ consists of laws of
$(X,S,A,Y)$ with $X$-marginal $\mu$ and conditional components
\[
\begin{array}{c|c|c}
(S,A)&P(S,A\mid X=x)&P(Y=1\mid X=x,S,A)\\\hline
(1,0)&\pi(x)&p_1(x)\\
(0,1)&(1-\pi(x))\lambda(x)&p_A(x)\\
(0,0)&(1-\pi(x))(1-\lambda(x))&q(x).
\end{array}
\]
The residual kernel ranges freely over
$q\in[\underline p_\Gamma,\overline p_\Gamma]$; $(S,A)=(1,1)$ has zero mass.
\end{definition}

\begin{lemma}[Historical compatibility]\label{lem:c-compatible-laws}
Every admissible $q$ defines a retained law whose marginal belongs to
$\mathcal P_\Gamma(P_{\rm obs})$, with original hidden kernel
\begin{equation}
 p_0=\lambda p_A+(1-\lambda)q\quad\text{on }\{\pi<1\}.
 \label{eq:c-original-kernel}
\end{equation}
The retained class is nonempty.
\end{lemma}
\begin{proof}
Apply the kernel construction of Appendix~\ref{app:compatibility} to the table.
Its masses sum to one and its selected component is unchanged. Conditional total
probability gives \eqref{eq:c-original-kernel}; convexity of the sensitivity interval
proves compatibility. Taking $q=\underline p_\Gamma$ proves nonemptiness.
\end{proof}
At $\lambda=0$ the historical class is recovered; at $\lambda=1$ no residual
kernel remains. Membership requires the audited and residual bounds separately:
a bound on their mixture $p_0$ alone does not imply either component bound.

Known randomization $A\perp Y\mid(X,S=0)$ supplies extra information: $p_A=p_0$
on the audited support, and $q=p_0$ wherever residual mass remains. Thus the
population audited law identifies $p_0$ wherever originally hidden cases have
positive audit probability. The retained class instead keeps its residual kernel
free within the displayed bounds. Finite audits additionally require estimation
and uncertainty control for the audited contribution; a positive population audit
probability alone supplies no finite-sample precision guarantee.

\subsection{Post-revelation endpoints and tail identification value}
\label{app:tiv}
Use $a^y_{f,\eta},d_{f,\eta},b^\sigma_{f,\eta}$ from Section~\ref{sec:population},
write $s_\Gamma=\overline p_\Gamma-\underline p_\Gamma$, and set
$w_\lambda=(1-\pi)(1-\lambda)$. The known contribution and a residual
kernel's threshold objective are
\begin{align}
 K_{f,\eta}(\lambda)
 &=\mathbb E[\pi(a^0_{f,\eta}+p_1d_{f,\eta})
 +(1-\pi)\lambda(a^0_{f,\eta}+p_Ad_{f,\eta})],
 \label{eq:c-known-contribution}\\
 G_q(\eta)&=\eta+\beta^{-1}\{K_{f,\eta}(\lambda)
 +\mathbb E[w_\lambda(a^0_{f,\eta}+q d_{f,\eta})]\}.
 \label{eq:c-objective-q}
\end{align}
Conditioning on the table and optimizing the affine residual term give
\begin{equation}
 \begin{split}
 L_{f,\eta}(\lambda)&=\eta+\beta^{-1}
 \{K_{f,\eta}(\lambda)+\mathbb E[w_\lambda b^L_{f,\eta}]\},\\
 U_{f,\eta}(\lambda)&=\eta+\beta^{-1}
 \{K_{f,\eta}(\lambda)+\mathbb E[w_\lambda b^U_{f,\eta}]\}.
 \end{split}\label{eq:c-retained-objectives}
\end{equation}
\begin{proposition}[Sharp retained-sensitivity interval]\label{prop:c-sharp-endpoints}
The attainable population risks are exactly
\[
 \{\rho_{\alpha,Q}(f):Q\in\mathcal P^{\mathrm{ret}}_{\Gamma,\lambda}\}
 =[\min_{\eta\in[0,1]}L_{f,\eta}(\lambda),
   \min_{\eta\in[0,1]}U_{f,\eta}(\lambda)].
\]
Both threshold minima and endpoint laws are attained.
\end{proposition}
\begin{proof}
On $H=\{x:\pi(x)<1\}$, the common kernels $q_f^L,q_f^U$ of
Appendix~\ref{app:extremal} satisfy
$a^0_{f,\eta}+q_f^\sigma d_{f,\eta}=b^\sigma_{f,\eta}$ for every $\eta\in[0,1]$.
On $H^c$, the residual weight $w_\lambda=(1-\pi)(1-\lambda)$ is zero.
Thus $L_{f,\eta}(\lambda)\leq G_q(\eta)\leq U_{f,\eta}(\lambda)$ for every
admissible $q$, with equality at the corresponding common kernel.
Taking minima proves the endpoint claims;
continuity of each hinge objective on $[0,1]$ gives attainment. For
$q_t=(1-t)q_f^L+tq_f^U$,
\[
 \sup_\eta|G_{q_t}(\eta)-G_{q_s}(\eta)|
 \leq |t-s|\beta^{-1}\mathbb E[w_\lambda s_\Gamma].
\]
The infimum inequality \eqref{eq:a:infimum-stability} makes the optimized risk
continuous in $t$. This admissible path joins the endpoints and fills their interval.
\end{proof}

\begin{proposition}[Exact revelation value]\label{prop:c-tiv}
With
\begin{align}
 \tau_{f,\eta}&=(1-\pi)s_\Gamma|d_{f,\eta}|/\beta,
 \label{eq:c-tiv}\\
 W_{f,\eta}(\lambda)&:=U_{f,\eta}(\lambda)-L_{f,\eta}(\lambda)
 =\mathbb E[(1-\lambda)\tau_{f,\eta}]
 =W_{f,\eta}(0)-\mathbb E[\lambda\tau_{f,\eta}],
 \label{eq:c-linear-gap}
\end{align}
every bounded measurable $h$ and feasible $t$ satisfy
\begin{equation}
 W_{f,\eta}(\lambda+th)-W_{f,\eta}(\lambda)
 =-t\mathbb E[h\tau_{f,\eta}].\label{eq:c-feasible-derivative}
\end{equation}
\end{proposition}
\begin{proof}
Subtract \eqref{eq:c-retained-objectives}: $K$ cancels and
$b^U-b^L=s_\Gamma|d|$. Substitution gives the perturbation identity, including
one-sided feasible directions. Since $0\leq\tau\leq\beta^{-1}$, all terms are
integrable. The fixed-threshold reduction is independent of $p_A$, even when
that law varies with the policy.
\end{proof}

\subsection{Minimum-cost precision and the exact frontier}
\label{app:frontier}
Fix $f,\eta,\Gamma$, abbreviate $\tau=\tau_{f,\eta}$, and let $c\geq0$ be
measurable with $\mathbb E[c]<\infty$. Set $T=\mathbb E[\tau]=W_{f,\eta}(0)$,
$a=(T-\varepsilon)_+$ for $\varepsilon\geq0$, and
$F=\mathbb E[\tau\mathbf1_{\{c=0\}}]$. By \eqref{eq:c-linear-gap}, the
precision constraint is $\mathbb E[\tau\lambda]\geq a$.

\begin{proof}[Proof of Theorem~\ref{thm:cert-frontier}]
We establish attainment and the primal--dual identity
\begin{equation}
 \operatorname{Cert}_{f,\eta}(\varepsilon)
 =\min_{0\leq\lambda\leq1}\{\mathbb E[c\lambda]:\mathbb E[\tau\lambda]\geq a\}
 =\sup_{u\geq0}\{ua-\mathbb E[(u\tau-c)_+]\}.
 \label{eq:c-primal-dual}
\end{equation}
Pointwise, $(c-u\tau)\lambda\geq-(u\tau-c)_+$, so every feasible policy obeys
\begin{equation}
 \mathbb E[c\lambda]\geq u\mathbb E[\tau\lambda]-\mathbb E[(u\tau-c)_+]
 \geq ua-\mathbb E[(u\tau-c)_+].\label{eq:c-weak-duality}
\end{equation}
The following four cases attain this lower bound, or its limit.

\emph{No reduction ($a=0$).} Take $\lambda=0,u=0$, giving zero cost; this includes $T=0$.

\emph{Free reduction ($0<a\leq F$).} Take
$\lambda=(a/F)\mathbf1_{\{c=0,\tau>0\}}$ and $u=0$. The reduction is exactly
$a$ at zero cost, including when the free component has atoms.

\emph{Interior reduction ($F<a<T$).} On $D=\{c>0,\tau>0\}$, use the finite
measure $\nu(B)=\int_B\tau\,d\mu$ and $r=c/\tau$. Put $b=a-F\in(0,\nu(D))$
and choose
\begin{equation}
 u_* =\inf\{v\geq0:\nu(D\cap\{r\leq v\})\geq b\}.
 \label{eq:c-weighted-quantile}
\end{equation}
Because $r\in(0,\infty)$ $\nu$-almost everywhere, $0<u_*<\infty$.
Continuity of a finite measure along monotone sets gives
$H:=\nu(D\cap\{r<u_*\})\leq b\leq H+J$, where
$J:=\nu(D\cap\{r=u_*\})$. Set $\theta=(b-H)/J$ if $J>0$ and $\theta=0$
otherwise (then $H=b$). The policy
\begin{equation}
 \lambda_* =\mathbf1_{\{c=0,\tau>0\}}
 +\mathbf1_{D\cap\{r<u_*\}}+\theta\mathbf1_{D\cap\{r=u_*\}}
 \label{eq:c-optimal-policy}
\end{equation}
achieves reduction $F+H+\theta J=a$ and equality in the pointwise dual bound.
Its cost equals $u_*a-\mathbb E[(u_*\tau-c)_+]$. It ranks by $\tau/c$ with
threshold $1/u_*$; fractional ties handle atoms without an atomlessness assumption.

\emph{Full reduction ($a=T>F$).} Feasibility forces
$\mathbb E[\tau(1-\lambda)]=0$, hence $\lambda=1$ almost surely on $\{\tau>0\}$.
The minimum cost is $\mathbb E[c\mathbf1_{\{\tau>0\}}]$, attained by revealing
exactly that set, while
\[
 uT-\mathbb E[(u\tau-c)_+]=\mathbb E[\min\{u\tau,c\}]
 \uparrow\mathbb E[c\mathbf1_{\{\tau>0\}}]
\]
by monotone convergence. Thus the dual supremum agrees even without a finite
maximizer. Its shortfall is $\mathbb E[(c-u\tau)_+\mathbf1_{\{\tau>0\}}]$,
so a finite optimizer exists exactly when $c\leq u\tau$ almost surely on $\{\tau>0\}$ for
some finite $u$. Full and free reduction correspond to value-density thresholds
$0$ and $\infty$. Zero-value locations can always be left unselected; if $F=T$
the whole frontier is free. These cases exhaust $0\leq a\leq T$.
\end{proof}

\subsection{From threshold precision to optimized CVaR precision}
\label{app:optimized-width}
Denote the endpoints in Proposition~\ref{prop:c-sharp-endpoints} by
$\underline\rho^{\rm ret}_{\alpha,\Gamma,\lambda}(f)$ and
$\overline\rho^{\rm ret}_{\alpha,\Gamma,\lambda}(f)$, and their difference by
$I^{\rm ret}_{\Gamma,\lambda}(f)$.
\begin{proposition}[Controlling the optimized CVaR interval]\label{prop:c-optimized-width}
For the fixed predictor,
\begin{equation}
 0\leq I^{\rm ret}_{\Gamma,\lambda}(f)
 \leq\sup_{\eta\in[0,1]}W_{f,\eta}(\lambda)=W_{f,0}(\lambda).
 \label{eq:c-optimized-width}
\end{equation}
Theorem~\ref{thm:cert-frontier} at $\eta=0$ therefore minimizes cost for the
sufficient condition $W_{f,0}(\lambda)\leq\varepsilon$.
\end{proposition}
\begin{proof}
If $\eta_L$ minimizes $L$, then
$\min_\eta U-\min_\eta L\leq U_{f,\eta_L}-L_{f,\eta_L}=W_{f,\eta_L}$;
nonnegativity follows from ordered endpoints. For
$v=\max_y\ell(f(x),y)$ and $w=\min_y\ell(f(x),y)$,
\[
 |d_{f,\eta}(x)|=(v-\eta)_+-(w-\eta)_+=(v-\max\{w,\eta\})_+.
\]
This decreases in $\eta$ and equals $v-w$ at zero. Integration against
$(1-\lambda)(1-\pi)s_\Gamma/\beta\geq0$ proves the supremum identity.
\end{proof}

\paragraph{Direct optimized-width design.}
Fix an admissible set $\Lambda_{\rm adm}$ of measurable revelation intensities and
a specified family of population audit laws
$\{P^\lambda:\lambda\in\Lambda_{\rm adm}\}$ sharing one $(X,S,Y)$-marginal.
Each $P^\lambda$ has intensity $\lambda$ and satisfies
Definition~\ref{def:c-retained-class}. Set
$p_A^\lambda(x)=P^\lambda(Y=1\mid X=x,A=1)$, with an extension within the
sensitivity bounds on zero-mass regions.
Write $I^{\mathrm{ret}}_{\Gamma,\lambda}(f;p_A^\lambda)$ for the width obtained
using this kernel in \eqref{eq:c-known-contribution}, retaining the original
free residual bounds. For the cost density $c$ above and $\varepsilon\geq0$, define
\begin{equation}
 \operatorname{Cost}^{\mathrm{CVaR}}_f(\varepsilon)
 :=\inf_{\lambda\in\Lambda_{\rm adm}}
 \left\{\mathbb E[c\lambda]:
 I^{\mathrm{ret}}_{\Gamma,\lambda}(f;p_A^\lambda)\leq\varepsilon\right\},
 \label{eq:c-direct-optimized-cost}
\end{equation}
with $\inf\varnothing=+\infty$. The audited contribution can affect the separate
threshold minima even though it cancels from $W_{f,\eta}(\lambda)$.

\paragraph{Strictness of the sufficient bound.}
Take constant $X$, zero-one loss, $f=0$, $\pi=p_1=p_A=1/2$, $\lambda=1/2$,
$\Gamma=2$, and $\beta=1/4$. The original bounds are $[1/3,2/3]$.
Under the retained lower and upper endpoint laws, the probabilities $r_L$ and
$r_U$ of unit loss $L_f=1$ are
\[
 r_L=\tfrac14+\tfrac18+\tfrac1{12}=\tfrac{11}{24},\qquad
 r_U=\tfrac14+\tfrac18+\tfrac16=\tfrac{13}{24}.
\]
Both exceed $\beta$, so \eqref{eq:a:bernoulli-cvar} gives optimized endpoints
one and $I^{\rm ret}_{\Gamma,\lambda}(f)=0$, although
$W_{f,0}(\lambda)=\{(1/2)(1/2)/(1/4)\}(2/3-1/3)=1/3$.
Thus the sufficient criterion can require revelation when optimized risk is
already identified. Predictor reoptimization and estimated acquisition values
require their own statistical or sequential analysis.

\subsection{Finite-cell computation and operational cost units}
\label{app:finite-cell}
For cells $j=1,\ldots,K$ with masses $\mu_j>0$, write
$\tau_j=\tau_{f,\eta}(j)$, $c_j=c(j)$, and set
$v_j=\mu_j\tau_j$, $w_j=\mu_jc_j$, $T=\sum_jv_j$, $a=(T-\varepsilon)_+$,
and $F=\sum_{j:c_j=0}v_j$. The exact fractional algorithm is:
\begin{enumerate}
\item If $a=0$, reveal nothing. If $0<a\leq F$, reveal fraction $a/F$ of the
zero-cost, positive-value component; both cases cost zero.
\item Otherwise reveal all useful free cells. Order the positive-cost,
positive-value cells by decreasing $\tau_j/c_j$; set
$R_k=\sum_{j\leq k}v_j$, $C_k=\sum_{j\leq k}w_j$, and $R_0=C_0=0$.
\item For $b=a-F$, let $k=\min\{j:R_j\geq b\}$. Reveal the cells before $k$,
fraction $\theta=(b-R_{k-1})/v_k$ of cell $k$, and none after it. Then
\begin{equation}
 \operatorname{Cert}_{f,\eta}(\varepsilon)
 =C_{k-1}+\theta w_k
 =C_{k-1}+\frac{b-R_{k-1}}{\tau_k/c_k}.
 \label{eq:c-finite-frontier}
\end{equation}
\end{enumerate}
Tied ratios may be ordered arbitrarily; full reduction gives $\theta=1$ in the
last useful cell. If there are no positive-cost useful cells, the first step
covers every feasible target. This is Theorem~\ref{thm:cert-frontier}'s policy,
so the frontier is piecewise linear. For fixed cell policies and audited laws,
the piecewise-affine argument of Proposition~\ref{prop:b:threshold-search}
optimizes both retained objectives over
$\{0,1\}\cup\{\ell(f(j),y):1\leq j\leq K,\ y\in\{0,1\}\}$.

If a hidden label costs $\kappa(x)\geq0$, the target-population cost density and
operational value density are
\begin{align}
 c(x)&=(1-\pi(x))\kappa(x),\qquad
 \mathbb E[c\lambda]=\mathbb E[(1-\pi)\lambda\kappa],\label{eq:c-label-cost}\\
 \frac{\tau_{f,\eta}(x)}{c(x)}
 &=\frac{s_\Gamma(x)|d_{f,\eta}(x)|}{\beta\kappa(x)}
 \quad\text{when }(1-\pi)\kappa>0.\label{eq:c-operational-density}
\end{align}
The hidden-mass factor cancels; unit label cost ranks already-hidden cases by
$s_\Gamma|d|/\beta$. Cell masses likewise cancel from $v_j/w_j$.
The algorithm allocates divisible population mass. Whole-individual batches add
integer constraints and sampling variation. Sequential retraining changes losses
and acquisition values; confidence guarantees across adaptive rounds additionally
require control of newly revealed contributions, acquisition history, and stopping.

\section{Experimental Protocols and Results}
\label{app:d}\label{app:experiments}

This appendix specifies the experiments in Section~\ref{sec:experiments}:
finite-cell identification and estimation, historical applicant assessment,
controlled masking, and targeted revelation. All results use saved outputs;
Appendix~\ref{app:implementation-inventory} identifies the accompanying provenance.

\subsection{Evaluation targets, interval construction, and aggregation}
\label{app:evaluation-targets}

Write $\beta=1-\alpha$, $\rho^L=\underline\rho_{\alpha,\Gamma}$,
$\rho^U=\overline\rho_{\alpha,\Gamma}$, and $I_\Gamma=\rho^U-\rho^L$.
Endpoints minimize their respective threshold objectives. Finite-support minima
are found exactly at $0$, $1$, and all label-specific loss breakpoints, including
when estimated hinge coefficients are signed. Sampling and lending studies use
Brier loss $(f-y)^2$; population sweeps and the synthetic frontier use absolute loss.

\begin{center}
\small\setlength{\tabcolsep}{3pt}
\begin{tabularx}{\textwidth}{@{}>{\raggedright\arraybackslash}p{.19\textwidth}>{\raggedright\arraybackslash}X>{\raggedright\arraybackslash}X>{\raggedright\arraybackslash}p{.18\textwidth}@{}}
\toprule
Study & Target & Metric & \shortstack[l]{Replications/\\aggregation}\\
\midrule
Finite-cell sampling & Sharp population interval; its endpoints & Oracle whole-interval containment; learned-endpoint RMSE & 60 worlds $\times$ 80 samples per $n$; means/RMSE\\
Historical applicants & Applicant-level regression-extension sensitivity interval & Estimated endpoints and strict comparisons & One cohort; five-fold cross-fitting\\
Controlled masking & Full-outcome CVaR of each fitted predictor & Reference coverage; comparison errors and unresolved decisions & 20 masks; counts and mean widths\\
Fixed revelation & Fixed-threshold retained gap $W_{f,\eta}$ & Remaining gap; cost to target & 120 synthetic worlds or 20 masks; within-run normalization, then medians/IQR\\
Sequential audits & Estimated optimized retained width; held-out predictive loss & Width, mean Brier loss, CVaR; paired TIV--random differences & 20 masks; medians and descriptive IQR\\
\bottomrule
\end{tabularx}
\end{center}

For fixed $f,\eta$, the revelation target is
$W_{f,\eta}(\lambda)=\E[(1-\lambda)\tau_{f,\eta}]$, where
$\tau_{f,\eta}=(1-\pi)s_\Gamma|d_{f,\eta}|/\beta$.
It differs from the optimized width, whose endpoints are minimized separately
(Appendix~\ref{app:optimized-width}). Oracle containment, masking-reference
coverage, and sequential plug-in precision have different calibration premises.
Fixed-audit reductions normalize within each run before aggregation; sequential
widths aggregate the recorded widths. Ratios of separately aggregated values
need not equal aggregated relative reductions.

\subsection{Finite-cell identification and learned-endpoint accuracy}
\label{app:finite-sample-experiments}

\paragraph{Generator and estimators.}
Each of 60 simulated populations has covariate support $\mathcal X_0$ consisting of 30
cells, with Dirichlet$(1,\ldots,1)$ masses, independent $\pi\sim U[.2,.95]$ and $p_1\sim
U[.05,.95]$, and $p_0\mid p_1\sim U[q_-(p_1),q_+(p_1)]$ at $\Gamma=2$.

Fix $f=\operatorname{clip}_{[0,1]}(p_1+N(0,.15^2))$ within each world,
use Brier loss and $\alpha=.75$, and compute exact endpoints and generating-law
CVaR on $T_f=\{0,1\}\cup\{f(x)^2,(1-f(x))^2:x\in\mathcal X_0\}$.
For each $n\in\{1000,3000,10000,30000,100000\}$, draw 80 samples per world
(4,800 replications). The oracle uses known $p_1$ and the bounded score of
Appendix~\ref{app:oracle-calibration}; its endpoints are expanded by
\[
r_{\rm or}=\frac1\beta\sqrt{\frac{\log(4|T_f|/\delta)}{2n}},\qquad \delta=.05,
\]
then clipped to $[0,1]$. Proposition~\ref{prop:b:oracle-calibration} gives
whole-interval containment with probability at least $.95$ for this fixed predictor.
Saved generating-law CVaR coverage also equals one at every reported sample size.

On the same records, randomly assign each observation to one of two folds. In a given training cell, let $N_{11}$ count records with $(S,Y)=(1,1)$, let $N_{10}$ count records with $(S,Y)=(1,0)$, and let $N$ be the total number of training records in that cell. Estimate
\[
\widehat p_1=\frac{N_{11}+1/2}{N_{11}+N_{10}+1},
\qquad
\widehat\pi=\frac{N_{11}+N_{10}}{N},
\]
clipped to $[.01,.99]$ and $[.1,1]$, respectively; an empty training cell receives propensity $1/2$. Evaluate the locally robust score using the other fold's fits. The resulting RMSEs measure endpoint accuracy. Learned-nuisance coverage requires a separate calibration argument.

\begin{table}[!htbp]
\centering\small
\setlength{\tabcolsep}{4pt}
\begin{tabular}{rcccccc}
\toprule
$n$ & Sharp width & Radius & Outer width & Coverage & RMSE upper & RMSE lower\\
\midrule
1{,}000 & .0949 & .260 & .614 & 1.000 & .0361 & .0425\\
3{,}000 & .0949 & .150 & .395 & 1.000 & .0171 & .0160\\
10{,}000 & .0949 & .082 & .259 & 1.000 & .0087 & .0076\\
30{,}000 & .0949 & .047 & .190 & 1.000 & .0049 & .0043\\
100{,}000 & .0949 & .026 & .147 & 1.000 & .0027 & .0023\\
\bottomrule
\end{tabular}
\caption{Repeated-sampling decomposition over 4,800 replications per sample size.
Widths and oracle radii are means; coverage is the proportion of oracle outer
intervals containing both exact endpoints. RMSE uses two-fold learned-nuisance estimates
on those same sampled records.}
\label{tab:finite-sample}
\end{table}

\paragraph{Observed-tail scaling.}
For the $d=16$ tail-cell construction of Appendix~\ref{app:learning-lower-bound}, use 240 replications at each of 54 settings:
$n\in\{600,900,1400,2200,3500,5500\}$,
$\beta\in\{.10,.15,.25\}$, and $\pi_0\in\{.20,.35,.55\}$.
Set $\gamma=\min\{.20,.16\sqrt{d/(n\beta\pi_0)}\}$ and recover each cell's
sign by majority vote among revealed tail labels, assigning ties the positive sign.
The metric is $2\gamma$ times the incorrectly recovered fraction, equal to excess
CVaR for this construction. Its least-squares guide against
$1/\sqrt{n\pi_0\beta}$ has slope $.5030$, intercept $-.0018$, and
$R^2\simeq.998$; this illustration does not establish a general matching upper
bound for Appendix~\ref{app:learning-lower-bound}.

\paragraph{Population sweeps.}
The five-region absolute-loss design has region masses $\mu_j=P(X=j)$ and full-data outcome probabilities $p_{\mathrm{full},j}=P(Y=1\mid X=j)$, for $j=1,\ldots,5$. It fixes
\[
\begin{aligned}
\mu&=(.45,.15,.15,.10,.15),&p_{\rm full}&=(.04,.75,.10,.40,.50),\\
s_1&=(.80,.03,.80,.80,.80),&s_0&=(.80,.60,.80,.80,.80),\\
f&=(.05,.25,.10,.40,.50),&&s_y=P(S=1\mid X,Y=y).
\end{aligned}
\]
Exact finite-support optimization evaluates 60 tail levels on $[.55,.95]$ at
$\Gamma=50$, and 60 common historical observation probabilities on $[.05,1]$ at
$\alpha=.85,\Gamma=50$, holding the original $p_1$ fixed in the latter sweep.
An archived sweep uses 60 log-spaced $\Gamma\in[1,100]$ at $\alpha=.85$;
values below the generating sensitivity shift do not claim generating-law coverage.

\begin{figure}[!tbp]
\centering
\includegraphics[width=\textwidth]{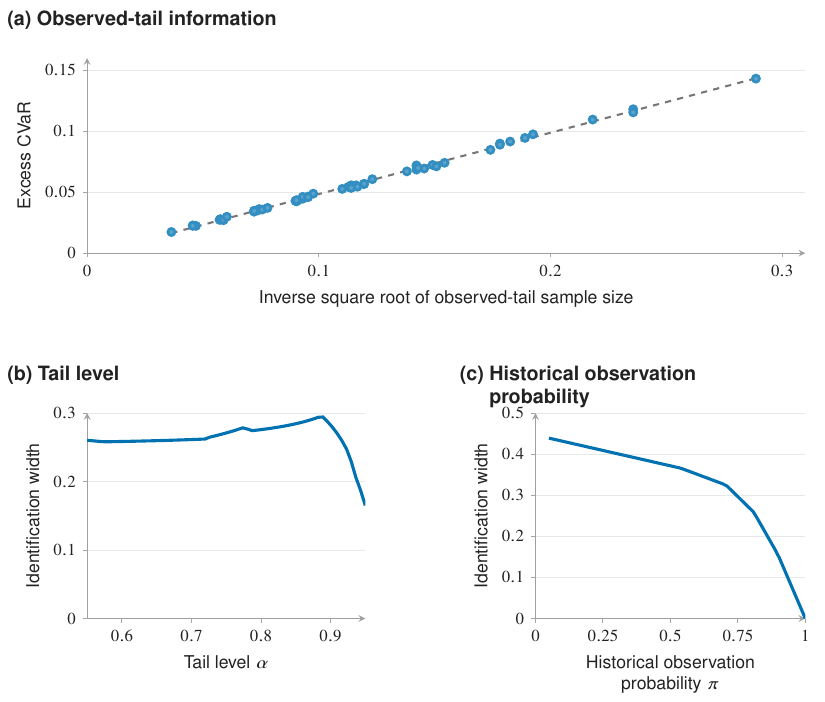}
\caption{\textbf{Observation scarcity and sensitivity shape tail-risk precision.} (a) Recorded
simulations of the Appendix~\ref{app:learning-lower-bound} learning construction, with a
least-squares guide against $(n\pi_0\beta)^{-1/2}$. Here $n\pi_0\beta$ is the expected
observed-tail sample count. (b--c) Exact identification widths for the fixed five-region
design as the tail level or historical observation probability varies. Width can vary
nonmonotonically with $\alpha$. Lines join recorded evaluations.}
\label{fig:diagnostics}
\end{figure}

\subsection{Historical LendingClub applicant assessment}
\label{app:historical-lending}

\paragraph{Cohort and shared preprocessing.}
From the public 2007--2018 accepted/rejected files \citep{george2019lendingclub},
retain loans issued and applications rejected in 2012. Mature issued outcomes
are charged-off/default ($Y=1$) or fully paid ($Y=0$), including corresponding
former-credit-policy statuses; exclude non-mature outcomes. The saved cohort has
390,644 applications: 53,367 issued loans (8,644 defaults) and 337,277 rejections.
Repayment under issuance is unobserved for rejected applications.

Features are recorded amount, debt-to-income ratio (DTI), FICO credit-score interval
midpoint or recorded risk score, employment years, state, three-digit ZIP prefix, and normalized purpose; exclude
accepted-only pricing/repayment fields, policy code, and calendar month.
Within each training sample, median-impute numeric fields with missingness
indicators, most-frequent-impute categorical fields, and one-hot encode categories
with minimum frequency 25. Standardize numeric fields for logistic regression.
Preprocessing, logistic regression, and stratified splitting use scikit-learn
\citep{pedregosa2011scikit}.

\paragraph{Eligibility, measurement, and extrapolation.}
\label{app:scope-support}
The 2012 FICO minimum was 660 \citep{lendingclub2012registration}.
Source checks identify 180,903 below-660 applications (46.31\%) and 23,097
missing standardized scores, with no issued loans in either group. All 53,367
issued loans lie among the remaining 186,644 records at or above 660.
This structural restriction does not make FICO eligibility sufficient for joint
covariate support or other underwriting requirements.

Recorded fields also differ \citep{lendingclubdictionary}: rejected applications
record FICO, whereas issued loans use an origination-score interval midpoint,
without common coarsening; issued \texttt{loan\_amnt} can reflect credit-department
reductions, whereas rejected \texttt{Amount Requested} is the requested total.
Issued purpose categories and rejected free-text titles are regex-normalized;
dates refer to funding month and application date, respectively. Thus these are
not identically measured predecision covariates.

In training fold $-k$, fit the issued-outcome regression $\widehat r_{-k}$ using
the boosting settings below, evaluate every held-out record, and specify
\begin{equation}\label{eq:d:historical-extension}
r_i=\operatorname{clip}_{[.01,.99]}\widehat r_{-k(i)}(X_i),
\qquad q_i\in[q_-(r_i),q_+(r_i)].
\end{equation}
On unsupported recorded covariates, $r_i$ is an extrapolated reference, not an
identified selected-label probability. Objectives use \eqref{eq:robust-score}
with $p=r_i$, clipped fitted propensity, and the selected-observation residual
correction. These residual-adjusted model-based estimates do not satisfy the
population-overlap premise of Theorem~\ref{thm:learning} on the full cohort.

\paragraph{Models, cross-fitting, and optimization.}
Candidates use standard classification objectives: $\ell_2$ logistic regression
($C=1$, L-BFGS, 1000-iteration limit), default-weighted logistic regression
(default class weight 2), and XGBoost \citep{chen2016xgboost} (450 trees,
depth 5, learning rate $.04$, minimum child weight 10, row/column subsampling
$.85$, $\ell_2$ penalty 2, histogram trees, recorded CUDA configuration).
Both nuisance classifiers use the boosting settings with separately fitted
preprocessing and distinct seeds: $p_1$ uses revealed outcomes and $\pi$
predicts $S$ from all training covariates. Shuffled five-fold cross-fitting is
stratified by $S$; clip $\widehat p_1$ to $[.01,.99]$ and $\widehat\pi$ to
$[.02,.98]$. Evaluate Brier loss at $\alpha\in\{.90,.95\}$ and
$\Gamma\in\{1,1.25,1.5,2,3,5,8,12\}$.

The corrected objective can be nonconvex because hinge coefficients can be signed. Enumerate
$0$, $1$, and both label-specific losses, $f(X_i)^2$ and $(1-f(X_i))^2$, at every observed
covariate $X_i$, including records whose actual outcomes are hidden. Deduplicate and sort
increasingly; among exactly equal computed minima, choose the smallest threshold, without
numerical tie tolerance. Clip endpoints to $[0,1]$; if they cross, replace both by their
midpoint. These clipping and crossing conventions are computational operations and do not
supply statistical coverage.

\paragraph{Fold enlargement and diagnostics.}
At each endpoint's selected threshold, compute five fold-specific objectives and divide
their sample standard deviation by $\sqrt5$ to obtain an empirical standard error. Let
$t_{.975,4}$ be the .975 quantile of a Student-$t$ distribution with four degrees of
freedom. Subtract this multiplier times the lower-endpoint standard error from the raw lower
endpoint and add the corresponding upper-endpoint quantity to the raw upper endpoint, before
clipping or crossing repair. Then clip both limits to $[0,1]$ and order them by minimum and
maximum. This empirical diagnostic uses dependent fold-specific quantities because their
training samples overlap; calibrated coverage requires the conditions and radius of
Theorem~\ref{thm:learning}.

Historical propensity AUC is $.9944$ and Brier score $.0201$; raw propensities
are below $.02$ for 75.26\% and above $.98$ for 1.80\% of applicants, and
inverse clipped propensities are at most 50, indicating substantial extrapolation.
Subgroups reuse full-cohort cross-fitted predictions and require at least
$\max\{250,\lfloor.002N\rfloor\}$ applications and 50 issued loans.
At $\Gamma=2,\alpha=.95$, boosting width is $.080$ for DTI above 30\%
(2.7\% issued) and $.032$ for FICO 700--739 (30.3\% issued).

\begin{table}[!htbp]
\centering\small\setlength{\tabcolsep}{3pt}
\begin{tabular}{llcccc}
\toprule
$\alpha$ & Predictor & Accepted-only & $\Gamma=1$ & $\Gamma=2$ & $\Gamma=3$\\
\midrule
.95 & Gradient boosting & .795 & .733 & [.703,.759] & [.682,.773]\\
.95 & Logistic & .801 & .805 & [.770,.831] & [.745,.843]\\
.95 & Default-weighted & .661 & .813 & [.782,.833] & [.757,.841]\\
.95 & Constant $1/2$ & .250 & .250 & [.250,.250] & [.250,.250]\\
.90 & Gradient boosting & .740 & .679 & [.625,.715] & [.575,.732]\\
.90 & Logistic & .741 & .721 & [.669,.760] & [.631,.778]\\
.90 & Default-weighted & .579 & .720 & [.679,.747] & [.646,.759]\\
.90 & Constant $1/2$ & .250 & .250 & [.250,.250] & [.250,.250]\\
\bottomrule
\end{tabular}
\caption{Historical applicants: accepted-only CVaR and model-based applicant-level
sensitivity intervals. Values other than the exact constant reference are estimates.
At $\Gamma=1$ the reference model fixes the hidden probabilities. Default-weighted denotes logistic
regression with default class weight 2.}
\label{tab:lc-parta}
\end{table}

For $f\equiv1/2$, both label-specific Brier losses equal $1/4$. Mean loss and every CVaR
endpoint therefore equal $1/4$, and $d_{f,\eta}=\mathrm{TIV}=0$ for every threshold. This
row provides an exact predictor reference; the acquisition policies compare audits for a
fixed or sequentially refitted boosting predictor.

\paragraph{Off-grid breakdown.}
For an accepted-only preferred $f$ and comparator $g$, set
$G(\Gamma)=\widehat\rho^L_\Gamma(g)-\widehat\rho^U_\Gamma(f)$.
At the first grid value $\Gamma_+$ with $G(\Gamma_+)\leq0$, if its predecessor
$\Gamma_-$ has positive gap, interpolate:
\[
\log\widehat\Gamma_{\rm br}=\log\Gamma_-+
\frac{G(\Gamma_-)}{G(\Gamma_-)-G(\Gamma_+)}
(\log\Gamma_+-\log\Gamma_-).
\]
These values interpolate the recorded grid. Boosting versus logistic gives approximate
breakdown $2.24$ at $\alpha=.95$, with first failed grid value 3, and $1.40$ at
$\alpha=.90$, with first failed value 1.5. Default-weighted versus logistic gives
approximately $1.01$ at $\alpha=.90$, with first failed value 1.25. The other three
accepted-only orderings fail already at $\Gamma=1$. A comparison with no crossing through 12
is recorded as right-censored above the grid.

\begin{figure}[!t]
\centering
\includegraphics[width=\textwidth]{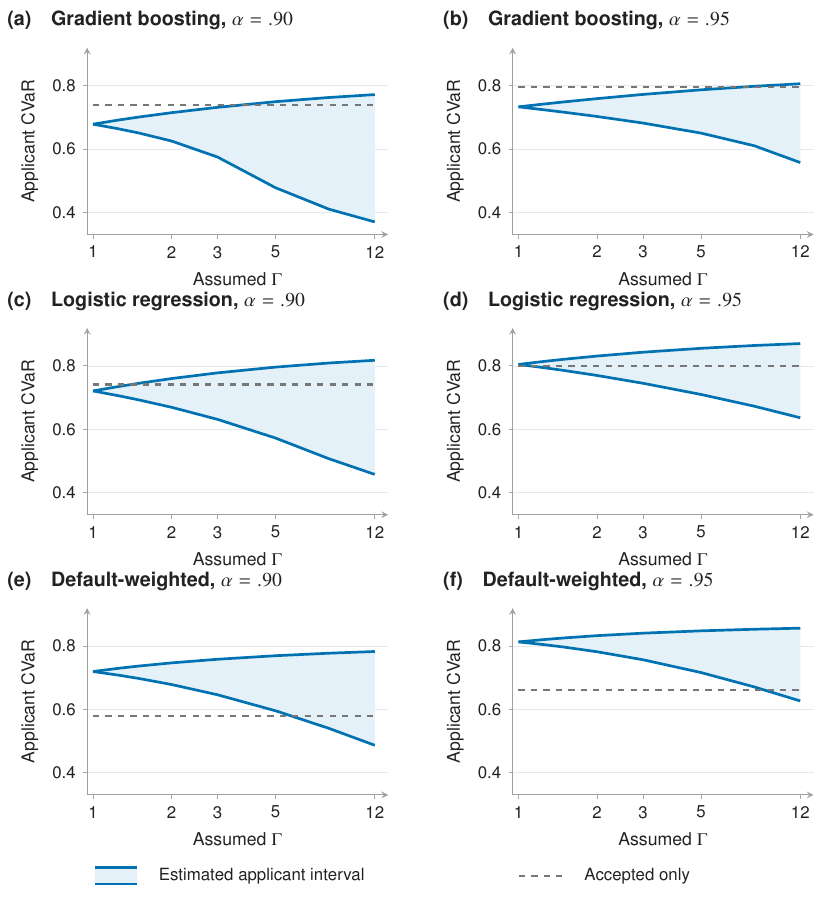}
\caption{\textbf{Complete applicant-level sensitivity analysis.}
All three predictors at $\alpha=.90$ and $.95$ over eight recorded sensitivity
levels. Blue boundaries and shading show estimated sensitivity endpoints under the applicant regression extension of Appendix~\ref{app:historical-lending}. Gray dashed lines show accepted-only CVaR. Horizontal
axes are logarithmic and vertical limits are shared. Default-weighted denotes
default-weighted logistic regression.}
\label{fig:full-sensitivity}
\end{figure}

\subsection{Controlled masking: empirical coverage and model comparisons}
\label{app:controlled-masking}\label{app:statistical-scope}

To evaluate coverage against known full-outcome risks, we use the 53,367 mature issued loans as the target population and independently draw 20 masks from
\[
\begin{aligned}
\operatorname{logit}P(S=1\mid X,Y)&=c+g(X)-Y\log3,\\
g(X)&=\operatorname{clip}_{[-3,3]}\bigl\{.65z(\text{score})-.35z(\text{DTI})\\
&\hspace{90pt}+.15z(\log(1+\text{amount}))\bigr\}.
\end{aligned}
\]
Each $z$ standardizes the median-imputed pseudo-population feature. Solve for
expected revealed share $.45$ ($c=-.0423744$). Bayes' rule gives
$\odds(p_0)/\odds(p_1)=3$, hence $\Gamma_0=3$; boundary outcome probabilities
coincide across hidden/revealed groups as in Appendix~\ref{app:compatibility}.
This validates artificial masking of mature issued loans, not outcomes for
historically rejected applicants.

All three predictors share shuffled five-fold splits stratified by imposed $S$; fitting,
clipping, threshold search, and fold enlargement follow
Appendix~\ref{app:historical-lending}. Reference CVaR uses every known outcome with that
predictor's cross-fitted predictions. Table~\ref{tab:lc-partb} counts intervals containing
this full-outcome reference risk. Table~\ref{tab:mask-comparisons} checks strict separation
of plug-in intervals against reference risk orderings. Three unordered predictor pairs per
mask and 20 masks give 60 comparisons at each tail level, or 120 across both levels.

\begin{figure}[!t]
\centering
\includegraphics[width=\textwidth]{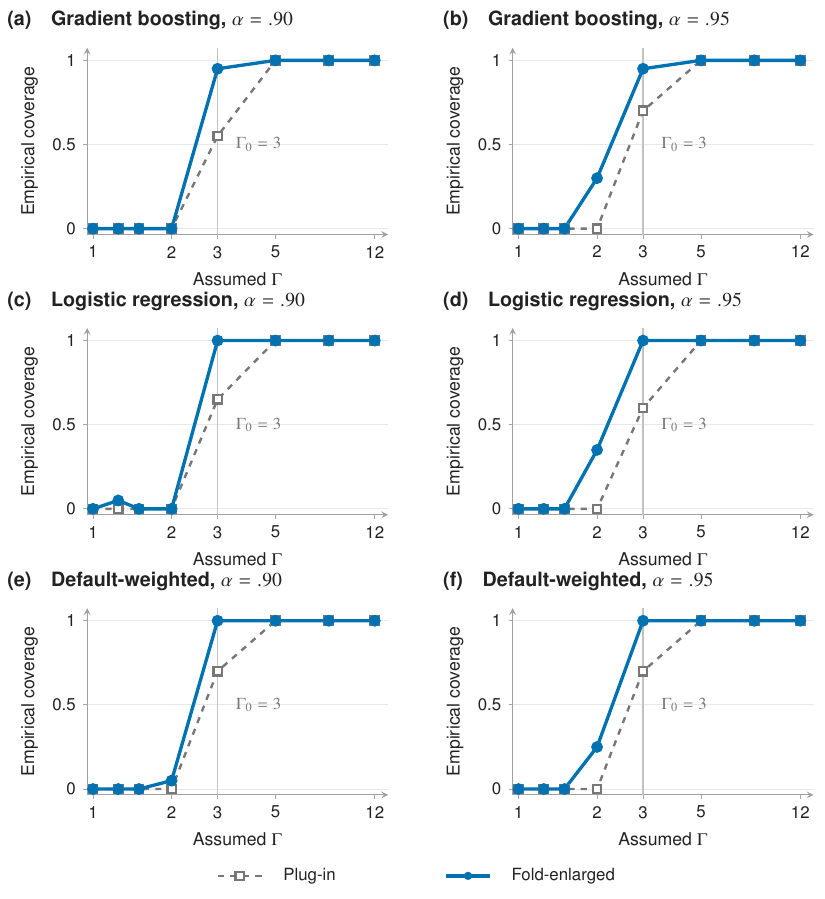}
\caption{\textbf{Empirical coverage by predictor and tail level.}
Each proportion uses 20 controlled masks. Blue circles show fold-enlarged intervals
and hollow gray squares plug-in intervals; the vertical reference marks the masking mechanism's sensitivity
$\Gamma_0=3$. Predictors share the same 20 masks, so the predictor-specific proportions are dependent. Default-weighted denotes default-weighted
logistic regression.}
\label{fig:detailed-coverage}
\end{figure}

\begin{samepage}
Figure~\ref{fig:detailed-coverage} retains the complete assumed-$\Gamma$ grid; plug-in
coverage is zero at every tested $\Gamma\le2$. Main-figure pooling averages 20 mask-level
means over three predictors, giving 60 equally weighted predictor--mask rows from 20
independent masks. Fold enlargement is an empirical calibration diagnostic; the class-wide
calibrated statement is Theorem~\ref{thm:learning}.
\par\end{samepage}

\begin{table}[!htbp]
\centering\small\setlength{\tabcolsep}{3pt}
\begin{tabular}{llcccc}
\toprule
$\alpha$ & Predictor & Plug-in & Enlarged & Mean width & Enlarged width\\
\midrule
.90 & Gradient boosting & 11/20 & 19/20 & .401 & .434\\
.90 & Logistic & 13/20 & 20/20 & .401 & .434\\
.90 & Default-weighted & 14/20 & 20/20 & .332 & .365\\
.95 & Gradient boosting & 14/20 & 19/20 & .070 & .089\\
.95 & Logistic & 12/20 & 20/20 & .066 & .084\\
.95 & Default-weighted & 14/20 & 20/20 & .110 & .139\\
\bottomrule
\end{tabular}
\caption{Controlled masking at $\Gamma=\Gamma_0=3$. The middle columns count
intervals covering the full pseudo-population reference CVaR over 20 masks;
the last two columns are mean widths. Fold enlargement follows the empirical diagnostic in Appendix~\ref{app:historical-lending}.}
\label{tab:lc-partb}
\end{table}

\begin{table}[!htbp]
\centering\small
\setlength{\tabcolsep}{7pt}
\begin{tabular}{@{}llrrrr@{}}
\toprule
$\Gamma$ & $\alpha$ & Comparisons & Resolved & Incorrect & Unresolved\\
\midrule
3 & .90 & 60 & 0 & 0 & 60\\
3 & .95 & 60 & 37 & 0 & 23\\
1 & Both levels & 120 & 120 & 16 & 0\\
\bottomrule
\end{tabular}
\caption{Model comparisons under controlled masking. Strict separation of plug-in intervals declares
an ordering, whose correctness is checked against known-outcome reference CVaRs. Counts give
resolved pairs, incorrect declared orderings, and unresolved pairs. Each tail level includes
three unordered predictor pairs in each of 20 masks.}
\label{tab:mask-comparisons}
\end{table}

\FloatBarrier
\begin{samepage}
\subsection{Synthetic fixed-threshold cost--precision frontier}
\label{app:frontier-experiments}

We next evaluate the fixed-threshold cost--precision frontier of Theorem~\ref{thm:cert-frontier} in synthetic populations with known outcome probabilities and acquisition costs.

\end{samepage}

\begin{samepage}
\paragraph{Generator.}
Use 120 worlds of 120 cells, $\alpha=.85$, $\Gamma=12$, absolute loss,
$f=p_1$, and $\eta=.5$. Masses are Dirichlet$(.75,\ldots,.75)$.
Independently draw $z_j,e_j,e_j^{\rm out},e_j^{\rm in}\sim N(0,1)$ across
cells/noise types. Hidden-tail cells satisfy $z_j+.35e_j$ at or above its
empirical $.65$ quantile (NumPy default linear interpolation). Outside/inside
these cells, respectively, set
\[
p_{1j}=\operatorname{sigmoid}(-1.35+.75z_j+.45e_j^{\rm out}),\qquad
p_{1j}=\operatorname{sigmoid}(-2.05+.35z_j+.30e_j^{\rm in}),
\]
clipped to $[.04,.72]$, and draw $\pi_j\sim U[.62,.95]$/$U[.02,.16]$.
Add to the log-odds of $p_{1j}$ a shift $N(0,.18^2)$/$U[.8\log10,\log10]$
to obtain $p_{0j}$, clipped to $[.02,.98]$. Costs are $c_j=U_j+B_jV_j$ with
$U_j\sim U[.45,1.75]$, $B_j\sim\operatorname{Bernoulli}(.20)$, and
$V_j\sim U[.75,1.75]$. Primitive draws are mutually independent and independent
of masses; hidden-cell membership induces the specified dependence of $\pi_j$
and log-odds shifts on the normal scores.

\end{samepage}

The generated $p_{0j}$ specifies the hidden-outcome distribution in cell $j$. The
fixed-threshold frontier follows from
$W_{f,\eta}(\lambda)=\sum_j\mu_j(1-\lambda_j)\tau_{f,\eta}(j)$ and cost
$\sum_j\mu_jc_j\lambda_j$. The frontier calculation uses $p_1,\pi,f,\eta,\Gamma,\mu,c$; the
generated $p_0$ does not enter these formulas.

\paragraph{Allocations and equivalent baselines.}
Here $c_j$ is population cost density: allocations cost
$\sum_j\mu_jc_j\lambda_j$, with full cost $\sum_j\mu_jc_j$.
Appendix~\ref{app:finite-cell}'s fractional-knapsack solution uses useful
zero-cost cells first, then decreasing $\tau_j/c_j$ on positive-cost/value
cells (this generator has positive costs). A fractional final cell implements
51 budgets from zero to full cost in $.02$ increments. TIV/cost and the exact
frontier share this ordering and all saved coordinates.

Seven baselines rank by raw independent $U[0,1]$ random score, cell mass,
negative revelation probability, selected-label expected absolute loss,
$p_1(1-p_1)$, sensitivity width, or absolute tail-excess contrast $|d_{f,\eta}|$. Only TIV/frontier
divide their scores by cost. Each world's random scores are independent of
generator draws and reused across budgets. Decreasing NumPy default sorting
($\texttt{argsort(score)[::-1]}$) imposes no stable secondary tie rule,
including clipping-induced ties. For $v_j=p_{1j}(1-p_{1j})$, predicted loss is
$2v_j$, predictive uncertainty is $v_j$, and
\begin{equation}\label{eq:d:equivalent-baselines}
s_\Gamma(p_{1j})
=\frac{(\Gamma^2-1)v_j}{\Gamma+(\Gamma-1)^2v_j}.
\end{equation}
The last expression increases strictly in $v_j$ for $\Gamma>1$, so these three
baselines share a mathematical weak ordering and their saved curves coincide.

Normalize remaining ambiguity by its initial value within each world before
taking medians/IQR. Cost-to-target is the first recorded budget reaching 25\%
of initial ambiguity, divided by TIV's corresponding cost in that world;
there is no interpolation between budgets. Normalization requires positive
initial ambiguity, full cost, and TIV target cost. This measures divisible
population revelation at fixed $f,\eta,\Gamma$ (Theorem~\ref{thm:cert-frontier}).

\begin{figure}[!tbp]
\centering
\includegraphics[width=\textwidth]{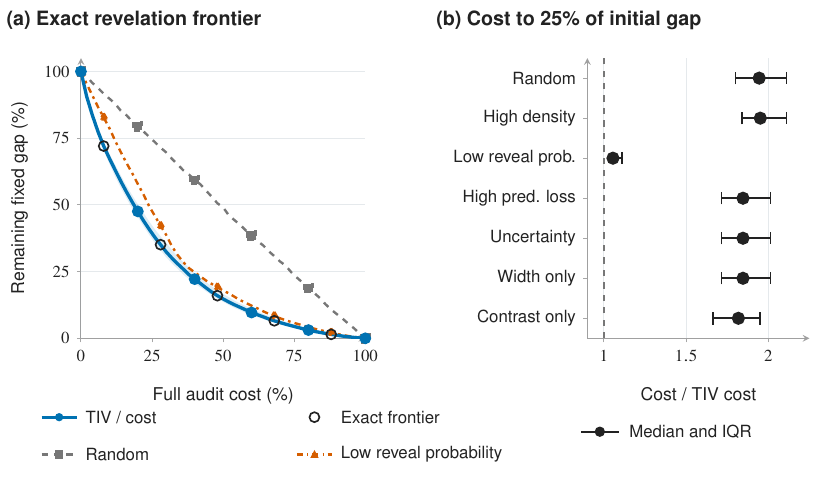}
\caption{\textbf{The exact revelation frontier and the cost of alternative policies.}
Across 120 finite-cell worlds, (a) median remaining fixed-threshold ambiguity,
normalized within each world; blue shading is the TIV interquartile range and
open markers identify the coincident exact frontier. (b) All seven baselines:
median cost of reaching 25\% of the initial gap relative to TIV in the same world,
with interquartile bars. Costs use the recorded 2\%-spaced grid.}
\label{fig:synthetic-frontier}
\end{figure}

\FloatBarrier
\subsection{Fixed and sequential audits of known repayments}
\label{app:repayment-audits}

Both protocols use boosting, $\alpha=.95$, $\Gamma=3$, the 20 controlled masks, and unit
cost per audited label. Let $N_{\mathrm{hidden,initial}}^{\mathrm{fix}}$ count initially
hidden cases in the fixed protocol's original pool, and let
$N_{\mathrm{hidden,initial}}^{\mathrm{seq}}$ count initially hidden cases in the sequential
protocol's 80\% training pool. Each protocol acquires eight batches of $\lceil
.02N_{\mathrm{hidden,initial}}^{\mathrm{protocol}}\rceil$ labels, capped by the remaining
eligible pool. Its budget fraction is the cumulative acquired count divided by that same
initial hidden count. Among candidates already known to be hidden, operational TIV is
$\widehat\tau/(1-\widehat\pi)=\widehat s_\Gamma|d_{f,\eta}|/\beta$, corresponding to
$c=1-\pi$ in Appendix~\ref{app:finite-cell}.

\begin{center}
\small\setlength{\tabcolsep}{3pt}
\begin{tabularx}{\textwidth}{@{}>{\raggedright\arraybackslash}p{.18\textwidth}>{\raggedright\arraybackslash}X>{\raggedright\arraybackslash}X@{}}
\toprule
Protocol & Fixed objective & Sequential retraining\\
\midrule
Data and predictor & Original hidden pool; initial cross-fitted predictor and historical nuisance predictions held fixed; upper threshold minimizes the empirical upper-endpoint objective in \eqref{eq:crossfit-endpoint}, built from the score \eqref{eq:robust-score}, before auditing. & 80/20 split jointly stratified by $(S,Y)$; acquire only in training pool, refit predictor on all known training labels each round\\
Evaluation & Empirical mean of remaining-hidden indicators times original per-label TIV; normalize by round zero within each mask & Training-pool retained plug-in width; held-out mean Brier and CVaR on the same reserved 20\%\\
Ordering/ties & One stable decreasing-score ordering per policy; ties retain masked-data order & Partial partition (\texttt{argpartition}); no stable secondary batch-cutoff rule; random scores refreshed each round\\
\bottomrule
\end{tabularx}
\end{center}

\begin{samepage}
The fixed protocol takes its upper threshold from the empirical upper-endpoint objective in
\eqref{eq:crossfit-endpoint}, built from the score \eqref{eq:robust-score}, using the
initial masking predictions of Appendix~\ref{app:controlled-masking} and the breakpoint/tie
rule of Appendix~\ref{app:historical-lending}. It freezes those predictor and nuisance
predictions, as well as this threshold, for every acquisition and fixed-gap calculation.
\par\end{samepage}

\begin{samepage}
\paragraph{Retained objective.}
Historical $p_1$ is fitted on initially revealed training outcomes, clipped as in
Appendix~\ref{app:historical-lending}, and frozen. For training size $n$, let
$A_{it}\leq1-S_i$ indicate audit by round $t$, $f_t$ be the refitted predictor,
and $\widehat p_i$ the frozen fit. The retained plug-in objective is
\begin{equation}\label{eq:d:sequential-objective}
\begin{aligned}
\widehat Q_t^\sigma(\eta)
={}&\eta+\frac{1}{n\beta}\sum_{i=1}^n\Bigl[
 S_i a^{Y_i}_{f_t,\eta}(X_i)
 +A_{it}a^{Y_i}_{f_t,\eta}(X_i)\\
&\quad +(1-S_i-A_{it})b^\sigma_{f_t,\eta}(X_i;\widehat p_i)
\Bigr],\qquad \sigma\in\{L,U\}.
\end{aligned}
\end{equation}
\end{samepage}
Known outcomes enter directly; still-hidden outcomes retain historical
sensitivity envelopes. Each endpoint uses breakpoint minimization, the threshold
tie rule, clipping, and midpoint collapse of Appendix~\ref{app:historical-lending}.
At full revelation $A_{it}=1-S_i$, both objectives equal
$\eta+(n\beta)^{-1}\sum_i a^{Y_i}_{f_t,\eta}(X_i)$ and both endpoints equal
full-training-data empirical CVaR; a regression test verifies this identity.
These training-prediction widths and held-out predictive risks are descriptive
trajectories, without fold enlargement or adaptive confidence guarantees.

\paragraph{Acquisition rules.}
Among still-hidden candidates, let $\widehat p_{\mathrm{known}}$ denote the fitted outcome
probability from currently known labels, and let $\widehat s_\Gamma=q_+(\widehat
p_1)-q_-(\widehat p_1)$ be the sensitivity width from the frozen historical outcome fit. In
the sequential protocol, let $\widehat\eta^U$ be the current upper-endpoint minimizing
threshold of \eqref{eq:d:sequential-objective}. The six acquisition scores are TIV per label
cost $\widehat s_\Gamma|d_{f,\widehat\eta^U}|/(\beta\kappa)$, with $\kappa=1$; independent
uniform random scores; predictive uncertainty $1-2|f-.5|$; negative fitted probability of
currently having a known label; predicted Brier loss $(1-\widehat
p_{\mathrm{known}})f^2+\widehat p_{\mathrm{known}}(1-f)^2$; and historical sensitivity width
$\widehat s_\Gamma$. The fixed protocol uses its frozen initial predictor, threshold, and
historical fits. Sequential TIV uses the current predictor and upper threshold with the
frozen historical sensitivity bounds. The propensity baseline refits the indicator of a
currently known label on all training covariates, with probabilities clipped to $[.02,.98]$.
Predicted loss refits the outcome model on currently known outcomes and uses its raw
probabilities. These models share the boosting settings above, with their respective
targets, available labels, and seeds. Refreshed acquisition fits guide label choice;
retained evaluation continues to use the historical outcome fit in
\eqref{eq:d:sequential-objective}.

\begin{table}[!htbp]
\centering\small
\begin{tabular}{@{}lrr@{}}
\toprule
Acquisition rule & \shortstack[r]{Fixed-threshold\\gap reduction}
& \shortstack[r]{Final median sequential\\interval width}\\
\midrule
TIV per label cost & 61\% & .035\\
Random & 16\% & .067\\
Lowest revelation propensity & 9\% & .084\\
Predicted loss & 0\% & .116\\
Predictive uncertainty & 0\% & .116\\
Sensitivity width & 0\% & .116\\
\bottomrule
\end{tabular}
\caption{Complete fixed and sequential audit comparison at the nominal 16\%
budget. Fixed reductions are medians of within-mask percentage reductions;
estimated sequential widths are medians across the 20 runs. Integer rounding gives actual median budget fractions of 16.013\% of the fixed initial hidden pool and 16.018\% of the sequential initial hidden training pool.}
\label{tab:audit-protocols}
\end{table}

\begin{table}[!htbp]
\centering\small
\begin{tabular}{lccc}
\toprule
Acquisition rule & Estimated width & Mean Brier & CVaR\\
\midrule
TIV & 0.0348 & 0.1362 & 0.8541\\
Random & 0.0671 & 0.1347 & 0.8682\\
Paired TIV$-$Random & $-0.0332$ & $+0.0014$ & $-0.0138$\\
\midrule
\multicolumn{4}{l}{\textit{Exact predictor reference}}\\
Constant $f\equiv1/2$ & 0.0000 & 0.2500 & 0.2500\\
\bottomrule
\end{tabular}
\par\smallskip{\footnotesize The constant forecast supplies an exact predictor reference.}

\caption{Precision and predictive loss after eight equal-budget audit rounds.
Policy rows are medians across 20 masks; the paired row is the median of the
within-mask TIV-minus-random differences. Estimated width uses the training pool;
mean Brier loss and CVaR use the same reserved 20\% split.}
\label{tab:sequential-performance}
\end{table}

\begin{figure}[!htbp]
\centering
\includegraphics[width=\textwidth]{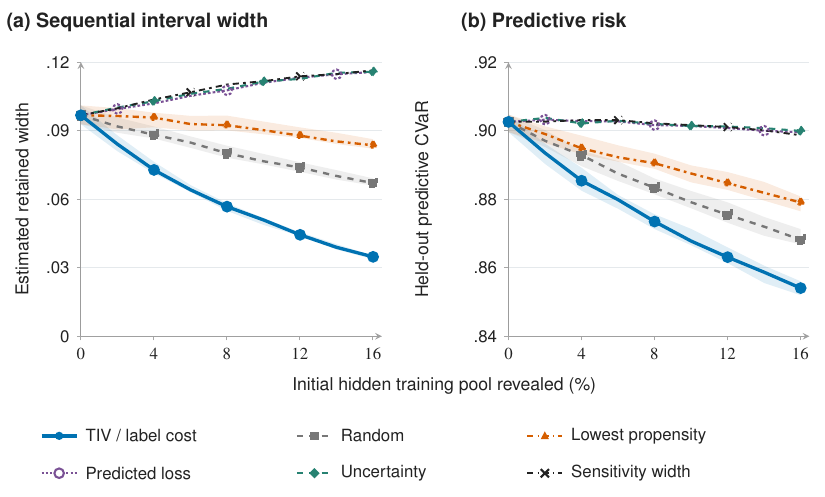}
\caption{\textbf{Sequential precision and predictive outcomes across all six acquisition rules.}
Lines are medians over 20 masks at $\alpha=.95$, $\Gamma=3$; light bands are interquartile
ranges for TIV, random, and lowest-propensity acquisition. (a) Estimated
retained-sensitivity width for the current predictor. (b) Its CVaR on the reserved 20\%
split. Predicted-loss, predictive-uncertainty, and sensitivity-width curves are plotted
separately and nearly overlap. Lines join recorded rounds.
Table~\ref{tab:sequential-performance} gives the exact constant-forecast reference. The horizontal coordinate is the median fraction of initially hidden training-pool labels acquired.}
\label{fig:all-revelation}
\end{figure}

\FloatBarrier
\begin{samepage}
In all 20 paired terminal runs, TIV has lower estimated width and held-out CVaR than random,
while random has lower mean Brier loss. The median paired differences are $-.0332$ in width,
$-.0138$ in CVaR, and $+.0014$ in mean Brier loss (Table~\ref{tab:sequential-performance}). The
exact constant reference has lower tail loss and higher mean loss than both policy medians.
Figure~\ref{fig:all-revelation} shows how the estimated width and held-out tail risk evolve
as each policy changes the training labels and predictor.

\par\end{samepage}

\FloatBarrier
\begin{samepage}
\subsection{Recorded implementation inventory}
\label{app:implementation-inventory}

Code, saved results, and reproduction instructions are available at
\url{https://github.com/E-zClap/hidden-tails}. The repository's manuscript folder
contains \path{reproducibility/README.md} and
\path{reproducibility/implementation_inventory.tex}, which record implementation entry points,
configurations, seeds, draw order, saved-output mappings, and runtime/hash logs. The
sequential results in Figure~\ref{fig:revelation-results}b, Figure~\ref{fig:all-revelation},
and Tables~\ref{tab:audit-protocols}--\ref{tab:sequential-performance} use the retained
plug-in objective \eqref{eq:d:sequential-objective}; the inventory identifies its runner and
configuration and records residual-corrected trajectories separately. It also documents the
cohort verification: all 53,367 reconstructed issued records match all 12 shared columns in
their original order, and all 20 archived mask diagnostics match.

\par\end{samepage}

\FloatBarrier

\section{Sensitivity Calibration and Extensions}
\label{app:e}\label{app:discussion}

\subsection{Choosing and calibrating the sensitivity range}
\label{app:gamma-calibration}

For interior probabilities, $\gamma=\log\Gamma$ bounds the hidden--selected log-odds discrepancy:
\begin{equation}
\begin{aligned}
|\operatorname{logit}p_0(x)-\operatorname{logit}p_1(x)|&\leq\gamma,\\
p_0(x)&\in[\operatorname{expit}\{\operatorname{logit}p_1(x)-\gamma\},
\operatorname{expit}\{\operatorname{logit}p_1(x)+\gamma\}].
\end{aligned}
\label{eq:e:log-odds-sensitivity}
\end{equation}
Overlap identifies the reference $p_1$; substantive knowledge or additional outcomes inform the allowed discrepancy, as in sensitivity analysis for nonignorable observation \citep{rosenbaum2002observational,scharfstein1999adjusting,tan2006distributional,zhao2019sensitivity}.

\paragraph{Probability-scale elicitation.}
For a reference probability $p\in(0,1)$ and elicited hidden range $[l,u]\subset(0,1)$, the smallest symmetric odds bound containing that range is
\begin{equation}
\Gamma_{\rm elic}(p,[l,u])=
\max\left\{1,\frac{\operatorname{odds}(u)}{\operatorname{odds}(p)},
\frac{\operatorname{odds}(p)}{\operatorname{odds}(l)}\right\},
\qquad\operatorname{odds}(v)=\frac{v}{1-v}.
\label{eq:e:elicited-gamma}
\end{equation}
For $p=.10$ and $[l,u]=[.05,.20]$, this gives $9/4$. A common $\Gamma$ must cover the whole target covariate support; uncertainty in $p$ can be included by maximizing over its confidence set.

\paragraph{Pilot calibration and covariate resolution.}
Declare a finite-cell model $X=g\in\mathcal G$, and write $p_{sg}=\Pr(Y=1\mid X=g,S=s)$ in cells with originally hidden mass. A pilot identifies $p_{0g}$ only when its design identifies that original hidden distribution: conditional random sampling suffices, while unequal probabilities require adjustment. An arbitrarily selected audited subset instead identifies its own law $p_A$ (Appendix~\ref{app:retained-class}). Suppose confidence sets satisfy $\Pr\{p_{sg}\in C_{sg}\text{ for all relevant }(s,g)\}\geq1-\delta_{\rm cal}$. For sets contained in $(0,1)$, define
\begin{equation}
\Gamma_{\rm cal}=\exp\left\{\max_{g\in\mathcal G}
\sup_{\substack{u\in C_{0g}\\v\in C_{1g}}}
|\operatorname{logit}u-\operatorname{logit}v|\right\}.
\label{eq:e:pilot-gamma}
\end{equation}
On the simultaneous coverage event this bounds every cell's true discrepancy. The supremum can be infinite as admissible probabilities approach incompatible boundaries. Clipping confidence sets to an interior interval preserves coverage only under a justified probability restriction or a confidence construction already supplying it. The finite-$\Gamma$ restrictions at $p_1=0,1$ are specified in Appendix~\ref{app:compatibility}.

Calibration in~\eqref{eq:e:pilot-gamma} controls the model with $X=g$. Cell averages do not control discrepancies at finer covariate resolution: either analyze with these cells or justify explicit within-cell homogeneity of the relevant probabilities. A certificate using finer covariates needs control at that finer resolution.

\paragraph{Joint error control.}
With an independent pilot, condition on it and apply the certificate at the chosen finite $\Gamma_{\rm cal}$ with valid conditional failure probability $\delta_{\rm cert}$, including nuisance estimation. A union bound gives total failure probability at most $\delta_{\rm cal}+\delta_{\rm cert}$ under the two procedures' assumptions. Reusing labels for calibration and revelation requires joint control of the sensitivity choice, audited contribution, and endpoint estimates; their dependence must enter the analysis. For a prespecified finite sensitivity grid, allocate error across its points, each already controlling the stipulated predictor class. A continuum requires uniform control in sensitivity. Let $\Gamma_{\max}$ be the largest sensitivity level under consideration. Population nesting implies that risk-limit or strict-separation claims valid at $\Gamma_{\max}$ hold throughout $[1,\Gamma_{\max}]$. Compare the calibrated range with the breakdown definition in Appendix~\ref{app:sensitivity}; Appendix~\ref{app:historical-lending} estimates breakdown numerically by interpolation on a log-sensitivity grid.

\subsection{Extensions of the outcome model}
\label{app:outcome-extensions}

Binary outcomes provide two label-specific losses whose ordering supplies one extremal hidden kernel for every threshold. A multiclass extension needs a sensitivity set for the hidden class-probability vector and a new extremality and threshold-optimization argument. Continuous outcomes require conditional-distribution sensitivity sets; unbounded losses additionally require tail conditions for finite statistical control. Population overlap and nuisance-error requirements remain those of Appendix~\ref{app:b}; the historical extension and masking calibration are specified in Appendices~\ref{app:historical-lending} and~\ref{app:controlled-masking}.

\subsection{Connections to related approaches}
\label{app:connections}

Selective-label evaluation concerns outcomes observed through historical decisions \citep{chen2025selective,coston2021fairness,lakkaraju2017selective}. Missing-data analysis and partial identification organize the resulting uncertainty through compatible laws \citep{little2019statistical,manski1989anatomy,manski1990nonparametric,manski2003partial}; here every point of the tail-risk interval has such a law. The CVaR representation accommodates discrete losses \citep{rockafellar2002conditional}. As in distributionally robust learning \citep{blanchet2019quantifying,delage2010distributionally,duchi2021uniform,mohajerin2018data}, the upper endpoint is a worst-case risk; the lower endpoint records the remaining identification information. Orthogonalization \citep{chernozhukov2018double,kallus2024localized,robins1994estimation} and cross-fitting \citep{chernozhukov2018double,kallus2024localized} control nuisance bias, and Rademacher complexity and contraction control uniform fluctuation \citep{bartlett2002rademacher,maurer2016vector,van1996weak}.

Active learning links queries to a learning objective \citep{cohn1996active}; TIV links acquisition to the precision of a specified risk claim. Its exact allocation theorem fixes $f,\eta,\Gamma$. The sequential study (Appendix~\ref{app:repayment-audits}) refreshes acquisition values after retraining while keeping the historical outcome regression fixed in the retained objective. Auxiliary fits to current known-label status or outcomes serve the corresponding acquisition baselines. The study evaluates interval trajectories and held-out predictive risk. Inference across adaptive rounds or at data-dependent stopping requires control of the acquisition history, newly revealed contributions, and stopping rule, beyond the fixed-sample uniform theorem.

\subsection{Broader impacts}
\label{app:broader-impacts}

By exposing population risks hidden by selective observation, this framework can help decision makers identify when available evidence is insufficient and direct audits toward informative outcomes. In lending or healthcare, however, a sensitivity range that excludes the true hidden outcome law, inaccurate nuisance estimates, or extrapolation into unsupported regions can produce misleading assurances and reinforce unequal access to credit or care. Even a valid population certificate does not establish fairness across demographic groups. Targeted collection may disproportionately burden particular groups and can expose sensitive financial or health information. Audit design should therefore consider subgroup coverage and acquisition burdens, use appropriate consent or authorization, and limit data collection, access, and retention. Certificates must be interpreted within the specified population, outcome, loss, and assumptions; they do not by themselves justify deployment or replace causal, fairness, privacy, and domain-specific assessments.
\FloatBarrier

\FloatBarrier
\clearpage
\section{Additional Recorded Experiments}
\label{app:f}\label{app:additional-experiments}
% Generated by figures/scripts/prepare_additional_experiments.py; do not edit.
\newcommand{\AdditionalHiddenObserved}{0.0611}
\newcommand{\AdditionalHiddenFull}{1.0000}
\newcommand{\AdditionalDiabetesGamma}{72.08}
\newcommand{\AdditionalDiabetesSelected}{0.6325}
\newcommand{\SupplementFiniteChoiceTable}{%
\begin{table}[!htbp]
\centering\small\setlength{\tabcolsep}{4pt}
\begin{tabular}{@{}lrcccc@{}}
\toprule
Rule & $f(B)$ & Observed CVaR & Full CVaR & Upper endpoint & Width\\
\midrule
Observed labels & .25 & 0.6447 & 0.7845 & 0.8877 & 0.2844\\
IPW & .25 & 0.6447 & 0.7845 & 0.8877 & 0.2844\\
Upper envelope & .65 & 0.6686 & 0.6582 & 0.8628 & 0.2286\\
Full-label oracle & .65 & 0.6686 & 0.6582 & 0.8628 & 0.2286\\
\bottomrule
\end{tabular}
\caption{Finite-candidate selection over 250 samples of size 1,400. Values average exact population evaluations of the chosen predictor; each rule selected the same candidate in every replication. The upper endpoint is a population sensitivity envelope, without statistical enlargement.}
\label{tab:additional-selection}
\end{table}
}
\newcommand{\SupplementFixedAuditTable}{%
\begin{table}[!htbp]
\centering\small\setlength{\tabcolsep}{4pt}
\begin{tabular}{@{}rccccc@{}}
\toprule
Labels & TIV & Random & Density & Uncertainty & Predicted loss\\
\midrule
\multicolumn{6}{l}{\emph{Fixed-threshold gap removed} (\%)}\\
600 & 42.4 & 30.6 & 40.7 & 15.8 & 15.8\\
1,200 & 76.2 & 61.6 & 67.8 & 47.1 & 47.1\\
1,800 & 100.0 & 91.4 & 95.5 & 88.9 & 88.9\\
\midrule
\multicolumn{6}{l}{\emph{Separately optimized empirical width removed} (\%)}\\
600 & 29.4 & 30.3 & 28.7 & 21.3 & 21.3\\
1,200 & 72.0 & 61.0 & 62.5 & 59.0 & 59.0\\
1,800 & 100.0 & 91.1 & 98.2 & 87.1 & 87.1\\
\bottomrule
\end{tabular}
\caption{One synthetic pool of 7,000 records, fixed predictor, and unit cost per hidden label. Reductions are normalized by that pool's initial gap or initial optimized width. These two targets need not rank policies identically; at 600 labels random revelation removes slightly more optimized width than TIV.}
\label{tab:additional-audit}
\end{table}
}
\newcommand{\SupplementReferenceForecastTable}{%
\begin{table}[!htbp]
\centering\small\setlength{\tabcolsep}{4pt}
\begin{tabular}{@{}lccc@{}}
\toprule
Specified forecast & Observed CVaR & Full CVaR & Upper endpoint\\
\midrule
\multicolumn{4}{l}{\emph{Synthetic populations: mean over 120 worlds}}\\
Selected-label probability $p_1$ & 0.8052 & 0.8703 & 0.8856\\
Sensitivity upper probability $q_+(p_1)$ & 0.8250 & 0.8102 & 0.8198\\
Full-label probability $p_{\rm full}$ & 0.7972 & 0.7464 & 0.8287\\
\midrule
\multicolumn{4}{l}{\emph{Diabetes-derived finite population}}\\
Selected-label probability $p_1$ & 0.2665 & 0.9252 & 0.9320\\
Sensitivity upper probability $q_+(p_1)$ & 0.8710 & 0.8693 & 0.8732\\
Full-label probability $p_{\rm full}$ & 0.5316 & 0.6760 & 0.7505\\
\bottomrule
\end{tabular}
\caption{Three specified reference forecasts, evaluated with absolute loss and $\alpha=.85$. The CSV role names ``Observed-selected,'' ``Certificate,'' and ``Oracle'' identify these formulas; they do not denote optimization over a candidate class. Synthetic entries use $\Gamma=12$; diabetes entries use its oracle-chosen sensitivity level.}
\label{tab:additional-forecasts}
\end{table}
}
\newcommand{\SupplementGridBridgeTable}{%
\begin{table}[!htbp]
\centering\small\setlength{\tabcolsep}{4pt}
\begin{tabular}{@{}rcccc@{}}
\toprule
 & \multicolumn{2}{c}{Synthetic median} & \multicolumn{2}{c}{Diabetes population}\\
Full-cost budget & Grid width & $W_{f,0}$ & Grid width & $W_{f,0}$\\
\midrule
0\% & 0.2562 & 1.3309 & 0.7295 & 1.1502\\
25\% & 0.0197 & 0.5393 & 0.2057 & 0.5857\\
50\% & 0.0080 & 0.1989 & 0.0084 & 0.1484\\
75\% & 0.0024 & 0.0599 & 0.0002 & 0.0025\\
100\% & 0.0000 & 0.0000 & 0.0000 & 0.0000\\
\bottomrule
\end{tabular}
\caption{Width connection along the fixed-$\eta=.5$ TIV-per-cost allocation. Each endpoint is minimized on the recorded 61-point cutoff grid. Synthetic columns are separate medians across 120 worlds; diabetes has one finite population. The sufficient bound is $W_{f,0}$, which is not itself a CVaR risk and can exceed one.}
\label{tab:additional-bridge}
\end{table}
}
\newcommand{\SupplementDiabetesCostTable}{%
\begin{table}[!htbp]
\centering\small\setlength{\tabcolsep}{4pt}
\begin{tabular}{@{}lr@{}}
\toprule
Allocation rule & First recorded cost to 25\% gap\\
\midrule
TIV/cost & 42\%\\
Fractional-knapsack frontier & 42\%\\
Random & 76\%\\
High density & 76\%\\
Lowest observation probability & 54\%\\
Predicted absolute loss & 44\%\\
Uncertainty & 44\%\\
Sensitivity width & 44\%\\
Excess-loss contrast & 86\%\\
\bottomrule
\end{tabular}
\caption{Diabetes-derived population: first hit on the 2\%-spaced full-cost grid. Random is the mean remaining gap over 100 random cell orderings, so its entry is the first crossing of that mean curve, not the mean of 100 crossing costs. Acquisition is divisible at the cell level.}
\label{tab:additional-diabetes-cost}
\end{table}
}

This appendix collects saved experiments that complement the studies in
Appendix~\ref{app:experiments}: simple observation mechanisms, finite-candidate
learning, additional diagnostics for the synthetic revelation frontier, and a
finite-population illustration constructed from diabetes data. They answer
questions about how the objectives behave under controlled conditions. Their
population quantities and empirical trajectories carry no additional
finite-sample confidence guarantee. No new simulations or model fits were run
when preparing these tables.

\subsection{Observation mechanisms, candidate learning, and fixed audits}
\label{app:additional-mechanisms}

\paragraph{Low observed risk can hide high population risk.}
The two-region example has masses $(.82,.18)$, deterministic outcomes $(0,1)$,
and observation probabilities $(.95,.04)$. Use absolute loss and $\alpha=.85$.
For the constant-zero predictor, the saved observed-label CVaR is
\AdditionalHiddenObserved{}, whereas population CVaR is
\AdditionalHiddenFull{}. The region-specific predictor $(0,1)$ has zero loss
under both populations. Thus this example demonstrates a large discrepancy
between observed and population risk; it does not exhibit a pairwise ranking
reversal. The saved envelope calculation uses $\Gamma=50$ and clips boundary
reference probabilities to $[10^{-8},1-10^{-8}]$, producing negligible nonzero
envelope values for the zero-loss reference.

\paragraph{Finite-candidate selection with known auxiliary functions.}
Use the five-region population specified under ``Population sweeps'' in
Appendix~\ref{app:finite-sample-experiments}. Outcomes are sampled from its
specified Bernoulli probabilities; observation is sampled conditional on the
outcome. The adverse-label observation probability in region $B$ is $.03$,
compared with $.60$ for the other label, giving an odds-ratio sensitivity shift
of $48.5$. Fix $\Gamma=50$, $\alpha=.85$, absolute loss, and
\[
 \mathcal F_0=\{(.05,b,.10,.40,.50):b\in\{.05,.25,.65,.85\}\}.
\]
Each of 250 independent training samples contains 1,400 records. Four rules
minimize, over $\mathcal F_0$, empirical full-label CVaR, empirical observed-label
CVaR, normalized inverse-propensity-weighted observed-label CVaR, or the empirical
upper sensitivity objective. The propensity rule uses the known marginal
$\pi(X)$; the upper objective uses known $p_1$ and combines observed excess loss
with the bounded hidden contribution. It does not estimate auxiliary models or
add the locally robust residual correction. All cutoff searches use exact loss
breakpoints. The full-label rule is an oracle benchmark unavailable to a learner
with selective labels.

Table~\ref{tab:additional-selection} evaluates each chosen predictor under the
exact population law. Each rule chooses the same candidate in every recorded
replication, so the table does not attach a sampling-error bar to its repeated
population evaluation. The difference between observed-label and
inverse-propensity learning, on one hand, and upper-envelope learning on the
other illustrates the consequence of outcome-dependent observation in this
particular candidate class. It is not a universal comparison between learning
algorithms.

\SupplementFiniteChoiceTable

\paragraph{A single fixed-predictor audit pool.}
A further sample of 7,000 records from the same population supplies one audit
pool. Fix the population observed-label-preferred predictor
$f=(.05,.25,.10,.40,.50)$ and the lower-envelope minimizing cutoff $\eta=.5$.
Only originally hidden outcomes can be acquired, at unit cost. The five static
rankings are per-hidden-label TIV $s_\Gamma|d_{f,\eta}|/\beta$, independent
uniform random scores, region mass, uncertainty $-|f-.5|$, and expected absolute
loss at the midpoint $(q_-+q_+)/2$ of the sensitivity interval. The implementation
sorts each score in decreasing order using NumPy's default \texttt{argsort}; it
does not impose a separate stable tie rule. The 16 recorded budgets range from
zero to 1,800 labels in increments of 120. The predictor and rankings remain
fixed throughout.

Revealed outcomes enter the known empirical contribution. The remaining hidden
outcomes retain the original sensitivity bounds. Table~\ref{tab:additional-audit}
separately reports the reduction of the fixed-cutoff gap and of the empirical
interval whose endpoints minimize their cutoffs independently. Both reductions
are normalized by their respective initial value in this one pool. The two
objectives need not favor the same allocation at every budget: random revelation
removes slightly more optimized width at 600 labels, although TIV removes more
fixed-cutoff ambiguity. Prediction uncertainty is also distinct from label-sensitive
tail ambiguity: the region with $f=.5$ has identical absolute loss under both
labels, hence zero TIV at every cutoff. These are mechanism observations from
one realized pool, without replication-based inference or a claim of adaptive
learning optimality.

\SupplementFixedAuditTable
\FloatBarrier

\subsection{Additional diagnostics for the synthetic frontier}
\label{app:additional-frontier}

Use the same 120 synthetic worlds, each with 120 cells, described in
Appendix~\ref{app:frontier-experiments}. The loss is absolute, $\alpha=.85$, and
$\Gamma=12$. The additional saved outputs compare three specified forecasts:
$p_1$, the upper sensitivity probability $q_+(p_1)$, and
$p_{\rm full}=\pi p_1+(1-\pi)p_0$. The CSV names for these rows are
``Observed-selected,'' ``Certificate,'' and ``Oracle.'' In this experiment they
are formulas fixed by the generator, not the results of minimizing empirical
risk or an upper endpoint over a candidate class. The full-label forecast uses
information unavailable from the observed law. Table~\ref{tab:additional-forecasts}
reports mean population evaluations over the 120 worlds, alongside the
single finite population introduced below. In particular, the $q_+(p_1)$ row
is not evidence that the risk-minimizing forecast has been found.

\paragraph{From a common cutoff to separately minimized endpoints.}
For each world, fix $f=p_1$ and acquire along the TIV-per-cost ordering at
$\eta=.5$. Cost is $\sum_j\mu_jc_j\lambda_j$, with a fractional terminal cell;
budget fractions divide by the full population cost $\sum_j\mu_jc_j$.
The known audited contribution is evaluated with the generative $p_0$, while the
remaining outcomes retain their sensitivity bounds. This isolates the retained
population objective from estimation of the newly audited contribution.

The saved diagnostic evaluates 21 cost fractions from zero to one. Its endpoint
search uses the grid
$\mathcal G=\{0,1/60,\ldots,1\}$, giving
\[
 I^{\mathcal G}_{\Gamma,\lambda}(f)
 =\min_{\eta\in\mathcal G}U_{f,\eta}(\lambda)
  -\min_{\eta\in\mathcal G}L_{f,\eta}(\lambda).
\]
This recorded \emph{grid-minimized width} is distinct from the exact
breakpoint-minimized width in Appendix~\ref{app:optimized-width}. The companion
bound maximizes the common-cutoff gap on the same grid. Because binary
absolute-loss contrasts satisfy $|d_{f,\eta}|\leq|d_{f,0}|$ and $0\in\mathcal G$,
that maximum equals $W_{f,0}(\lambda)$. All 2,520 saved world--budget rows satisfy
$0\leq I^{\mathcal G}_{\Gamma,\lambda}(f)\leq W_{f,0}(\lambda)$ to numerical
precision. Table~\ref{tab:additional-bridge} gives selected budgets. This check
illustrates the sufficient-width connection; it neither makes the $.5$-cutoff
allocation optimal for optimized width nor turns the finite grid into exact
threshold minimization. The gap bound can exceed one because it is not itself
a bounded-loss CVaR.

\subsection{A diabetes-derived finite-population illustration}
\label{app:additional-diabetes}

\paragraph{Measured outcomes and imposed observation.}
The construction uses all 442 records in the diabetes dataset distributed with
scikit-learn \citep{pedregosa2011scikit}. It binarizes the measured disease-progression
outcome at its empirical 75th percentile, assigning $Y=1$ at or above that
cutoff. A standardized logistic regression, fitted to all these outcomes with a
5,000-iteration limit and the recorded software defaults, supplies scores $r_i$.
Fourteen empirical-quantile score cells define the finite covariate population.
The score is used to construct strata, not evaluated as a held-out predictive
model. In particular, the construction has access to outcomes subsequently
called hidden.

For each record, impose the observation probability
\[
 s_i=\operatorname{clip}_{[.03,.95]}
 \{.85-.25r_i-.55Y_i-.20Y_i r_i+\epsilon_i\},
 \qquad \epsilon_i\sim N(0,.01^2).
\]
The saved construction integrates these probabilities; it does not draw a new
binary observation mask. For a cell $j$ with $n_j$ records, set
\[
 \mu_j=\frac{n_j}{442},\qquad
 \pi_j=\frac{1}{n_j}\sum_{i\in j}s_i,\qquad
 p_{1j}=\frac{\sum_{i\in j}Y_i s_i}{\sum_{i\in j}s_i},\qquad
 p_{0j}=\frac{\sum_{i\in j}Y_i(1-s_i)}{\sum_{i\in j}(1-s_i)}.
\]
Before evaluation, the implementation clips $p_{1j}$ and $p_{0j}$ to
$[10^{-4},1-10^{-4}]$ and uses the resulting Bernoulli mixture
$p_{{\rm full},j}=\pi_jp_{1j}+(1-\pi_j)p_{0j}$. Consequently, the reported full
risks refer to this smoothed finite-cell law. The measured progression outcomes
and covariates define the construction; the observation channel and acquisition
costs are imposed. The population observation share is
\AdditionalDiabetesSelected{}.

\paragraph{Sensitivity, loss, and costs.}
Use absolute loss and $\alpha=.85$. The sensitivity level is chosen with oracle
knowledge of both cell kernels:
\[
 \Gamma=\max\!\left\{2,\;1.10\exp\!\left(
 \max_j|\operatorname{logit}p_{0j}-\operatorname{logit}p_{1j}|\right)\right\}
 =\AdditionalDiabetesGamma{}\quad\text{(rounded)}.
\]
This construction ensures compatibility for the known cell law. It is not a
calibration procedure justified using observed labels alone, and it supplies no
simultaneous or data-dependent-$\Gamma$ confidence statement. With zero-based
cell indices, the imposed population cost density is
$c_j=.65+1.15\bar r_j+.20\mathbf1\{j\bmod3=0\}$, where $\bar r_j$ is the
cell's mean fitted score.

The three reference forecasts and their population risk evaluations follow
Appendix~\ref{app:additional-frontier}. For the acquisition comparison fix
$f=p_1$ and $\eta=.5$, use divisible cell allocations, and evaluate 51 fractions
of full cost in increments of $.02$. The acquisition rules are the seven
baselines and TIV-per-cost/frontier ordering described in
Appendix~\ref{app:frontier-experiments}. Random acquisition averages the remaining
gap over 100 independent uniform-random cell orderings. The 21-budget width
diagnostic uses the same 61-point cutoff grid as the synthetic study.

\SupplementReferenceForecastTable
\SupplementGridBridgeTable
\FloatBarrier

Table~\ref{tab:additional-diabetes-cost} reports the first recorded cost fraction
at which each curve reaches 25\% of its initial fixed-cutoff gap. For random
acquisition, this is the crossing of the mean gap curve, rather than the mean of
100 individual crossing costs. TIV and the exact fractional-knapsack frontier
coincide by their common ordering. The data illustrate how observation and costs
shape a retained population certificate when all relevant quantities are known;
they do not establish clinical effectiveness, generalization to a new patient
population, or finite-sample coverage. They also do not convert passive revelation
into identification of outcomes under untaken actions.

\SupplementDiabetesCostTable
\FloatBarrier

\subsection{Saved-output provenance and reproduction order}
\label{app:additional-provenance}

The numerical sources are the \texttt{experiment1}, \texttt{experiment2}, and
\texttt{experiment3} CSVs, and the \texttt{experiment6} and
\texttt{experiment7\_semisynthetic} CSVs under \path{experiments/results/}.
The mechanism examples use the full-sequence random stream with master seed
20260503. The current frontier and diabetes CSVs instead match
\path{run_summary_main_figure.json}, generated by running
\path{experiments/run_tail_observability_experiments.py} with
\texttt{--main-figure}. The full-run summary's older frontier output uses a different random-stream
position. The archived \path{run_summary_semisynthetic.json} instead records an
earlier 569-record malignancy example, with malignancy-rate fields; it is not
the 442-record diabetes construction in this appendix. Neither archived output
is substituted for the current CSVs.
Reproductions should use separate output copies or follow the documented run
order so that one entry point does not silently overwrite another run's saved
results.

The standard-library extraction script
\path{figures/scripts/prepare_additional_experiments.py} regenerates these tables
from the saved CSVs. Its \texttt{--check} mode verifies the generated tables,
source checksums, agreement with the matching run summary, repeated-sample
aggregation, TIV/frontier equality, and the recorded width inequalities.
\path{figures/supplement_additional_provenance.json} records those checks and the
source-to-entry-point mapping. The corrected lending sequential experiment,
original-cohort verification, and finite-sample endpoint study remain documented
in Appendix~\ref{app:experiments} and the accompanying reproduction material.

\FloatBarrier
\end{document}